\documentclass[sigconf]{acmart}
\AtBeginDocument{%
  }

\copyrightyear{2026}
\acmYear{2026}
\setcopyright{cc}
\setcctype{by-nc-nd}
\acmConference[KDD '26]{Proceedings of the 32nd ACM SIGKDD Conference on Knowledge Discovery and Data Mining V.2}{August 09--13, 2026}{Jeju Island, Republic of Korea}
\acmBooktitle{Proceedings of the 32nd ACM SIGKDD Conference on Knowledge Discovery and Data Mining V.2 (KDD '26), August 09--13, 2026, Jeju Island, Republic of Korea}
\acmDOI{10.1145/3770855.3818482}
\acmISBN{979-8-4007-2259-2/2026/08}

\usepackage{algorithm2e}
\usepackage{amsmath} 
\usepackage{multirow}
\usepackage{graphicx} 
\usepackage{subfig}
\begin{document}

\title{Smallholder Agricultural Landscape Understanding at a National Scale}

\author{Radhika Dua}
\author{Aditi Agarwal}
\author{Aishwarya Jayagopal}
\author{Depanshu Sani}
\author{Ishan Deshpande}
\affiliation{\institution{Google DeepMind}
  \city{}
  \country{}}

\author{Vishal Batchu}
\author{Gaurav Aggarwal}
\author{Alok Talekar}
\email{atalekar@google.com}
\author{Vaibhav Rajan}
\email{vaibhavrajan@google.com}
\affiliation{\institution{Google DeepMind}
  \city{}
  \country{}}

\author{Alex Wilson}
\author{Hoang Tran}
\author{Bogdan Floristean}
\author{Neelabh Goyal}
\author{Ramya Cheruvu}
\author{Yan Mayster}
\affiliation{\institution{Google}
  \city{}
  \country{}}

\renewcommand{\shortauthors}{Radhika Dua et al.}

\begin{abstract}
Comprehensive agricultural landscape understanding is critical for addressing global challenges in food security, climate change, and resource management. This requires mapping not just crop fields, but also vital features like trees and water bodies which form an intricate mosaic in complex \textit{smallholder} systems dominating the Global South. 
Previous efforts to develop such land use maps have been limited by a narrow focus on methods for field delineation only, and also do not 
develop robust post-processing steps essential for real-world deployment. Further, to our knowledge, no prior system for smallholder farms has been deployed and evaluated at a national scale. This work addresses these limitations by presenting the first national-scale agricultural mapping system that moves beyond simple field delineation to 
enable segmentation of agricultural instances like fields, trees and water bodies.
Our system is refined for real-world application using novel post-processing heuristics to ensure map consistency and accuracy, and is validated through a rigorous, multi-faceted evaluation process.
Fine-grained land use maps generated by our system are publicly accessible via an API at \textit{\href{http://agri.withgoogle.com}{http://agri.withgoogle.com}}, enabling a wide range of applications from precision agriculture and policy-making to advancing global sustainability development goals.
\end{abstract}




\maketitle

\section{Introduction}
Large-scale agricultural landscape understanding -- the comprehensive mapping and analysis of agricultural environments -- is foundational for addressing the interconnected crises of food insecurity, climate change, and resource degradation \cite{fao2023state,oecd2023agricultural}. At its core, this involves accurately delineating crop fields, which is a prerequisite for detailed, field-level analytics \cite{Yan2016Conterminous,Maxwell2018Implementation,Ma2019Deep}. Such data enables precise crop statistics, monitoring of farm health, identification of yield gaps 
and informs economic policies on resource allocation and mechanization \cite{rudel2009agricultural,Mueller2012Closing,Cai2018HighPerformance,Kang2019FieldLevel}. Furthermore, knowing the precise location and extent of fields unlocks the potential of precision agriculture by enabling tailored advice on planting, fertilization, and pest control \cite{Segoli2012Should,Salek2018Bringing,Bramley2019Farmer}.

Beyond fields, delineating trees and agroforestry systems is critical for both sustainability and resilience \cite{pancholi2023role}. Mapping these agricultural features helps quantify their role in sequestering carbon, which is key to transforming agriculture from a net source of carbon emissions into a climate solution \cite{FAOSTAT_ClimateChange2023,Gao2023GreenhouseGE}. It also informs the implementation of agroforestry to make farms more resilient to climate impacts like drought and extreme heat \cite{lasco2014agroforestry}. Similarly, identifying and monitoring water resources -- down to on-farm ponds and wells -- is indispensable for sustainable water management \cite{cofie2015water,prasad2022can}. This allows for the planning of efficient irrigation, the tracking of agricultural runoff to protect water quality, and targeted investments in water access that are crucial for boosting farmer 
incomes and food security \cite{magidi2021informing,su2024advancing}. In essence, this holistic and granular mapping of fields, trees, and water systems provides the foundational data layer required to drive smart policy, empower farmers, and eventually build a productive, sustainable, and resilient global food system.

While remote sensing and machine learning (ML) offer a scalable way to generate such land use maps, their application in smallholder farms presents significant data and modeling challenges. 
These challenges stem primarily from the inherent nature of smallholder systems, which are characterized by an intricate mosaic of very small fields (< 2 ha \cite{shdefinition}), diverse land uses, and interspersion with natural vegetation and water resources \cite{Lesiv2019Estimating,Samberg2016Subnational,Rada2019New}. 
Smallholder systems are responsible for more than half of the 
global agricultural production \cite{Samberg2016Subnational,sylvester2015success}, and continue to face the problem of large yield gaps \cite{Mueller2012Closing}.
Most successful ML models, which were developed for large-scale industrial farming systems using extensive government-provided labeled datasets, often fail to generalize effectively to different regions, satellite sensors, or image resolutions without significant recalibration \cite{donohue2018towards,GarciaPedrero2019Deep,Wagner2020Extracting,Marvaniya2021Small,Masoud2020Delineation}. 
Further, standard, freely available satellite imagery (e.g., Landsat at 30 m/pixel, Sentinel-2 at 10 m/pixel) that have been used extensively are too coarse to resolve the boundaries of individual smallholder fields \cite{Jain2016Mapping}.
Beyond the lack of labels for smallholder regions, remote sensing data, in general, presents unique technical complexities not common in standard computer vision tasks. These include the importance of the temporal dimension (capturing changes over a growing season), the use of multispectral data channels beyond RGB, high variance within the ``field'' class (due to diverse crops, shapes, and sizes), and low variance between fields and other land use classes like forests \cite{KernerEtAl2023Multi}. Compounding these issues is the distribution shift as field characteristics vary dramatically between different regions, climates, and seasons \cite{KernerEtAl2023Multi,rs14225738}.

These challenges have been addressed both at data and methodological levels in 
recent works. Increasing accessibility of very-high-resolution (VHR) satellite imagery (e.g.,  WorldView-3 data at 0.31 to 2 m/pixel), has enabled more accurate delineation of field boundaries in smallholder systems.
Most recent works have tackled the field boundary delineation problem by modeling it as a semantic segmentation task, i.e., per-pixel binary classification where the classes are determined based on either boundary or extent of the field, e.g., \cite{waldner2020deep,waldner2021detect,mei2022using}.
To achieve the final goal of delineating each field as a distinct object (i.e., instance segmentation), approaches have either relied on post-processing steps (like watershed algorithm \cite{vincent1991watersheds}) after semantic segmentation or employed direct object detection methods (e.g., Mask-RCNN \cite{he2017mask}). Finally, to address the problems of insufficient labeled data and distributional shift across geographies, transfer learning has also been utilized \cite{KernerEtAl2023Multi,rs14225738}.

Despite these advances, previous research remains limited in the following practically important ways.
First, prior research for smallholder agriculture has narrowly focused on the delineation of field boundaries alone. 
As discussed earlier, a comprehensive mapping requires a holistic understanding of multiple agricultural features, including trees and water sources like wells. 
This integrated approach is especially critical for smallholder farms, where these different landscape elements often exist together and are intricately interconnected.
Second, while previous studies have evaluated the benefits of direct instance segmentation over two-step approaches, they have not assessed the efficacy of panoptic segmentation techniques from computer vision, which unifies both semantic and instance segmentation into a single, cohesive framework.
Third, the scope of method development in prior work has often been confined to the design of the core machine learning models. This overlooks the practical necessities of real-world deployment, which demand robust post-processing steps to handle challenges such as inconsistent predictions across different satellite images that may not be geographically or temporally aligned.
Finally, a crucial gap in the existing literature is that no previous work has presented a large-scale deployment and a comprehensive evaluation of their proposed mapping system in a real-world context.
Our work addresses these limitations and makes the following contributions.

\subsection{Our Contributions}
\begin{itemize}
\item \textbf{First National-Scale Land Use Map with emphasis on Smallholder Farms} We develop and evaluate a satellite-based system for comprehensive mapping of agricultural land use at a national scale with sufficient granularity to support targeted interventions, particularly for smallholder farms. Our maps are publicly accessible through an API at \textit{\href{http://agri.withgoogle.com}{http://agri.withgoogle.com}}.

\item 
\textbf{Agricultural Instance Segmentation} 
Our instance segmentation approach enables identification of individual fields, trees and water bodies. To the best of our knowledge, we are the first to train and evaluate instance segmentation techniques on diverse agricultural features. 

\item \textbf{Refinement for real world deployment}
We design novel heuristics to post-process the model inference outputs to generate the final land use maps. These steps are crucial for ensuring the accuracy, consistency, and usability of inferred maps for downstream applications.

\item\textbf{Rigorous Evaluation} To ensure the reliability and real-world applicability of our model, we conducted meticulous evaluation, which includes evaluation on public benchmark datasets, 
comparison with in-situ surveys and 
 on-ground validation in collaboration with external partners.
\end{itemize}

\begin{figure*}[h!]
\centering
\includegraphics[width=\textwidth]{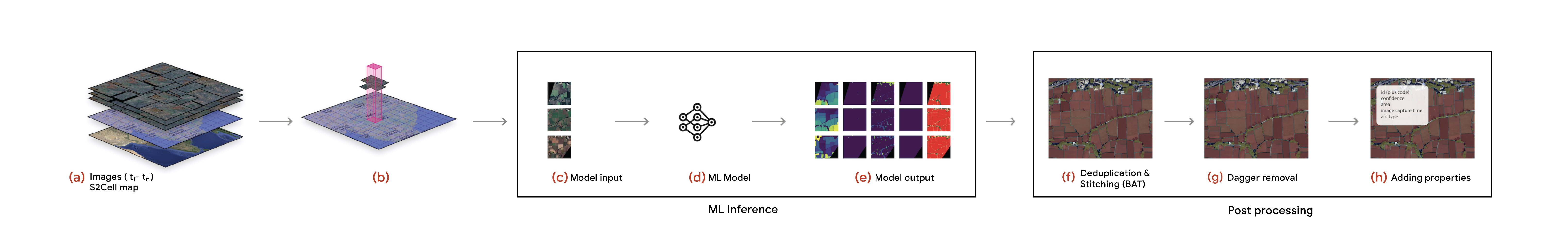}
\caption{Overview of our ALU system: (a) Complete coverage requires multiple satellite images acquired at different times ($t_{1}$ to $t_{n}$) due to the large study area and satellite orbital constraints. (b) To ensure scalability, the processing is spatially partitioned using S2 cells, a gridded sub-region of the Earth. (c-e)  A machine learning model performs multi-class panoptic segmentation on each image to identify agricultural features. (f-h)  Post-processing steps, including vectorization, de-duplication, and merging, reconcile information from overlapping images and generate a final output partitioned at s2 cell.}
\label{fig:alu_overview}
\end{figure*}

\section{Background and Related Work}
\subsection{Land Use Monitoring}
Traditional methods for land-use monitoring such as
manual and smartphone-based surveys \cite{land_survey_manual, assam_land_record, mh_land_record} 
remain valuable for localized assessments, but they are inherently labor-intensive and time-consuming. Remote sensing methods, utilizing UAVs and satellites, have emerged as a more efficient and cost-effective solution for large-scale land-use monitoring. These technologies capture imagery across extensive areas and a broad spectrum of wavelengths, providing rich data for analysis. While satellite imagery may have lower spatial resolution compared to UAV imagery, its expansive coverage and historical data availability render it particularly well-suited for monitoring agricultural landscapes and informing land management initiatives.

\subsection{Smallholder Farm Boundary Delineation} 
Numerous studies have focused on delineating field boundaries from remote sensing data by adapting techniques from computer vision (CV).
E.g., multi-task convolutional neural networks (CNN) with UNet architectures have been adapted by Waldner et al. \cite{waldner2020deep,waldner2021detect} for instance segmentation of satellite images. 
Most previous approaches have used moderate-resolution satellite images (e.g., LandSat, Sentinel-2) and have focused on large-scale farming systems in the global North \cite{GarciaPedrero2019Deep,Wagner2020Extracting,Marvaniya2021Small,Masoud2020Delineation}.

Recognizing that high-resolution satellite images are crucial to delineate boundaries in smallholder farms, recent approaches have begun using very-high-resolution imagery.
E.g., Mei et al. \cite{mei2022using}, adapt a popular object detection method, Mask-RCNN \cite{he2017mask}, and use WorldView-3 satellite imagery (0.5 m) to delineate farm boundaries in India.
Transfer learning based techniques have utilized a combination of moderate and high resolution satellite imagery to combat both problems of insufficient labelled data in smallholder regions and  
differences in farm sizes across regions.
Examples include the works of Wang et al. \cite{rs14225738} and
Kerner et al. \cite{KernerEtAl2023Multi} 
to predict farm boundaries in India and Kenya respectively.

A large number of methods have been developed, in CV, for joint instance and semantic segmentation, including 
panoptic segmentation methods which output panoptic masks simultaneously identifying semantic classes and instances,
thereby yielding comprehensive scene understanding
\cite{elharrouss2021panoptic,de2022panoptic,chuang2023deep}.
The benefit of simultaneous modeling of these two tasks, over most previous approaches
which had separate modules for the tasks and multiple passes for inference, have been demonstrated by these methods on benchmark CV datasets.
However, to our knowledge, they have not been evaluated for smallholder farm boundary delineation. 
In addition, our work differs from previous works focussed on smallholder farms in the following ways. 
We go beyond farm boundaries to enable delineation of other features of agricultural relevance -- trees, wells and water bodies -- in smallholder farms.
Further, we describe postprocessing steps required to generate accurate land use maps from ML model outputs, which are crucial for real-world deployment.


\subsection{Image Segmentation Approaches}
Image segmentation has evolved into a diverse set of tasks, most notably instance and panoptic segmentation. While instance segmentation focuses on identifying and delineating individual, countable objects (``things''), panoptic segmentation aims to unify instance and semantic segmentation into a single cohesive pipeline by also classifying amorphous background regions (``stuff''). 
Initial efforts in comprehensive scene understanding focused specifically on building panoptic pipelines by merging distinct task branches. For example, Panoptic FPN~\cite{kirillov2019panoptic} achieved panoptic segmentation by appending a dense-prediction semantic branch parallel to a proposal-based instance branch on a shared backbone. Similarly, Panoptic-DeepLab~\cite{cheng2020panoptic} was designed explicitly as a bottom-up panoptic framework, utilizing dual decoders to independently predict semantic labels and instance centers. 
More recently, the field has shifted toward fully unified architectures capable of performing semantic, instance, or panoptic segmentation seamlessly without requiring task-specific branching. K-Net~\cite{zhang2021k} introduced a unified, box-free framework that handles any of these tasks by leveraging dynamic, learnable kernels and bipartite matching. Advancing this universal approach, Mask2Former~\cite{cheng2022masked} utilizes masked attention to extract localized features, establishing a highly flexible framework for any image segmentation task, including pure instance segmentation. \\
While panoptic and unified models excel at global scene understanding, our specific objective is targeted purely at delineating discrete agricultural entities. Therefore, our method focuses exclusively on instance segmentation. Our application does not require categorizing ``stuff'' classes, such as general background, soil, or other non-agricultural regions. Furthermore, most of these general segmentation approaches have been predominantly evaluated on natural images, which differ vastly in scale, density, and texture from remote sensing imagery. To demonstrate our method's efficacy within this specialized domain, we compare our proposed instance-focused approach against the recent and adaptable Mask2Former~\cite{cheng2022masked}, evaluating it specifically on its instance segmentation capabilities.

\subsection{Remote Sensing Foundation Models}
With the availability of large scale unlabelled satellite data, remote sensing foundation models have gained traction over the recent years. SatMAE~\cite{cong2022satmae} extended the Masked Autoencoder (MAE) framework by incorporating temporal embeddings and distinct spectral positional encodings to effectively model multi-spectral time-series data. To tackle the inherent variance in spatial resolutions across different sensors, models like ScaleMAE~\cite{reed2023scale} introduced scale-aware pre-training, utilizing Ground Sample Distance (GSD) based positional encodings to ensure representations remain robust across varying image scales. Similarly, the Prithvi architecture~\cite{szwarcman2025prithvi} explicitly targets large-scale temporal dynamics by applying spatial and temporal transformer attention over massive harmonized datasets, integrating precise location metadata to enrich the embedding space. However, most of these methods require time series remote sensing data and work on multi spectral satellite data, unlike the high resolution RGB satellite snapshot data we use. As such, we benchmarked ALU against SAM2~\cite{ravi2024sam}, a foundational segmentation model which can work on RGB images.

\section{Method}

\subsection{Overview}

Figure \ref{fig:alu_overview} illustrates the overall workflow for our Agricultural Landscape Understanding (ALU) system. 
To map large geographic extents (e.g., all of India, in our study), we need to rely on multiple satellite images acquired at different times ($t_{1}$ to $t_{n}$) to achieve complete coverage (\ref{fig:alu_overview}a), as orbital constraints prevent a single satellite image from encompassing the entire region of interest. 
To ensure scalability and computational efficiency, the processing workflow is sharded by S2 cells \cite{s2_cell} (\ref{fig:alu_overview}b). 
S2 cells are a hierarchical spatial indexing system that subdivides the Earth's surface into a grid of cells, providing a convenient and efficient way to organize and access geospatial data. This spatial partitioning strategy allows for parallelization of the image processing and analysis, enabling the efficient handling of vast amounts of data required for national-scale land-use mapping.
A trained machine learning model consumes each of the collected images and performs multi-class panoptic segmentation to identify and delineate various agricultural features (\ref{fig:alu_overview}c-e). Section \ref{sec:mlmodel} describes our ML model development. 

Note that the spatio-temporal variation in image acquisition introduces the challenge of reconciling potentially disparate representations of the landscape, as weather patterns and seasonal phenomena (e.g., changes in crop stage) can significantly alter the appearance of agricultural features over time. 
To reconcile these variations, we rely on sufficient spatial overlap between images acquired at different times. The resulting segmentation masks are then subjected to a series of post-processing steps, including vectorization, stitching,  de-duplication, and  boundary refinement, to generate a comprehensive and spatially accurate representation of the agricultural landscape, effectively merging information from overlapping images while accounting for spatio-temporal inconsistencies (Fig. \ref{fig:alu_overview}f-h).
These steps are outlined in section \ref{sec:postprocessing}.
Finally, we mention additional considerations which are important for final deployment in section \ref{sec:deployment}.

\subsection{ML Model Development}
\label{sec:mlmodel}

\begin{figure}[hbt!]
\centering
\includegraphics[width=0.15\textwidth]{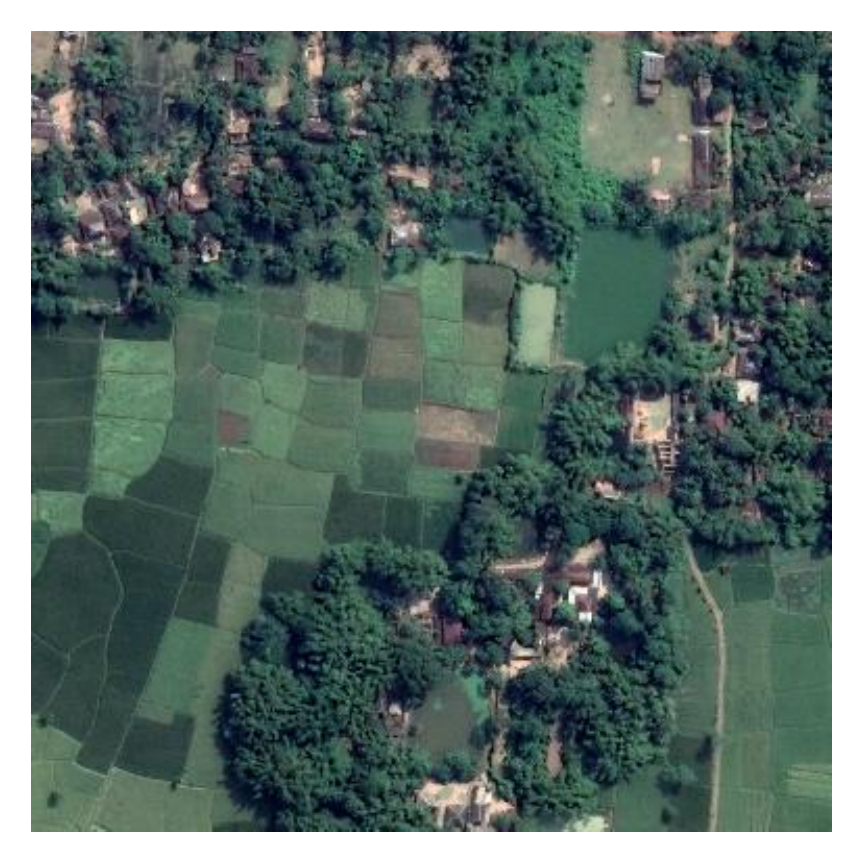}
\includegraphics[width=0.15\textwidth]{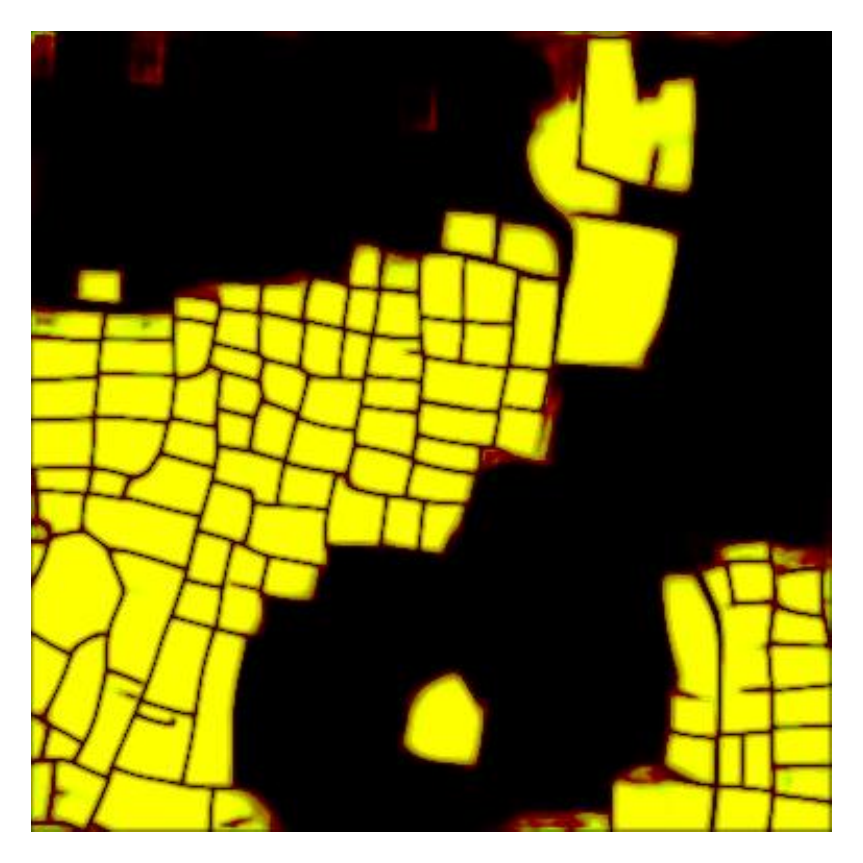}
\includegraphics[width=0.15\textwidth]{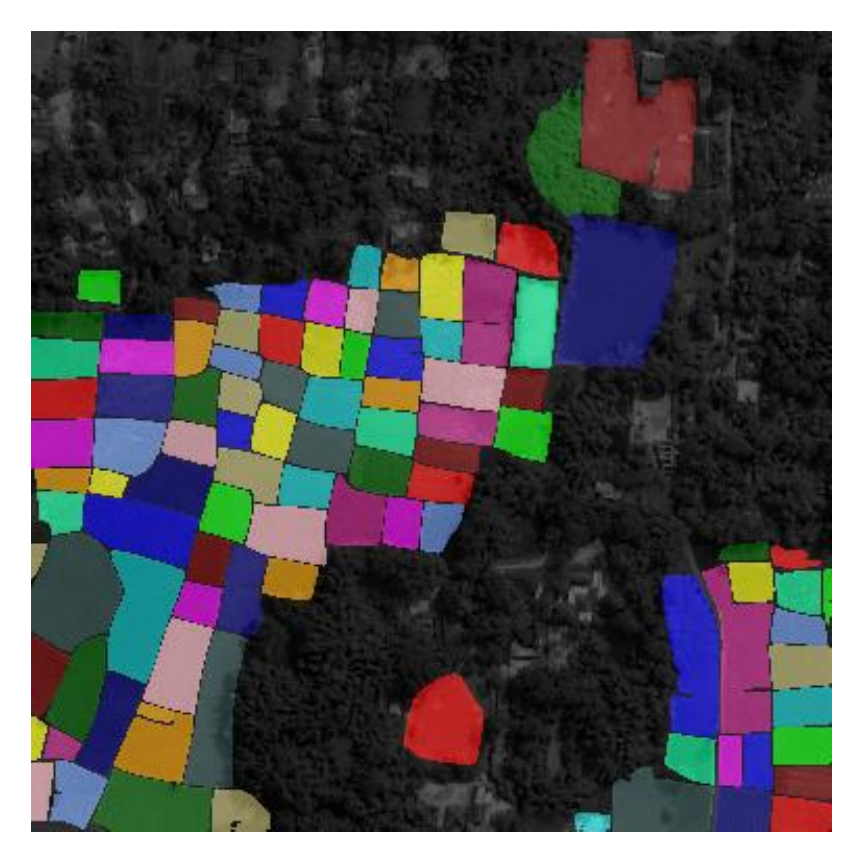}
\caption{Given an input satellite image with RGB bands (left), our goal is to generate both a semantic segmentation map (center) and an instance segmentation map (right). Outputs shown for the ground layer.}
\label{fig:three_images}
\end{figure}

The goal of the ML model is to generate semantic and instance segmentation maps for agricultural features from input satellite images (see fig. \ref{fig:three_images}).
In the following we describe how our dataset was collected, annotated and preprocessed, and define the exact semantic and instance classes.
We describe the model that had the best performance in our evaluation (shown in \S \ref{sec:mleval}) on this dataset.

\subsubsection{Data and Annotations}
\label{sec:data}

\begin{figure}[h!]
\centering
\includegraphics[width=0.5\textwidth]{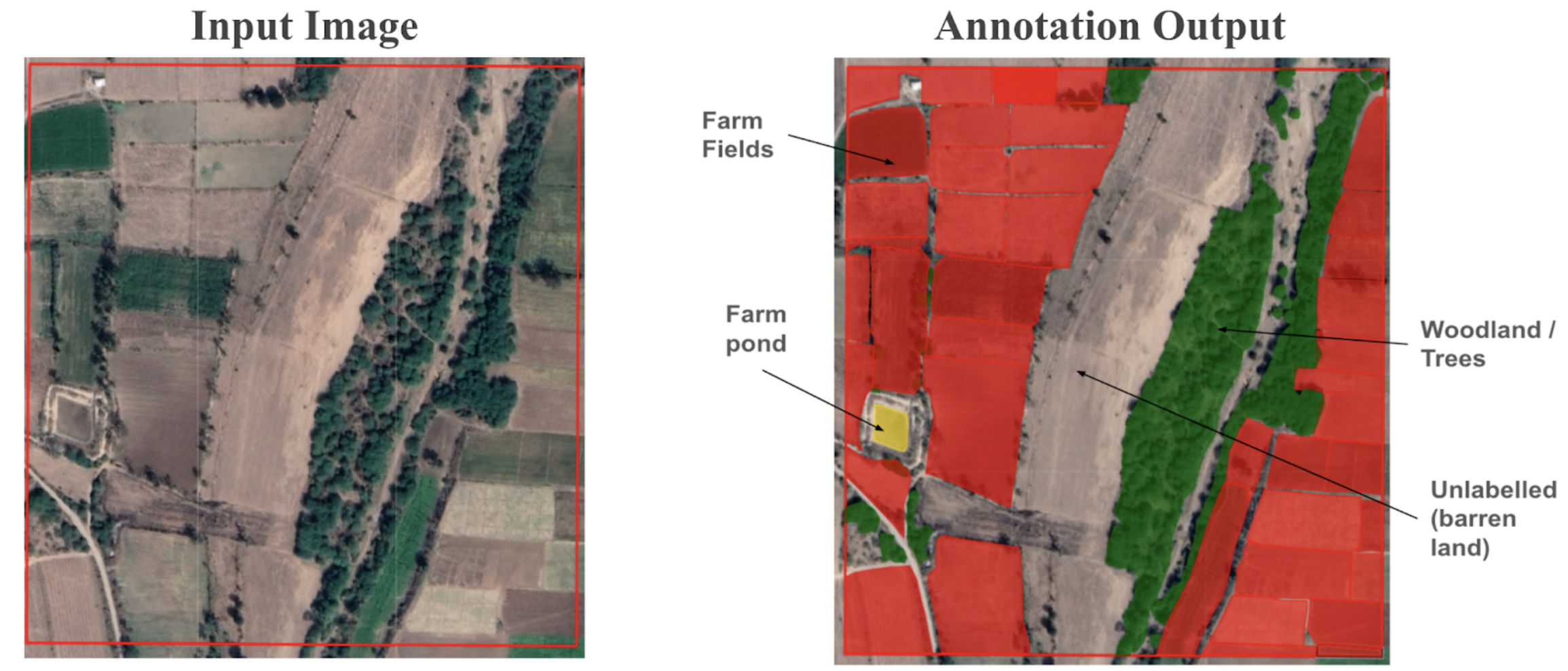}
\caption{Example: input high resolution satellite image (left); human labeled annotations (right).}
\label{fig:dataset_annotated_sample}
\end{figure}

We randomly sampled 4549 tiles from high- resolution satellite imagery obtained
from Google Maps
over India between 2019 and 2021.
Each tile comprises 1500 x 1500 pixels (corresponding to a ground sampling distance of 30 cm and a spatial extent of 450 m x 450 m).
Agricultural structures in our dataset are listed in
Table \ref{tab:dataset_layer_classes} (right column) with additional statistics in
Table \ref{tab:data_dist_table} in Appendix \ref{app:labeling}.
Annotators were hired to delineate various agricultural structures within each image --
e.g., see Figure \ref{fig:dataset_annotated_sample}.

We define four distinct layers of entities: ground, well, tree, and cloud, which conceptually represent the height of the landscape structures,
e.g., ground features like fields and ponds are at the same height and belong to the same layer.
Classes within a layer cannot overlap (e.g., ponds and fields), but classes across layers can (e.g., trees within a field).
The `tertiary' layer has two classes --  class `background' denotes pixels not belonging to the other classes,
and class `ignore' indicates masked regions excluded from the analysis.
Appendix \ref{app:labeling} provides more detail on the annotation process.
\begin{table}[h]
\centering
\caption{The defined layers and their corresponding classes.}
\begin{tabular}{ll} 
\toprule 
Layer & Classes \\
\midrule 
Ground & fields, farm ponds \& other water bodies\\ 
Well & dug wells \\
Tree & trees/woodland \\
Cloud & opaque cloud, transparent cloud \\
Tertiary  & background, ignore (non-feature layer) \\
\bottomrule 
\end{tabular}
\label{tab:dataset_layer_classes}
\end{table}

Note that this layer-wise categorization allows us to 
(i) Model overlap: predict multiple classes across layers which a pixel can belong to (e.g., tree and field, for a pixel in a tree which is within a field), 
(ii) Model physical constraints: a pixel cannot belong to two classes in the same layer, e.g., field and pond.

\subsubsection{Preprocessing}

We performed rasterization to convert the vector annotations into a raster format where each pixel is assigned a label.
To prepare the data for input to model training, we created 5 distinct layers: 1 for semantic segmentation and 4 for instance segmentation (one for each layer, except the tertiary layer).
To facilitate efficient processing and memory management during model training 
we created a downsampled version of the dataset with
resolution of 500x500 at 0.9m/pixel. 
Finally, the dataset was divided into train-validation-test splits, with
3159, 695 and 695 
images respectively. The geographical coverage of our test set is shown in Figure \ref{fig:choropleth_mediou}, Appendix \ref{app:stratified}.

\subsubsection{Machine Learning Model}

We adopt the model of Gao et al \cite{gao2019ssap} after empirical comparison with alternative techniques (\S \ref{sec:mleval}).
Our model consists of 2 parts:
(i) a U-Net based convolutional network which generates per-pixel semantic class and pixel-pair affinities in a single pass, and
(ii) a cascaded graph partitioning algorithm which uses these predictions to obtain instance segmentation.
In the following we give a brief overview; \cite{gao2019ssap} has more details.

\subsubsection{Network Architecture}

\begin{figure}[h!]
\centering
\includegraphics[width=0.5\textwidth]{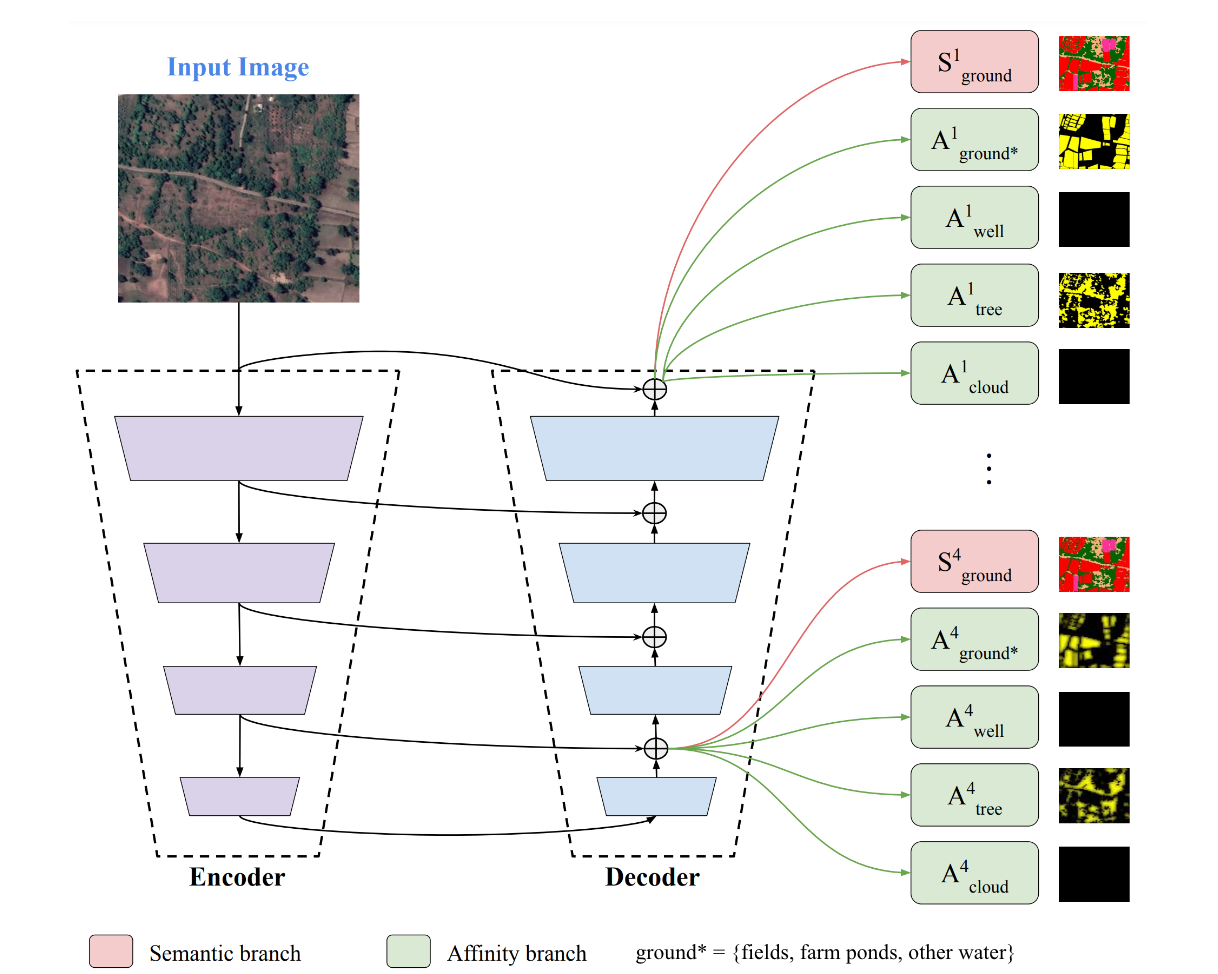}
\caption{Our model has a U-Net style encoder-decoder framework, with 5 distinct output heads: 1 semantic segmentation head and 4 affinity mask heads per upsampling stage.}
\label{fig:methodology}
\end{figure}

Our model architecture, illustrated in Figure \ref{fig:methodology}, is based on U-Net \cite{ronneberger2015u} with a ResNet50 encoder that has been pre-trained on ImageNet. 
The decoder is modified to output (i) per-pixel semantic class label
and (ii) pixel-pair affinity which specifies the probability of a pair of pixels belonging to the same instance.
Moreover, both these outputs are obtained at multiple resolution scales of 1x,  $\frac{1}{2}$x, $\frac{1}{4}$x, and $\frac{1}{8}$x.
For each pixel, affinities to neighboring pixels within an $r \times r$ window are learned.
Keeping the window size fixed, coarser long-range affinities and finer short-range affinities are learned at lower and higher resolutions respectively.
This obviates the need for varying the window sizes to learn affinities at different ranges and also makes computation tractable for subsequent graph partitioning.
This multi-scale approach is particularly useful in our application as it enables us to identify objects at vastly different levels of detail, e.g., field sizes can vary significantly
in area \cite{rs14225738}.
In our implementation, we generate one semantic segmentation map ($S_{\text{ground}}^i$) and affinity masks ($A_j^i$) for each layer $j \in $ \{ ground, well, tree, cloud \} at four different resolution scales $i$, and use $r = 5$.

\subsubsection{Instances from Affinities}
The postprocessing pipeline for instance segmentation begins with the identification of core pixels, defined as interior elements that exhibit a high average affinity within a specified neighborhood radius. These high-confidence seed pixels are subsequently grouped into initial core segments through connected component analysis. This is followed by an iterative expansion phase, where core segments grow into adjacent unassigned pixels based on affinity scores. The semantic labels are assigned to each instance by determining the argmax of the aggregated average class probability distribution.

\subsubsection{Training Objective}
We use the same combination of loss functions from \cite{gao2019ssap} (see Appendix \ref{app:loss}):
(i) $\alpha$-balanced variant of the focal loss, extended to multiple classes, and 
(ii) average L2 loss for $r^2$ affinity predictions made within an $r \times r$ affinity window for each pixel.

\subsubsection{Model Selection Criteria}

We use the mean instance-level Intersection over Union (IoU), averaged over the classes, on the validation set to choose the best checkpoint in the training process.

\subsection{Large-Scale Inference and Post-processing}
\label{sec:postprocessing}

The trained ML model inference is run on a vast satellite imagery archive where any specific location may be covered by a deep stack of images with varying resolution, age, and quality (e.g., clear, hazy, cloud-covered). 
Consequently, a single physical object (e.g., a field) may be detected with inconsistent shapes across different images may be missed entirely, incorrectly partitioned, or fused with nearby objects. Thus, post-processing is required to select the single best shape for each underlying physical object (See fig. \ref{fig:bat}). The central challenge is balancing \textit{contour quality}, derived from high-quality sensors, against \textit{freshness} from more recent imagery, as these attributes are often inversely correlated.
We employ a workflow based on BAT (Best Autogen Tracker) \cite{vector_map_patent}, a system of  heuristics that evaluate evidence across the image stack to validate an object's existence and determine its optimal shape.

\begin{figure}
\centering
\includegraphics[width=0.5\textwidth]{figures/main_paper/BAT3.png}
\caption{Left to right: Multiple input satellite images ($V_1, \ldots ,V_n$) obtained for the ground location beneath ($G$), model inference outputs ($O(V_1), \ldots ,O(V_n)$), BAT-based stitching and deduplication yields the final output ($F$).}
\label{fig:bat}
\end{figure}

\textbf{Stitching and Deduplication}
This process is enabled by an inference framework that partitions the continental-scale imagery into overlapping tiles. The overlap between tiles provides the necessary spatial context to stitch fragmented detections into complete objects. 
Detected objects that fall on the boundaries of these tiles are clustered with boundary objects in neighboring tiles.
Multiple sequential heuristics based on overlapping coverage area and image quality are used to de-duplicate the objects, and prioritize the most reliable predictions. 
Image quality is scored using the image resolution, age and a proxy measure of model performance.
The workflow is outlined in Algorithm \ref{algo:bat_workflow}, with 
complete details in Appendix \ref{sec:bat}.
\begin{algorithm}[h]
\caption{BAT Workflow for Stitching and Deduplication.}
\label{algo:bat_workflow}
\KwIn{A set of detections $D$, processed tiles $T$, and image metadata $M$}
\KwOut{A final, deduplicated set of detections $S_{final}$}
\SetAlgoLined

\textbf{1. Data Integrity and Preparation:}\\
\quad Filter $D, T, M$ to remove all entries with broken data links.\\
\quad $T_{incomplete} \leftarrow$ Identify and mark all tiles in $T$ with incomplete RGB content or a maxed-out detection count.

\textbf{2. Geographic Bucketing for Parallelization:}\\
\quad Partition $D, T, M$ into geographic buckets (e.g., using S2 cells), each with a wide margin. \\
\quad \emph{The following steps (3-6) run in parallel for each bucket.}

\textbf{3. Stitching Fragmented Detections:}\\
\quad $D_{boundary} \leftarrow$ Identify all boundary detections in the current bucket. \\
\quad $D_{stitched} \leftarrow$ Cluster detections in $D_{boundary}$ using the \emph{cluster membership criterion} and create new shapes by unioning the geometries in each cluster.

\textbf{4. Conflict Resolution:}\\
\quad $D_{complete} \leftarrow$ All non-boundary detections in the bucket. \\
\quad $S \leftarrow$ Set of post-stitched detections created after resolving overlaps between $D_{stitched}$ and $D_{complete}$ [Algorithm \ref{algo:clipstitched2}]

\textbf{5. Chronological Validation and Deduplication:}\\
\quad Compute quality scores [equation \ref{eqn:imquality}] for all images associated with the bucket. \\
\quad $M' \leftarrow$ Select the top-$n$ highest-scoring images from $M$. \\
\quad $S' \leftarrow$ Create a candidate list of detections from $S$ that originate from images in $M'$, sorted first by image quality and then by model confidence. \\
\quad $S_d \leftarrow$ Validate each detection in $S'$ against evidence from $M'$ and merge the accepted detections into a final, non-overlapping result set [Algorithm \ref{algo:finalvalidation}].

\textbf{6. Final Output Generation:}\\
\quad Filter $S_d$ to remove detections whose centroids lie outside the bucket's core area (excluding the margin).

\Return{The filtered $S_d$ from each parallel process.}\\
\end{algorithm}

\begin{figure}[h]
\centering
\includegraphics[width=0.5\textwidth]{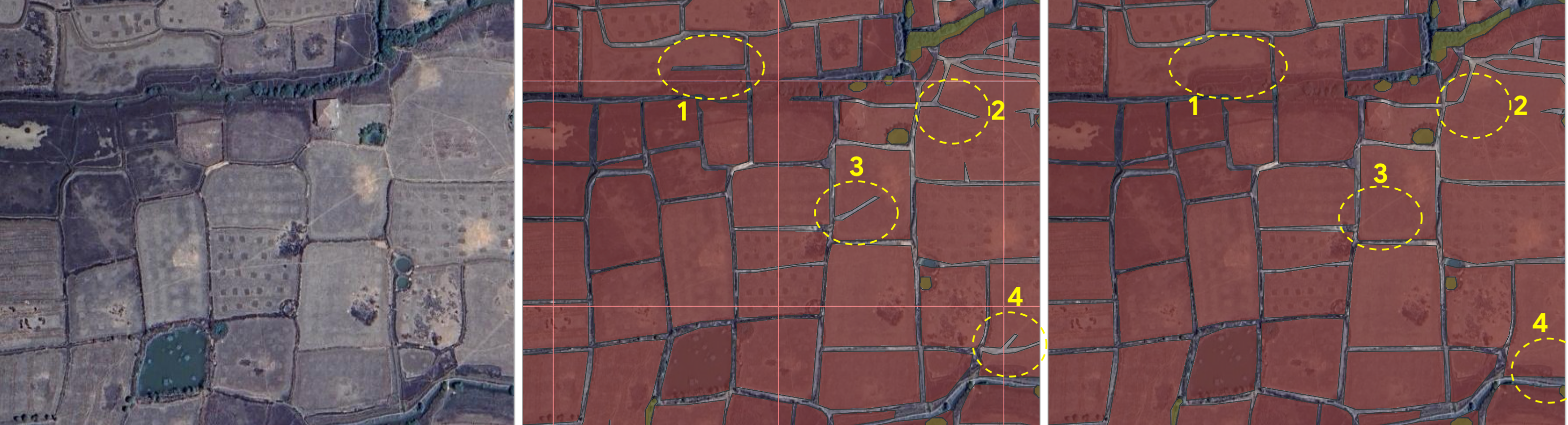}
\caption{Left to right: Input satellite image, model output, model output post processed via \textit{Dagger removal} (the sharp protrusions present in the field polygons are smoothed out).}
\label{fig:dagger_removal}
\end{figure}

\textbf{Boundary Refinement} 
We observe that the selected polygons after postprocessing exhibit irregularities or artifacts due to limitations in the model's predictions or noise in the input imagery. 
To address this, a smoothing step is performed to refine the boundaries of the polygons, resulting in a more visually appealing and accurate representation of the land-use features. This step involves removing sharp angles or ``daggers" (as indicated by the dashed polygons in Figure \ref{fig:dagger_removal}) and applying smoothing heuristics.
The key idea is to first find the base of the dagger in the convex hull of the polygon, where the base is detected by geometric rules to identify the `dagger-like' shape of the inward protrusions in the polygon, and then, close the polygon at the base by removing the points that constitute the dagger 
(more details are in Appendix \ref{sec:dagger}). 
This step yields more natural and realistic shapes.

\noindent
Running time of our inference pipeline is discussed in Appendix \ref{app:time}.

\subsection{Additional Considerations for Deployment}
\label{sec:deployment}

The following steps are performed to provide user-friendly outputs from the API.

\textbf{Vectorization} 
of pixed-based segmentation masks, which yield polygon representations of each object, provides a more compact and efficient representation of the land-use features, facilitating subsequent processing and analysis.

\textbf{Feature Identification} 
is enabled, for efficient indexing and referencing of individual agricultural features, by assigning an identifier based on the Plus Code \cite{pluscode}, comprising the latitude and longitude, of its centroid. 

\textbf{Spatial Indexing and Data Partitioning} is done at S2 cells at level 13
to facilitate efficient data storage, retrieval, and analysis. 
This partitioning strategy allows users to query and retrieve data for specific regions of interest, facilitating targeted analysis and application. 
A level 13 S2 cell covers an area of approximately 1 km x 1 km and typically requires about 1 MB of storage. This granularity offers a good balance between data volume and spatial coverage, making it suitable for sharing data as API responses while providing sufficient information for various analyses and applications. 
This sharding strategy allows users to access data for specific regions by querying with S2 cell IDs, geographic coordinates, or region names. 


\section{Pre-Launch Evaluation}

We evaluate the ALU system which includes both the ML model and the postprocessing steps, 
by comparing ALU outputs with those reported in previous works on field boundary delineation, on their respective datasets.
We also present our empirical comparison of representative ML models for this task, which led to our choice, i.e., the model of \cite{gao2019ssap}, within ALU.

\subsection{System Evaluation on external data}
\label{sec:syseval}

We consider the following works which have evaluated their techniques on 
smallholder farms in Asia and Africa, and have publicly released their datasets. 
Note that the satellite inputs, training/inference data, model and post processing steps differ across each of these works and ALU. 
Further, due to varying temporal and spatial frequencies of satellite data, both outputs may not be available for all the fields in the considered regions. 
Consequently, precise matching of test sets is not feasible, and we report
comparative performance on a subset of the data where the outputs are heuristically matched.
The details of the heuristic matching are in Appendix \ref{app:extdata}.
Therefore, these results serve merely as approximate indicators of comparative performance.

Each of the previous works use different evaluation metrics (See Appendix \ref{app:metrics}) on boundary and/or extent of the inferred fields.
As discussed in previous works, e.g., see Appendix A.2 in \cite{rs14225738}, pixel-based metrics such as accuracy or F1 score are not reliable indicators of performance for the task of field delineation, when there is class imbalance, evaluation on partial labels, post-processing for instance 
segmentation, and comparison across satellite imagery of different resolutions
(see Appendix \ref{app:extdata} for more details).
Hence, we use only the reported instance segmentation metrics in each work to assess ALU's performance. 

\subsubsection{Data and Metrics}

\begin{itemize}
\item 
\textbf{Kenya}: 
Kerner et al \cite{KernerEtAl2023Multi} report results on smallholder fields in Kenya using their transfer learning based approach where a spatio-temporal U-Net adapted from \cite{aung2020farm} is used to first train on Sentinel 2 data from France, then tuned on PlanetScope data from South Africa, evaluated on PlanetScope data on Kenya (all on visible RGB bands only).
They evaluate the instance segmentation performance through mean intersection-over-union (mIoU)
for both boundary and extent prediction of fields.
\item
\textbf{India}:
Wang et al \cite{rs14225738} also use a transfer learning based approach where a model trained with PlanetScope images in France is fine tuned on 1.5m VHR Airbus Spot images in India and used to predict field extent in India.
They evaluate their instance segmentation performance with median IoU and IoU50 (fraction of fields in the test set with IoU greater than 50\%).
\item
\textbf{Vietnam and Cambodia}:
The AI4SmallFarms dataset \cite{10278130} provides smallholder farm boundaries in Vietnam and Cambodia. Their best performing model is trained on 0.5m VHR images from Google Map.
They use the Polis score to evaluate the performance of instance segmentation.
The PoLis score \cite{avbelj2014metric} takes matched pairs of the prediction and ground truth polygons as input. The Polis distance between two polygons A and B  is the average of distances between each vertex in A and its closest point in B (not necessarily a vertex). The Polis score sums the directional distances to obtain a symmetric score. Lower score indicates better performance.

\end{itemize}







\subsubsection{Results}

Table \ref{tab:system_comparison} shows the performance of ALU on the benchmark datasets and the best reported scores in the respective works. 
The India dataset is closest in terms of the resolution of satellite data as well as geographical region on which the ML model is trained.
Unsurprisingly, our median IoU is similar to theirs with a marginally better IoU50.
ALU is able to delineate the farms well in Kenya, Vietnam and Cambodia, suggesting that our model generalizes well to other regions containing smallholder farms. 
Such generalization capabilities, even without explicit transfer learning, have also been reported in earlier works, e.g.,  \cite{mei2022using,waldner2021detect}.
Previous works have also reported higher performance from methods trained directly on VHR data, e.g.,  \cite{10278130}, and we observe the same trend here in Kenya.
We also compare the performance of ALU without the (optional) dagger removal step, denoted by ALU-DR.
We note that all the metrics in all the datasets deteriorate without this step. The effect is most apparent in the largest Vietnam and Cambodia dataset. Also see Appendix \ref{app:notebat} on system-level ablations.

\begin{table}[h!]
\centering
\small
\caption{Performance Comparison on Benchmark Datasets}
\label{tab:system_comparison}
\begin{tabular}{lccccc}
\toprule
Data 
& \multicolumn{2}{c}{\textbf{Kenya} \cite{KernerEtAl2023Multi}} 
& \multicolumn{2}{c}{\textbf{India} \cite{rs14225738}} 
& \begin{tabular}[c]{@{}c@{}} \textbf{Vietnam} \& \\ \textbf{Cambodia} \cite{10278130}\end{tabular} \\
\cmidrule(lr){2-3} \cmidrule(lr){4-5} \cmidrule(lr){6-6}
Metric & \begin{tabular}[c]{@{}c@{}}mIoU \\ (Border)\end{tabular} & \begin{tabular}[c]{@{}c@{}}mIoU \\ (Extent)\end{tabular} & medIoU & IoU50 & Polis \\
\midrule

Previous & 0.58 & 0.53 & 0.85 & 0.89 & 20.3 \\
ALU (our) & 0.614 & 0.613 & 0.844 & 0.928 & 11.89  \\
ALU-DR & 0.613 & 0.612 & 0.841 & 0.928 & 12.69 \\
\bottomrule
\end{tabular}
\end{table}

\subsection{ML model evaluation}
\label{sec:mleval}

Our choice of ML model is determined by an empirical comparison on our annotated dataset (\S \ref{sec:data}).
We compare the instance segmentation performance with two categories of baseline methods
-- field boundary delineation methods in the remote sensing literature
and panoptic segmentation methods in the computer vision literature.
In the former, deep learning techniques have been specifically developed for 
boundary delineation of fields only. 
So, we use only ground instances as defined earlier for training and evaluation.
In the latter case, we evaluate performance on all the instance layers. 
In both cases, we train all the models on our training data and evaluate on the same held-out test set.

\subsubsection{Metrics}

In addition to mean and median IoU, we also report metrics which quantify under-segmentation (US) and over-segmentation (OS), and false postive rate (FPR) and false negative rate (FNR) -- for these four metrics, lower values indicate better performance.
In addition, we report Average Precision and Recall at IoU \cite{lin2014microsoft}
Definitions are in Appendix \ref{app:metrics}.

\subsubsection{Comparison with field boundary delineation methods}
We compare our model performance on the ground instance layer with the following methods: 
ResUNet-a \cite{waldner2020deep},
DECODE \cite{waldner2021detect},
Mask-RCNN \cite{he2017mask},
SAM-2~\cite{ravi2024sam},
and
DinoV3~\cite{simeoni2025dinov3}.
Implementation details can be found in Appendix \ref{app:baselines}.
Note that we only compare the performance of their model architectures on our dataset.
Preprocessing steps and inputs differ across these works, e.g., some of these works use NIR in addition to RGB and \cite{waldner2020deep,waldner2021detect} aggregate predictions across time in different ways.


Table \ref{tab:model_comparison} shows the performance of these methods. 
The ALU model has the best mean and median IoU values with Mask-RCNN being the next best.
Average Precision and Recall at IoU are shown in Table \ref{tab:coco_metrics_fieldseg} 
and show the same trend.
We believe the improved performance could be attributed to (a)
direct instance segmentation in \cite{gao2019ssap} -- which aligns with the general trend found in the literature where joint methods outperform 2-step approaches \cite{mei2022using}, and 
(b)  auxiliary signal from other ALU classes in 
our multi-task approach.
In terms of over and under segmentation metrics they are comparable, with marginally better values for MasRCNN in OS and US.
Oversegmentation, we believe, affects downstream applications less than undersegmentation -- if a field is segmented into multiple smaller fields, then analysis of a particular area, e.g., in crop management, may not change.
However, when actual counts are utilized, both metrics are equally important.
Qualitative comparisons for some images are shown in Table \ref{tab:ground_qual} in Appendix \ref{app:qual}, where we also illustrate failure cases of ALU.







\begin{table}[htbp]
\centering
\small
\caption{Performance comparison with previous field boundary delineation methods.}
\label{tab:model_comparison}
\setlength\tabcolsep{4pt} 
\begin{tabular}{l rrrrrr}
\toprule
\textbf{Model} & \textbf{mIoU} & \textbf{medIoU} & \textbf{OS} & \textbf{FNR} & \textbf{US} & \textbf{FPR} \\
\midrule

ALU (ours) & 0.66 & 0.82 & 1.15 & 8.46 & 1.12 & 8.91 \\
MaskRCNN \cite{mei2022using} & 0.64 & 0.81 & 1.12 & 9.14 & 1.11 & 20.43 \\
DECODE \cite{waldner2021detect} & 0.56 & 0.69 & 1.19 & 2.03 & 1.24 & 22.14 \\
ResUNet-a \cite{waldner2020deep} & 0.08 & 0.04 & 1.37 & 42.90 & 1.46 & 38.85 \\
SAM2~\cite{ravi2024sam} & 0.49 & 0.46 & 1.16 & 5.83 & 1.35 & 46.04 \\
Dino V3 \cite{simeoni2025dinov3} & 0.17 & 0.11 & 1.64 & 44.48 & 1.00 & 52.47 \\

\bottomrule
\end{tabular}
\end{table}


\begin{table}[h]
    \centering
    \caption{Performance Comparison with field boundary delineation methods (COCO Metrics: AP - Average Precision, AR - Average Recall, maximum detections = 1000).}
    \begin{tabular}{l rr}
    \toprule
    \textbf{Model} & \textbf{AP@IoU=0.5:0.95} & \textbf{AR@IoU=0.5:0.95} \\
    \midrule
    ALU & 0.27 & 0.49 \\
    MaskRCNN \cite{mei2022using} & 0.26 & 0.49 \\
    DECODE \cite{waldner2021detect} & 0.10 & 0.40 \\
    ResUNet-a \cite{waldner2020deep} & 0 & 0 \\
    SAM2~\cite{ravi2024sam} & 0.03 & 0.31 \\
    Dino V3 \cite{simeoni2025dinov3} & 0 & 0.01 \\
    \bottomrule
    \end{tabular}
    \label{tab:coco_metrics_fieldseg}
\end{table}

\subsubsection{Comparison with panoptic segmentation methods}

We compare ALU with representative transformer-based panoptic segmentation methods
Mask2former \cite{cheng2022masked} and SAM2~\cite{ravi2024sam} (fine-tuned independently for each class).
Implementation details are in Appendix \ref{app:baselines}.
To extract instances for the pond layer, we extract the instances from the field layer, with predicted semantic mask as ponds.
SAM2 had the best performance on our dataset, as shown in Table \ref{tab:panopt_comparison}.
Qualitative comparisons for some images are shown in Appendix \ref{app:qual}.
We find that the simpler, CNN-based ALU model outperformed these models.
This could be due to the relatively small dataset on which the models were fine-tuned, and the distributional difference between natural images and satellite images.
Similar trends of classical CNN-based models outperforming transformer-based foundation models have been reported in recent empirical comparisons using satellite data \cite{xu2025specialized}.
We also observed that using a pretrained ALU model trained on large scale, proprietary Earth Observation data, instead of the ImageNet backend, only provides a marginal average improvement of mIoU (0.006 $\pm$ 0.02).
Techniques to improve these models, such as using SAM adapters, for satellite images remains an avenue for future work.

\begin{table}[htbp]
\centering
\caption{Performance comparison on all layers with panoptic segmentation.}
\label{tab:panopt_comparison}
\small
\setlength\tabcolsep{4pt} 

\begin{tabular}{@{}l l rrrrrr@{}}
\toprule
& \textbf{Class} &
\textbf{mIoU} &
\textbf{medIoU} &
\textbf{OS} &
\textbf{FNR} &
\textbf{US} &
\textbf{FPR} \\
\midrule
\multirow{5}{*}{\rotatebox{90}{ALU (ours)}} 
& Fields & 0.66 & 0.82 & 1.15 & 8.46 & 1.12 & 8.91 \\
& Ponds & 0.00 & 0.00 & 1.00 & 99.15 & 1.00 & 33.33 \\
& Trees  & 0.21 & 0.00 & 1.12 & 56.35 & 1.08 & 13.30 \\
& Clouds & 0.29 & 0.08 & 1.03 & 13.16 & 1.17 & 55.14 \\
& Wells  & 0.07 & 0.00 & 1.00 & 63.60 & 1.00 & 43.10 \\
\midrule
\multirow{5}{*}{\rotatebox{90}{Mask2Former}}
& Fields & 0.38 & 0.31 & 1.27 & 19.24 & 1.34 & 32.96 \\
& Ponds & 0.00 & 0.00 & 1.00 & 100 & 1.00 & 100 \\
& Trees & 0.00 & 0.00 & 1.04 & 70.31 & 1.18 & 62.13 \\
& Clouds & 0.00 & 0.00 & 1.00 & 100 & 1.00 & 100 \\
& Wells & 0.00 & 0.00 & 1.00 & 100 & 1.00 & 100 \\
\midrule
\multirow{5}{*}{\rotatebox{90}{SAM2}}
& Fields & 0.49 & 0.46 & 1.16 & 5.82 & 1.35 & 45.97 \\
& Ponds & 0.00 & 0.00 & 1.00 & 92.93 & 1.02 & 99.64 \\
& Trees & 0.21 & 0.06 & 1.25 & 21.50 & 1.13 & 58.18 \\
& Clouds & 0.15 & 0.00 & 1.50 & 55.26 & 1.04 & 99.14 \\
& Wells & 0.00 & 0.00 & 1.00 & 100 & 1.00 & 100 \\
\bottomrule
\end{tabular}
\end{table}

\subsubsection{Additional Results}
We show ALU's performance on field delineation  in India stratified by (1) agro-climatic zones, (2)
agro-ecological regions and (3) seasons, (4) S2 cells  and (5) field sizes in Appendix \ref{app:stratified}.
We discuss the zero-shot performance of our model (trained on data from India) on other countries -- Malaysia, Vietnam and Bangladesh in Appendix \ref{app:generalizabiliy}.
The performance on these countries in the Global South is comparable to that in India (Table \ref{tab:panopt_comparison}).

\section{Post-Launch Evaluation}

\subsection{Comparison with Census Data}

We obtain granular geo-spatial statistics of agricultural features at a national scale with the ALU API.
Detailed tabular data containing state-wise counts and areas of fields, dug wells, ponds and trees are shown in Appendix \ref{geospat}.
On comparing these areas with those reported in the census \cite{allindiareport} (see Appendix \ref{app:census}), we find that 
our estimates vary across the country, with several states like Arunachal Pradesh, Himachal Pradesh, Jharkhand, Kerala, Meghalaya, Mizoram, Nagaland, Sikkim and Tripura showing high agreement.


\subsection{On-ground validation}\label{onground}

We performed rigorous on ground validation of the ALU API outputs with our partners -- the Department of Agriculture of the state Government of Telangana and a local startup, TeamUp
\cite{teamupwebsite}.
During field observations in 19 pilot villages, 
errors in ALU inference were studied and reported as feedback.
In 19 villages in Telangana, we obtained a field level accuracy of \textbf{82\%}.
We observed that 2.19\% fields were under-segmented, 7.45\% fields were over-segmented for our use case. 8.18\% fields were reported to have boundary errors (due to either temporal shift in land use, or lack of clarity of instructions in reporting process).
Table \ref{tab:village_survey_results} (Appendix \ref{app:teamup}) has village-level details.

\section{Discussion}
\label{sec:discussion}

In this work, we present the development of a complete, end-to-end system for agricultural landscape understanding, comprising an integrated pipeline that combines instance segmentation with crucial post-processing heuristics designed for real-world deployment. Our approach is more holistic than prior efforts, moving beyond a narrow focus on field boundaries to comprehensively map diverse agricultural features, including trees and wells. 
To validate our system, we conducted a thorough, multi-faceted evaluation which included performance analysis on public benchmarks, comparison with in-situ surveys, and, most importantly, post-deployment on-ground validation in collaboration with external partners. 
While each of these assessments have limitations due to the complexity and heterogeneity of such systems, overall the evaluations confirm the practical utility and reliability of ALU for downstream applications.

A key outcome of this work is the first comprehensive mapping of agricultural land use in India at a national scale, achieving a granularity sufficient to support targeted interventions. We have successfully identified hundreds of millions of smallholder fields and millions of minor irrigation structures, the vast majority of which were previously unmapped and undocumented. This focus is critical, as smallholder farmers, cultivating less than two hectares of land, comprise 86.2\% of India's agricultural workforce and manage approximately half of the total cultivated area \cite{allindiareport}.
Our publicly accessible API explicitly incorporates both quality scores and inference timestamps with the segmentation outputs. This facilitates time-sensitive analysis for our API users, which is crucial for dynamic landscapes like India's, where agricultural features undergo rapid transformations due to seasonal variations, crop rotations, and land-use modifications. 
The underlying data is refreshed at 6 month cadence or earlier. 
Each API query returns the map for an S2 cell, where level 13 -- approximately 1 sq km and 1 MB in size -- is chosen to optimize for mobile apps.

Our API facilitates the development of critical downstream applications for agricultural experts, environmentalists and policy makers. For instance, ALU has been used at scale by our partners: 
(1) Ministry of Agriculture, India, for real-time, parcel-level intelligence to replace outdated district-level approximations in their AgriStack~\cite{agristack} and Krishi-DSS~\cite{krishidss} Agricultural Decision Support System platforms, (2) Government Water Resource Department (e.g., ACIWRM in Karnataka, India~\cite{kwris}) to measure crop water productivity and irrigation efficiency at the individual parcel level, identifying vulnerable farming communities, (3) CarbonFarm~\cite{carbonfarm}, for accurately delineating rice paddies to measure wet/dry cycles and issue carbon credits for methane reduction and (4) TerraStack~\cite{terrastack}, for creating spatial identities that unlock formal credit and insurance for underserved farmers.

\subsubsection{Limitations}
While our results suggest that ALU generalizes well, with promising evaluations in Kenya, Cambodia, Malaysia, Vietnam and Bangladesh  we acknowledge that a more thorough evaluation across a wider range of diverse geographies is required to fully establish its robustness. 
From a modeling perspective, the detection accuracy for smaller, less frequent features like ponds and wells is not as high as for larger features. This can be addressed through future algorithmic improvements such as techniques designed to handle class imbalance, as well as by additional data collection. 
Further, our system is currently limited to just four classes; future work should expand this vocabulary to include other agriculturally significant features like rivers, canals, and warehouses. 
From a data perspective, our system currently relies on very-high-resolution imagery, which is typically available only once every one to two years. To capture the dynamic nature of agriculture, future systems must leverage low-resolution, high-revisit-frequency satellite imagery to identify in-season changes. 
Also, our current model exclusively uses RGB channels; future work can explore the integration of other spectral bands.
Our current BAT-based post-processing pipeline is designed to resolve spatial conflicts and select the single best shape for each object at a given timepoint. The chronological validation and deduplication (step 5) operates at timescales that are much smaller than the scales at which agricultural temporal dynamics is monitored (typically months). However, we expect longer-term temporal changes to be estimated with these ALU outputs. Alternative approaches are possible where temporal estimates are obtained over long periods of time, directly from multiple satellite inputs, with BAT being an intrinsic part of the estimation process itself -- this would be a novel and impactful direction of future work.

\section{Conclusion}

This work presents a novel, rigorously evaluated system that delivers the first national-scale, multi-class map of agricultural landscapes with a unique focus on smallholder agriculture.
 Our method is distinguished not only by its holistic inclusion of fields, trees, and water bodies but also by its practical design, which incorporates essential post-processing steps for real-world usability.
Detailed maps obtained from our system are made publicly accessible through an API at \url{http://agri.withgoogle.com}. 
We believe ALU can empower a wide range of applications in this socially impactful domain.

\begin{acks}
We thank
Musty Krishna Chaithanya, 
Anumas Ranjith Reddy from TeamUp for their on-ground validation of ALU outputs in Telangana;
Aashish Kumar and Betala Laxmi Tirumala from the Government of Telangana, for their assistance in coordination and validation in Telangana;
Muqthar Mohammad and 
Kiranmai Chennuru for 
coordinating the data annotation;
Amandeep Kaur,
Nikita Saxena,
Ujwal Singh,
Jitendra Jalwaniya,
Gaurav Singh,
Chandan Nath, 
Arnab Basu,
Nishi Shah, Avneet Singh, Dinesh Tewari, 
Agata Dondzik, KC Chung, 
Sharath Holla, Bindiya Kurle, Olana Missura, Rahul Aggarwal, Kshitij Pancholi and Shubhika Garg
for their assistance in data collection, 
software implementation and visualizations.
\end{acks}

\bibliographystyle{ACM-Reference-Format}
\bibliography{sample-base}

\appendix
\section{Dataset Statistics}
\label{app:datasetstats}

Agricultural structures in our dataset are listed in Table \ref{tab:data_dist_table}. Our dataset is split into train, validation and test splits comprising 3159, 695 and 695 images respectively. 
The table depicts the specific number of instances spread out over a set of images, for each class in each split in the dataset. For example, the train data contains 150273 field instances, spread over 2467 of the 3159 images available for training.

\begin{table}[!h]
\centering
\caption{
Number of annotated instances, and number of images in which they occur, for each class, in our dataset splits.}
\label{tab:data_dist_table}

\small
\setlength{\tabcolsep}{3pt}
\begin{tabular}{lrrrrrr}
    \toprule 
    & \multicolumn{2}{c}{\textbf{Train}} & \multicolumn{2}{c}{\textbf{Validation}} & \multicolumn{2}{c}{\textbf{Test}} \\
    \textbf{Features} & Instances & Images & Instances & Images  & Instances & Images \\
    \midrule
    Fields & 150273 & 2467 & 37346 & 556 & 35663 & 566\\
    Ponds  & 1269   & 505  & 281   & 114 & 257   & 112 \\
    Trees  & 292002 & 2774 & 67263 & 621 & 65059 & 617 \\
    Wells  & 2380   & 807  & 452   & 168 & 571   & 180\\
    Clouds & 267    & 109  & 42    & 18 & 74    & 27 \\
    \bottomrule
\end{tabular}



\end{table}

\section{ALU: Time Complexity}
\label{app:time}
We measured the time taken by our inference pipeline which includes all our postprocessing steps and is being served through the publicly available API, on an area of 89,000 sq km over a 4 year period comprising 400,000 images (resolution 1536 x 1536 px). This yields an average inference time of 0.0315 seconds/image and 0.14 seconds/sqkm.

We separately evaluated the inference time of the ML model (without postprocessing steps) on a V100 GPU on 512 x 512 px images with batch size of 16. The average time taken is 26.68 ms.

Since the maps are computed at a frequency of roughly 6 months and stored, latency due to inference is not relevant to the API user.
\section{Comparison with External Benchmark Datasets}
\label{app:extdata}

\subsection{Kenya}

Each polygon in the input geojson is associated with a survey date. We also determine the S2 level 8 and level 13 cells that the polygon is contained in. We consider the ALU output with the closest available date (on or before) the survey date, for the same S2 cell.
Total number of polygons compared: 284.

\subsection{India}

In their paper, train/val/test splits are created by dividing India into 20 x 20 grid cells as shown in the figure 2 in \cite{rs14225738} . They assigned 64\% of grid cells to the training set, 16\% to the validation set, and 20\% to the test set. All images that fell into a grid cell were assigned to the corresponding dataset split. Input images are not shared publicly. Latitude and longitude coordinates of the grid cells and the split they belong to were shared (via personal communication) by Sherrie Wang. 

To find the field instances that belong to the test grid, we employ the following heuristic. For all the grid coordinates (lat, long pairs), we obtain their enclosing s2 level 8 cell and the 8 neighboring cells that share an edge or vertex with the enclosing cell. We consider all the input field instances belonging to these s2 cells as the test set. Visual inspection confirms that the fields selected in this manner  roughly correspond to the fields in the test set.

For each polygon in the test set, we search for an ALU polygon in the same level 13 S2-cell, and consider the one which has maximum intersection in area. If no intersecting polygons are found, then the closest ALU polygon is considered.
Total number of polygons compared: 4703.

\subsection{Vietnam and Cambodia}

For each input polygon in the provided files, we determine the S2 level 8 and level 13 cells it is contained in and find the ALU outputs corresponding to the S2 cells, between July and October 2021.
Among these ALU outputs, the polygon which has maximum intersection with the input polygon is considered as the predicted ALU polygon.
Total number of polygons compared: 1,702 in Vietnam and 11,573 in Cambodia, for a total of 13,275 polygons.

\section{Training Loss Definitions}
\label{app:loss}

\textbf{Semantic segmentation} 
To address the problem of foreground background imbalance, we use the Cross-Entropy Focal Loss \cite{lin2017focal} to train the
semantic segmentation map at each scale.
We use the $\alpha$-balanced variant of the focal loss, extended to multiple classes. 
For the case of two classes, focal loss is
given by $FL(p_t) = -\alpha_t (1-p_t)^{\gamma} \log{p_t}$,  where $p_t = \{p \text{ if } y=1, \text{ and } 1-p$ otherwise\}; $y \in \{\pm 1\}$ specifies the ground-truth class and $p \in [0,1]$ is the model's estimated probability for label $y = 1$ (computed through a sigmoid activation in the network).
Note that the cross entropy loss, $CE(p_t) = -\log{p_t}$.
Here, $\gamma \ge 0$ is the tunable focussing parameter, which enables the scaling factor to decay to zero as confidence in the correct class increases, and $\alpha_t \in [0,1]$ is a hyperparameter.


\textbf{Affinity prediction} 
For each pixel, $r^2$ affinity predictions are made within an $r \times r$ affinity window. The average L2 loss is used during training: $\frac{1}{r^2}\sum_{j=1}^{r^2}(z_j - a_j)^2$, where $a_j$ is the predicted affinity between the current pixel and the $j^{\rm th}$ pixel in its affinity window, and $z_j$ is the ground-truth affinity set to 1 for pixel pairs belonging to the same instance and to 0 otherwise.
Sigmoid activation is used to obtain $a_j \in (0,1)$.
The loss is computed across affinity masks at all resolutions. 
We add additional weights to the edge pixels to improve learning for better boundary delineation.

\section{Baselines}
\label{app:baselines}

\subsection{Field Boundary Delineation Methods}

\begin{itemize}
\item 
ResUNet-a \cite{waldner2020deep}: a multi-task UNet-based CNN which predicts extent, boundary and distance to nearest boundary, and then applies watershed to delineate field instances.
We implemented the model based on the original paper, and tuned hyperparameters like learning rate and batch size on our dataset. The number of epochs was set to 100 and trained on a V100 GPU.
\item
DECODE \cite{waldner2021detect}: a network called Fractal ResUNet which uses FracTAL ResNet building blocks in the same multi-task backbone as ResUNet-a, and obtains instances through a hierarchical watershed algorithm.
We implemented DECODE as in the original paper, with hyperparameter tuning done on learning rate, fractal depth, number of fractal units, normalization layers, with our dataset. The model was trained on a batch size of 32 for 100 epochs.
\item
Mask-RCNN \cite{he2017mask}: which follows the implementation described in \cite{mei2022using} for smallholder field delineation. 
\item 
SAM-2~\cite{ravi2024sam}: which uses a promptable mask decoder from SAM that generates a segmentation mask based on image embeddings and prompts.
\item  
DinoV3~\cite{simeoni2025dinov3}: 
We utilise the ViT backbone pretrained on satellite dataset (SAT-493M) and their semantic segmentation probe over the backbone. Our approach involves unfreezing the DINOv3 ViT backbone weights and training the entire model (backbone + lightweight segmentation head) jointly on ALU’s training dataset specifically for  fields only. This is followed by a watershed algorithm post-processing step to generate precise field instance segmentation. Finetuning is done for 20,000 iterations on a batch size of 16, at which point we saw training loss saturation and evaluation loss degradation.
\end{itemize}

\subsection{Panoptic Segmentation Methods}

All the baseline models were initialized with ResNet-50 checkpoints trained on the COCO dataset.
\begin{itemize}
    \item 
    Mask2Former~\cite{cheng2022masked}: consists of a low resolution pixel encoder model, a pixel decoder and a modified transformer decoder that makes use of masked attention instead of cross attention. 
    \item 
    SAM2~\cite{ravi2024sam}: SAM2 is a promptable video and image instance segmentation model. To derive panoptic segmentation results, we use a class-specific instance segmentation approach by fine-tuning the SAM2-L model independently for each class. To generate class-targeted prompts, an auxiliary binary semantic segmentation model is trained using binary cross-entropy loss. During both training and inference, point prompts are sampled from the output masks of this semantic model, effectively guiding the SAM2 instance segmentation process toward the intended class. This is a common technique utilized to generate target panoptic outputs from the SAM family of models. During finetuning of the SAM2 model, each class-specific model is trained for 1,700 epochs with a batch size of 8.
\end{itemize}

\section{Evaluation Metrics}
\label{app:metrics}

\subsection{Metrics used for ALU System level evaluation}





Pixel-wise metrics that incorporate True Negatives such as Accuracy and Matthews correlation coefficient MCC (which measures the correlation between predictions and true labels) exhibit limited reliability, as their values can fluctuate significantly based on the resolution and boundary width of the generated segmentation masks. This variability stems from the nature of the data and the methodology employed for comparisons. 

The ground truth test set comprises a predefined collection of fields (polygons) within a specific geographical area. Typically, not all fields within a given region are included in the test set (e.g., in Kenya and India). When data consisting of such field polygons for a fixed region is rasterized, every pixel located outside the field (boundary/extent, depending on the evaluation) is considered as a ``negative,'' even though in reality, certain pixels may belong to a field (which are not included in the test set). Increasing the resolution of the rasterized image during this process could lead to a higher increase in the number of such ``negative'' pixels than the relatively fewer ``positive'' pixels. In a field-by-field comparison, especially when only boundaries are considered, this results in a substantial count of true negatives, which can artificially inflate the values of accuracy and MCC.

Conversely, pixel-wise metrics that do not utilize True Negatives (Precision, Recall, and F1) consider only true positives, false positives, and false negatives. While generally more reliable, these metrics are not entirely foolproof. ALU outputs may contain ``true'' fields absent from the ground truth in a region-wise comparison. In a field-by-field comparison, wherein the ALU field with the maximum intersection with the ground truth is selected, inaccuracies may arise due to over- and under-segmentation over time. This can occur even when all fields are considered (e.g., for Cambodia and Vietnam), as the same fields are not guaranteed to be present during inference from the baseline model and ALU, given differences in satellites used and their capture times. Instance segmentation metrics (IoU, IoUk) and polygon shape metrics (Polis), are more reliable compared to Accuracy and MCC, but are also subject to these limitations.

\subsection{Metrics used for ALU ML model evaluation}



We evaluate our model on our labeled data with the following metrics. The main difference from previous works is the way in which predicted instances are matched to ground truth instances.

\begin{enumerate}
    \item \textbf{IoU (Intersection-over-Union)} 
    Our per-instance IoU is defined as follows, $IoU_I = (P_{m}^i \cap P^i)/(P_{m}^i \cup P^i)$, where 
    $P_{m}^i$ is the total number of pixels in all the predicted $i^{\rm th}$ merged and matched instances (see below) which have more than $t \%$ overlap with the  $i^{\rm th}$ ground truth instance and $P^i$ is the total number of pixels in the ground truth $i^{\rm th}$ instance. 
    
    A predicted instance is considered to match with a ground truth instance if they both belong to the same class AND the overlap covers at least $t=10\%$ of the ground truth instance. All predicted instances that match the same ground truth instance are merged together, and predicted instances that do not match any ground truth instance are ignored. We then compute the IoU of the merged predictions w.r.t the matching ground truth instances. Mean and median are computed over all ground truth instances per class. 
    
    We allow multiple predictions to match with a ground truth because, for agricultural landscape understanding we find that it does not hurt any downstream applications. E.g., if one field is segmented into multiple smaller fields then any analysis of a particular area of interest remains unaffected.

    \item \textbf{False Negative and False Positive Rates} FNR is defined as the fraction of ground truth instances that do not have any matching predicted instance (match defined in the same way as in IoU). FPR is the fraction of predicted instances that are not matched with a ground truth instance.

    \item \textbf{Over segmentation and Under segmentation metrics}
    Over-segmentation occurs when objects are subdivided, resulting in fragmented detections. Under-segmentation occurs when objects are merged or not adequately separated.
  
    The \textbf{OS} metric quantifies the degree of over-segmentation by a model, i.e., how frequently it breaks up one ground truth instance across multiple predicted instances. 
    We compute it as a ratio of the total predicted instance count for each ground truth instance to the number of ground truth instances that have at least one corresponding predicted instance: $$OS = \frac{\sum_{i=1}^{N} \mathrm{PRED\_inst\_count\_per\_GT\_inst}_i}{N_{GT}^{pred>0}}$$

where 
$\mathrm{PRED\_inst\_count\_per\_GT\_inst}_i$ is the number of unique overlapping inferred instances per $i\text{-}th$ ground truth instance,
and 
$N_{GT}^{pred>0}$ is the number of ground truth instances that have at least one corresponding predicted instance.
Note that if $\mathrm{PRED\_inst\_count\_per\_GT\_inst}_i > 1$, then the instance is over-segmented.

Similarly, the \textbf{US} metric quantifies the degree of under-segmentation by a model, i.e. how frequently does a model cover up multiple ground truth instances with a single predicted instance. It is defined as: 
$$ US = \frac{\sum_{i=1}^{N} \mathrm{GT\_inst\_count\_per\_PRED\_inst}_i}
{N_{pred}^{GT>0}}
$$

where 
$\mathrm{GT\_inst\_count\_per\_PRED\_inst}_i$ i is the number of unique overlapping ground truth instances per $i\text{-}th$ predicted instance.
$N_{pred}^{GT>0}$ is the total number of predicted instances for which at least one GT instance is present.
An instance is under-segmented if  $\mathrm{GT\_inst\_count\_per\_PRED\_inst}_i  > 1$. 

\end{enumerate}

In addition to the metrics above, we report the following metrics based on the COCO evaluation framework~\cite{lin2014microsoft}
\begin{itemize}
\item 
\textbf{Average Precision and Recall at IoU (AP/AR@IoU)}: We calculate average precision (AP) and average recall (AR) at IoU threshold values $t$, ranging from 0.5 to 0.95 in increments of 0.05. Instances with IoU greater than threshold $t$ are treated as true positives.
$$AP_{0.5:0.95} = \frac{1}{10}\Sigma_{t\in\{0.5, 0.55, ..., 0.95\}}AP_{t}$$
$$AR_{0.5:0.95} = \frac{1}{10}\Sigma_{t\in\{0.5, 0.55, ..., 0.95\}}AR_{t}$$
where $AP_t = \frac{1}{11}\Sigma_{s\in\{0, 0.1, ..., 1.0\}}P_{s}$ ($P_s$ is the maximum precision for recall $\ge\ s$ and $t$ is the threshold and the fraction 1/11 results from the 11 equidistant bins of width 0.1 between 0 and 1) and $AR_{t}$ is the maximum recall at a threshold $t$ for a specific number of detections in an image (1000 detections in our case). 
\end{itemize}
\section{Labelling Process}
\label{app:labeling}

Figures \ref{fig:annotation_guide1} and \ref{fig:annotation_guide2} illustrate the annotation process with example input images and their corresponding annotation output. 
As shown in the figure, annotators meticulously outlined the boundaries of different features, such as farm fields, farm ponds, and woodlands, ensuring accurate representation of the diverse elements within the agricultural landscape. This detailed annotation process captures both the semantic categories and the spatial extent of individual instances, providing the necessary information for training and evaluating our panoptic segmentation model.


\subsection{Annotation Guidelines}\label{annotation}
\begin{figure*}[h]
    \centering
    \includegraphics[width=0.8\textwidth]{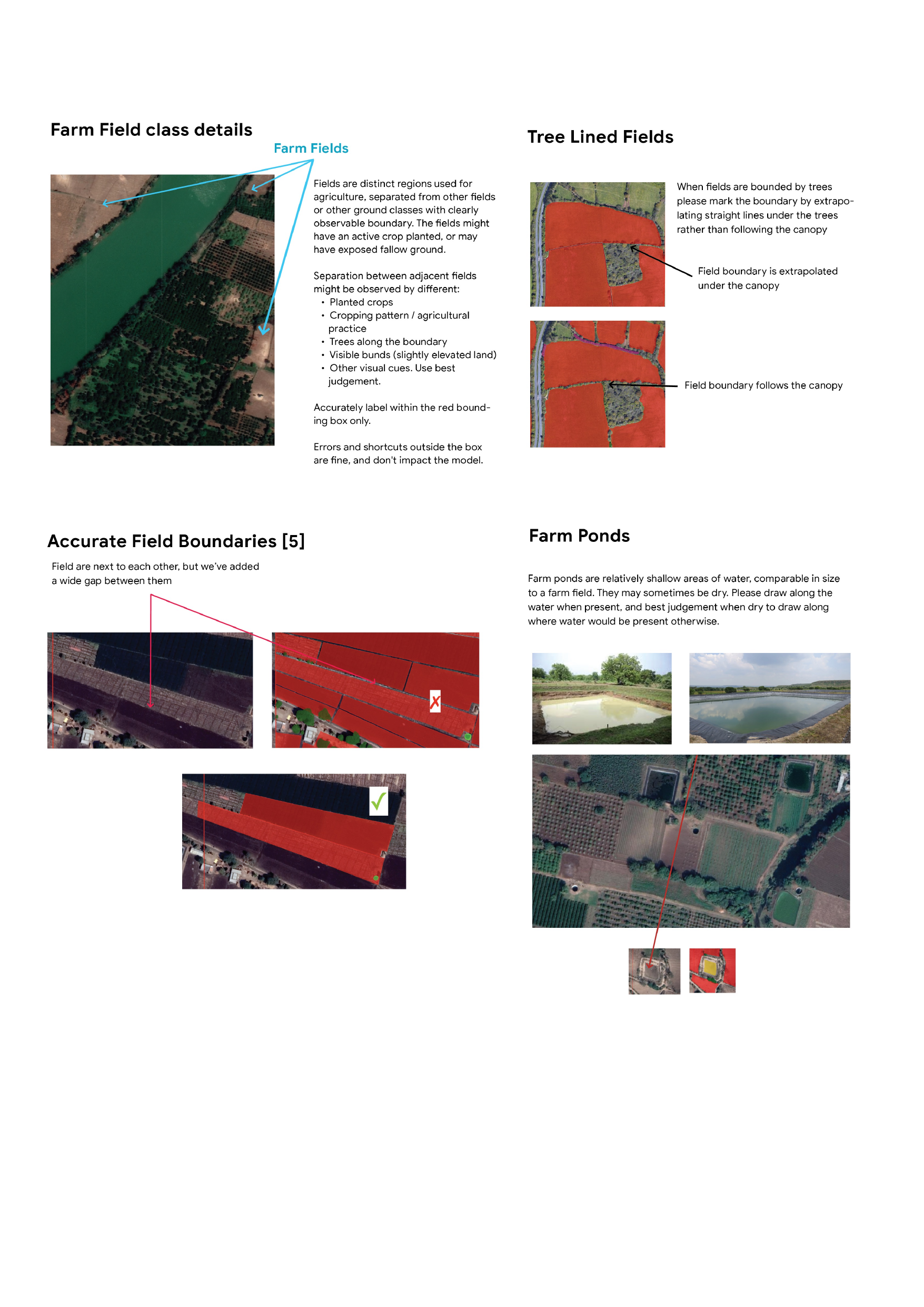}
    \caption{Annotation Guidelines for ground instance classes: fields and ponds}
    \label{fig:annotation_guide1}
\end{figure*}

\begin{figure*}[h]
    \centering
    \includegraphics[width=0.8\textwidth]{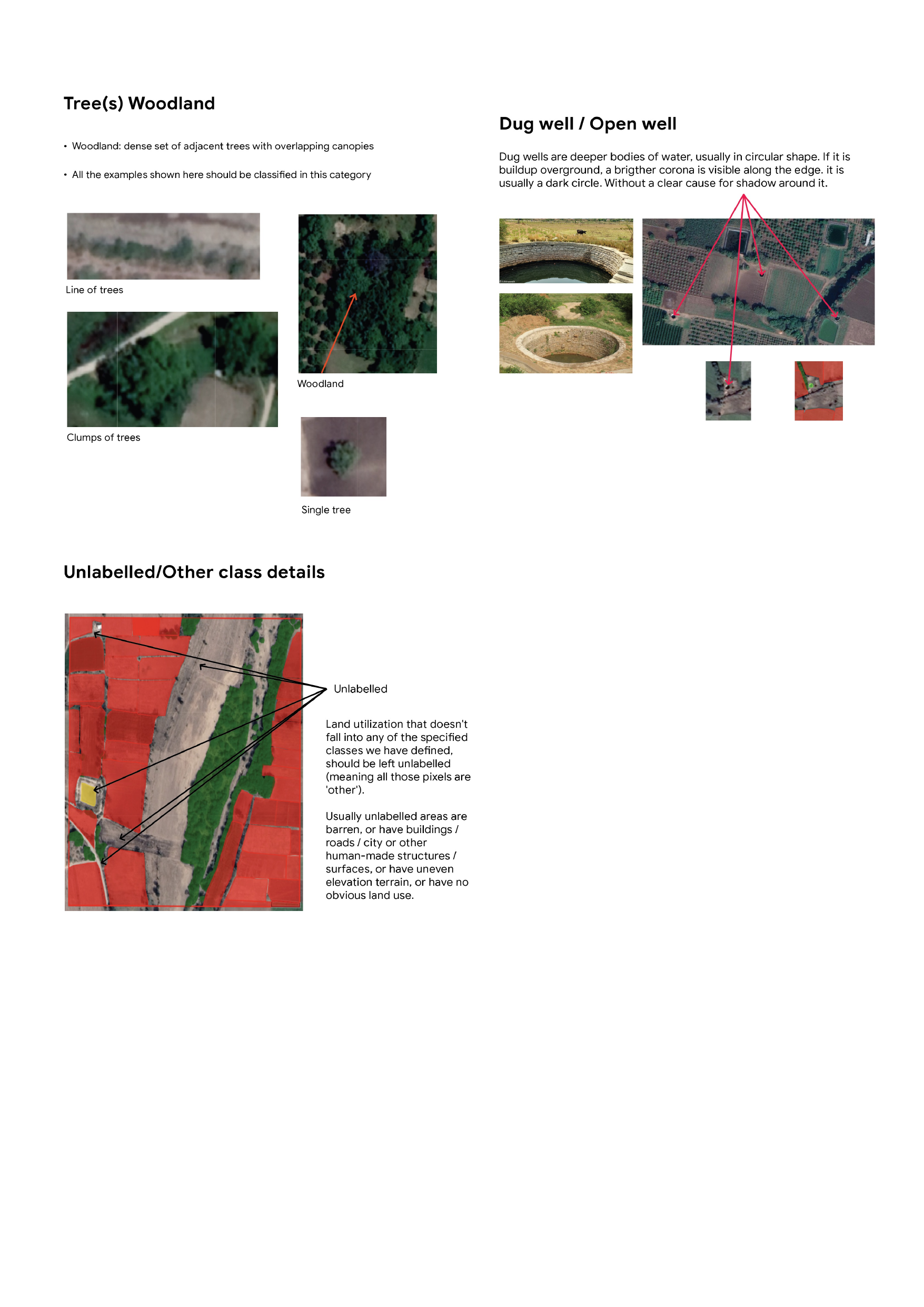}
    \caption{Annotation Guidelines (contd.) for trees, wells and other classes}
    \label{fig:annotation_guide2}
\end{figure*}

To facilitate accurate and consistent labeling of agricultural features by annotators, we established a comprehensive annotation protocol (see Figures \ref{fig:annotation_guide1} and \ref{fig:annotation_guide2}). Recognizing the variability in class definitions across different implementations, we aimed to create a standardized benchmark for the following classes we annotated. These guidelines were validated by agricultural experts and partners.

For ambiguous cases, such as differentiating barren farm ponds from fields, we provided detailed morphological criteria. For instance, a `farm pond' was defined as exhibiting clear depth along its edges, even when dry. Similarly, `wells' were distinguished from `tree shadows' based on their regular, often circular shape. Annotators were instructed to encompass entire clusters when labeling features like `trees,' rather than individual instances, to capture the spatial extent of these features accurately. Additionally, we permitted overlapping polygons to reflect the potential overlap of ground, tree, and well layers in the real world.


To ensure annotation quality, we ran pilots to train and calibrate annotators. 
Each image was annotated by 1 annotator, to avoid reconciliation issues across multiple annotators. Average time taken was $\sim$90 mins/image at a rate of $\sim$10-15 USD/hour.



\section{De-Duplication: Stitching And Selection}\label{sec:bat}

Our model processes a vast satellite imagery archive where any location is covered by a deep stack of images with varying resolution, age, and quality (e.g., clear, hazy, cloud-covered). This heterogeneity generates significant duplication and noise in the raw model output. Consequently, a single physical object may be detected with inconsistent shapes across different images or not at all. A missed detection can indicate the object's absence at the time of capture, or it may result from poor image quality or other model confusion. Furthermore, the model can erroneously partition a single object or fuse it with nearby features.
To resolve these issues, post-processing is required to select the single best shape for each underlying physical object. The central challenge is balancing \textbf{contour quality}, derived from high-quality sensors, against \textbf{freshness} from more recent imagery, as these attributes are often inversely correlated. We employ BAT (Best Autogen Tracker) \cite{vector_map_patent} for this task —- a system of heuristics that evaluates evidence across the image stack to validate an object's existence and determine its optimal shape, as outlined in the following subsections. While originally developed to de-duplicate building detections from aerial imagery, BAT's core logic is general enough to be adapted, with configuration tuning, to satellite imagery and other polygonal object classes.

\subsection*{The Tiled Inference Framework}

Processing continental-scale imagery necessitates partitioning it into manageable chunks. The framework invokes the model on \textbf{tiles}: N x N pixel squares defined on a Universal Transverse Mercator (UTM) projection grid. Typically, N = 1024. Each tile consists of a unique 924x924 pixel core area, surrounded by a 50-pixel margin on each side.

\begin{figure}[h]
    \centering
    \includegraphics[width=0.25\textwidth]{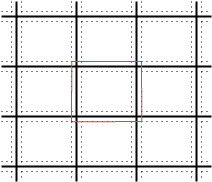}
    \caption{Imagery tiles with margin in a UTM grid. Solid lines form the grid of ``proper'' non-overlapping 924x924 tiles, while the dotted lines show how far each tile extends when we account for the 50 pixel margin assigned to it.}
    \label{fig:tileswithmargin2}
\end{figure}

As shown in Figure \ref{fig:tileswithmargin2}, the core areas of adjacent tiles form a seamless, non-overlapping grid (solid lines). However, each tile's margin extends into the core areas of its neighbors (dashed lines). This intentional overlap is fundamental to the system, as it provides the necessary spatial context for subsequent cross-tile analysis and stitching —- the process of connecting detections made separately in different tiles into complete representations of the underlying physical features.

\subsection*{The BAT Workflow}

BAT operates on three interlinked data entities: \textbf{detections (D)} inferred from the ML model, the \textbf{tiles (T)} they derive from, and \textbf{image metadata (M)}.
The BAT workflow, described below, is structured as a sequence of steps designed to systematically refine the raw, tiled model output into a clean, de-duplicated set of polygons.

\begin{figure}[]
    \centering
    \includegraphics[width=0.5\textwidth]{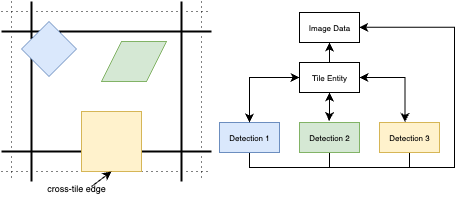}
    \caption{Left: An illustration of two complete shapes (green and blue) and one boundary (yellow) shape within the same tile with margin (note that we show only one tile with its margin here, without showing the boundaries of the neighboring tiles' margins as was shown in the previous figure). The yellow shape has a single cross-tile shape and is thus a ``boundary'' shape. Right: Data representation used in BAT with tiles and detections mutually linked and both data types pointing to the metadata record for the image used during inference.}
    \label{fig:detectionshapes2}
\end{figure}

\subsubsection*{1. Data Integrity and Preparation}

The process begins with a critical data preparation phase. 

\begin{itemize}
\item 
Integrity Filtering: The system first validates all links between the three input data entities. Any records with broken or missing links—which can result from asynchronous data ingestion, system failures, or the removal of source imagery—are filtered out. This ensures that all subsequent operations are performed on a complete and consistent dataset.
\item 
Marking Incomplete Tiles: Certain tiles are flagged as providing unreliable evidence and are added to a set $T_{incomplete}$. This includes:
\begin{itemize}
\item 
RGB Incompleteness: Tiles near an image edge may contain dummy pixel values. These are marked to prevent them from being used as false negative evidence.
\item
Model Output Incompleteness: Tiles containing the maximum number of detections allowed by the model's configuration. The absence of a detection here is ambiguous, as more objects may have been present than could be reported.
These marked tiles are not discarded but are excluded from the validation score calculation in a later step.

\end{itemize}

\end{itemize}

\subsubsection*{2. Geographic Bucketing for Parallelization}

To enable massively parallel processing, the global dataset is partitioned into geographic ``buckets'' typically defined by S2 cells (e.g., level 11: about 16 sq km). Each bucket is augmented with a substantial margin (e.g., 500 meters) to ensure that objects crossing bucket boundaries can be fully reconstructed. The subsequent steps of BAT are then executed independently within each bucket.

\subsubsection*{3. Stitching Fragmented Detections}

\begin{figure}[]
    \centering
    \includegraphics[width=0.45\textwidth]{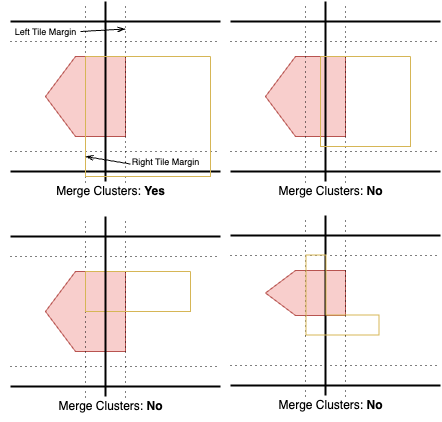}
    \caption{Top left shows a case where the membership criterion is met and the detections lock onto each other. Top right shows a case where no merging is done because the detection with the yellow outline shape does not have a cross-tile edge (it does not reach all the way to the margin of the right tile), is thus not a boundary shape and is not used in clustering. Bottom left and right show two cases where both shapes have cross-tile edges but there is no mutual containment of at least one such edge of each shape by the other. On the bottom left, only about half of the red's cross-tile edge is inside the yellow shape. On the bottom right, red's cross-tile edge is not contained by the yellow shape, nor are either of the two cross-tile edges of the yellow shape contained by the red shape.}
    \label{fig:mergingclusters2}
\end{figure}

The tiled inference approach often fragments large objects across multiple tiles. This step reconnects those fragments.
\begin{itemize}
    \item 
Identify Boundary Shapes: Detections are categorized as either ``complete'' or ``boundary''. A boundary shape has at least one ``cross-tile edge'' that is flush (or nearly so) with its tile's margin, indicating the object likely continues into a neighboring tile. See figure \ref{fig:detectionshapes2}.
\item
Cluster Fragments: Only boundary shapes are subject to clustering. The process uses a strict cluster membership criterion: two clusters are merged only if the union polygon of each contains a cross-tile edge from the other. This ``docking'' logic is far more robust than simple overlap, preventing the erroneous fusion of distinct but proximate objects. See figure \ref{fig:mergingclusters2}.
\item
Create Stitched Detections: For each resulting cluster, the constituent fragments are replaced by a single new stitched detection, whose geometry is the geometric union of all fragments.

\end{itemize}

\subsubsection*{4. Conflict Resolution}

After stitching, conflicts can arise between new stitched shapes and the original complete shapes. These are resolved via the heuristic in Algorithm \ref{algo:clipstitched2}, which generally prioritizes complete detections:

\begin{algorithm}[h]
\SetAlgoLined

  \For{\text{all stitched detections} $S$}{

    \For{\text{all complete detections} $C$}{

       \If{$C$ \text{overlaps} $S$ \text{by more than X\% of } $Area(S)$}{
         {\text{Drop } $S$}
       }
       \ElseIf{$S$ \text{overlaps} $C$ \text{by more than X\% of } $Area(C)$}{
         {\text{Drop } $C$}
       }
       \Else{
         {\text{Clip $S$ using $C$}}
       }
    }
  }
\caption{Resolving overlap conflicts between stitched and non-boundary detections.}
\label{algo:clipstitched2}
\end{algorithm}

Surviving boundary fragments that failed to stitch are propagated to subsequent steps to maximize recall. The resulting set of new and surviving shapes is termed post-stitched detections $S$.

\subsubsection*{5. Chronological Validation and Deduplication}

This stage validates each post-stitched detection against evidence from the entire image stack to select the best final representation.

\begin{itemize}
    \item 
Image Quality Scoring: First, we compute a quality score for every image $I$ in the bucket as a weighted combination of contributions from image age, resolution, and a proxy measure of model performance on that image (typically, we choose the number of image detections in the bucket as a proxy for cloud absence and color balance):
\begin{equation}\label{eqn:imquality}
\begin{aligned}
Q =& w_{age} * \left(1 - \frac{min(age(I), A_{cap}}{A_{cap}})\right) + \\
  &w_{res} * (1 - min(1.0, res(I))) + \\
 &w_{perf} * \frac{min(perf(I), P_{cap}}{P_{cap}},
\end{aligned}
\end{equation}
where $w_{age}, w_{res}, w_{perf}$ are the weights determined empirically based on the desired trade-offs between up-to-date-ness, image resolution, and model performance (don't need to sum up to 1),
$age(I)$ is the age of $I$ in years, $res(I)$ is the resolution of $I$ in meters per pixel, $perf(I)$ is the performance score of the model on $I$ expressed as the number of detections obtained in the bucket cell, $A_{cap}$ is the age of the oldest acceptable image, and $P_{cap}$ is the cap on the ``detection population'' within the bucket cell ($I$ can have more but won't get a further score boost).

The top n scoring images ($M'$) are selected for validation. All detections from $S$ that link to images in $M'$ are collected in $S'$, sorted by image id according to the order in $M'$ with a secondary descending sort by model confidence value.

\item
Chronological Validation:
For each candidate detection s' in $S'$, a chronological presence/absence sequence (0s and 1s) is built by checking for confirming overlaps from all other images in $M'$. Incomplete tiles ($T_{incomplete}$) overlapping $s'$ are ignored.
This sequence is traversed from most recent to oldest. A running $validationScore$ is tallied, adding or subtracting the $simpleQualityScore$ of each image based on whether it confirmed (1) or denied (0) the detection's presence. The $simpleQualityScore$ uses only age and resolution,
 dropping the proxy contribution for model performance, and uses sigmoid functions for rapid decay as $age(I)$ and $res(I)$ grow:
\begin{equation}
simpleQualityScore = \frac{\frac{1}{1 + e^{age(I)}} + \frac{1}{1 + e^{1 + res(I)}}}{0.5 + \frac{1}{1 + e}},
\end{equation}
where, in order to normalize, the denominator is just the maximum achievable score in the numerator.

\item
Decision and Merging:

\begin{itemize}
    \item 

Early Exits: The process stops if the validationScore crosses an early accept ($earlyAcceptMinScore$) or early reject           ( $earlyReject\ MaxScore$) threshold. Otherwise, the detection is accepted if its final score is positive.
\item
Freshness Priority: To prioritize newness, the accept threshold can be set low enough to be cleared by a single, recent, high-quality image, correctly handling newly built objects.
\item
Merge into Results: Accepted detections are added to the final set $S_d$, provided they don't significantly overlap with existing results. An incoming detection can ``kick out'' a previously accepted one if its validation score is substantially higher, and non-boundary shapes can displace leftover boundary fragments.
\end{itemize}
\end{itemize}

The complete algorithm for this validation step is as follows:
\begin{algorithm}[h]
\SetAlgoLined

  \For{\text{each detection $s'$ in} $S'$}{
  
    $m' \leftarrow image(d)$
    
    $validationScore = 0$
    
    \For{\text{$m \in M' | m \neq m'$ in ascending order of $age(m)$}}{
      Skip $m$ if any tile from $m$ that overlaps $s'$ is in $T_{incomplete}$.
      
      Compute $O \leftarrow$ overlap area of $s'$ with $\cup\{{s \in S' | image(s) = m}\}$.
      
      \If{$O > minValidationArea$}{$validationScore += simpleQualityScore(m)$}
      \Else{$validationScore -= simpleQualityScore(m)$}

      \If{$validationScore > earlyAcceptMinScore$}{\text{\emph{Merge} $s'$ into result $S_d$ if no overlaps exceed $T$}}
      \ElseIf{$validationScore < earlyRejectMaxScore$}{\text{Drop $m'$ and goto next $s'$}}
    }
    \If{$validationScore > 0$}{\text{\emph{Merge} $s'$ into result $S_d$ if no overlaps exceed $tol$}}
  }
  \caption{Validation of Post-stitched detections}
  \label{algo:finalvalidation}
\end{algorithm}

\subsubsection{Final Output Generation}
  
Finally, to prevent duplicate outputs from adjacent parallel buckets, a detection is only included in a bucket's final output if its geometric centroid lies within that bucket's core area (excluding the processing margin). This ensures each object is represented exactly once in the final, consolidated dataset.

The complete algorithm is shown in Algorithm \ref{algo:bat_workflow}.

\section{Dagger Identification and Removal}\label{sec:dagger}

\begin{figure}[h]
    \centering
    \includegraphics[width=0.23\textwidth]{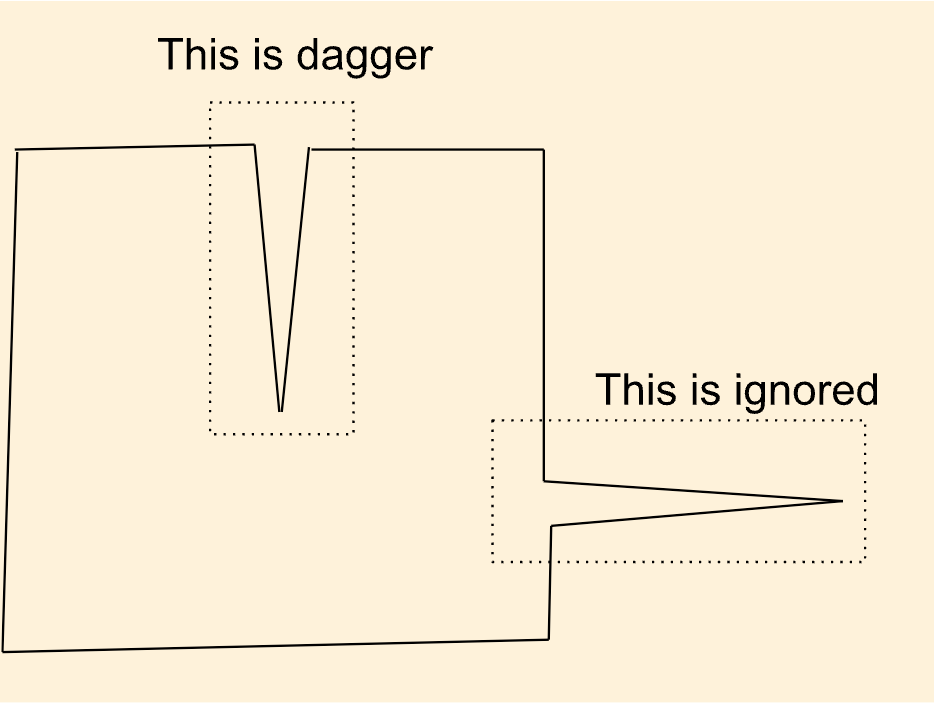}
    \includegraphics[width=0.23\textwidth]{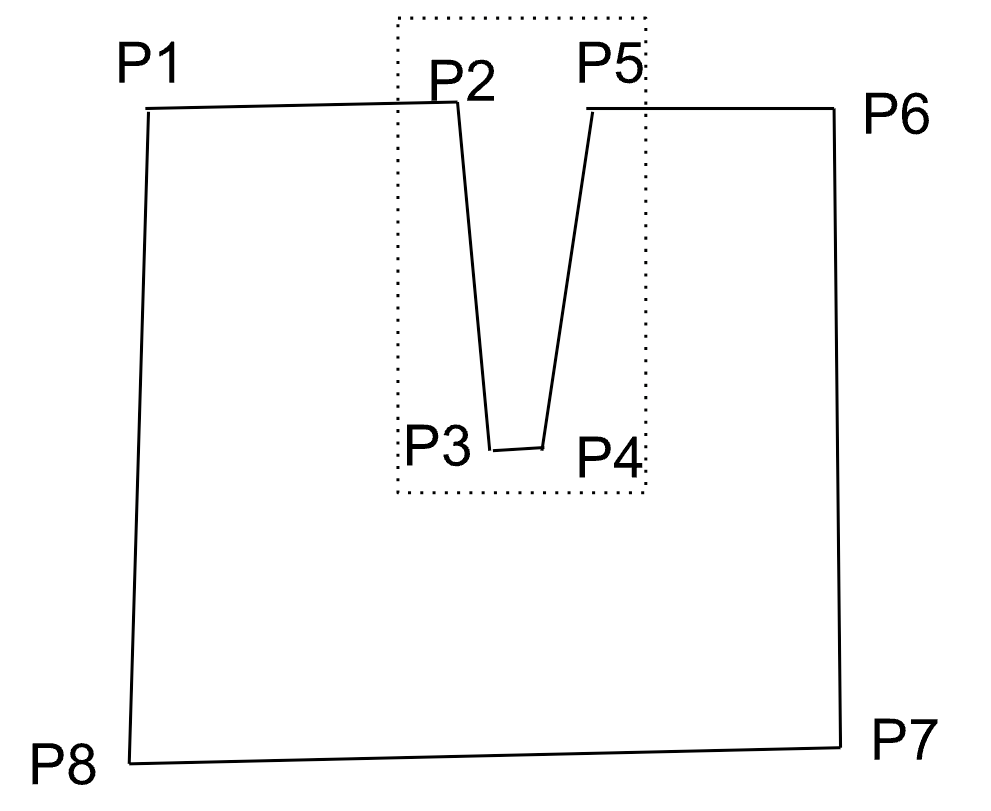}
    \caption{Examples of daggers. Points $p2,p5$ form the base of the dagger on the right.}
    \label{fig:dagger_example}
\end{figure}

Informally, daggers refer to `dagger-shaped' formations in the boundary of a polygon, which are pointing inwards. 
Some examples are shown in Figure \ref{fig:dagger_example}.

\subsection*{Dagger Removal}

\begin{figure}
    \centering
    \includegraphics[width=0.4\textwidth]{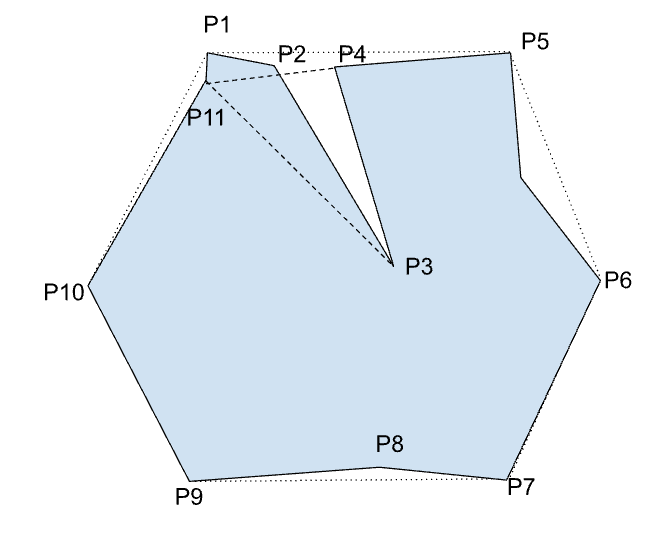}
    \caption{We only consider daggers such as \{P2, P3, P4\}, which fully reside within the segment (P1, P5) of the convex hull.  Even though \{P3, P4, P11\} looks like a dagger, smoothening it will cut away a big part of the original polygon and change its convex hull.}
    \label{fig:daggerhull}
\end{figure}

Let $P= \{p_1, p_2, ..., p_n\}$ denote a sequence of vertices of a polygon.
Daggers consist of consecutive points $p_i,\ldots,p_j$
on the polygon boundary.
We assume $i < j$ following modular arithmetic [$mod(n)$] for indices.
Our dagger smoothening algorithm will try to preserve the convex hull of the polygon, and so we only consider daggers that `fully reside' within a side of its convex hull (see figure \ref{fig:daggerhull}, because otherwise the dagger smoothening algorithm (discussed below) may cut away some part of the convex hull bound, and change the polygon’s convex hull.

We check all pairs of points on the polygon which lie between some consecutive pair of points on its convex hull, 
$\mathcal{H}$,
and evaluate if the pair is a `base' of a dagger (see Figure \ref{fig:dagger_example} and Algorithm \ref{algo:dagger}).
The main heuristic employed is in identifying the base of a dagger (in $isDagger()$), which we describe in the following.
Once a dagger is identified, the set of points in the polygon are updated by removing all the points of the identified dagger, except for the base points (in $removeDagger()$).

\begin{algorithm}[h]
\SetAlgoLined
\KwIn{A Polygon with vertex sequence $P= \{p_1, p_2, ..., p_n\}$}
\KwOut{Polygon without daggers with vertices $P' \in P$}

$\mathcal{H}:$ Convex Hull of $P$ 

$Q = \mathcal{H} \cup P$ \tcp*{Points in $P$ which are on the hull}

\tcp{Iterate in sequence}
\For{$q_i \in Q$}{

$R$: sequence of points in $P$ lying between $q_i$ and $q_{i+1}$

\For{each pair $(r_i, r_j) \in R \times R $}{
    \If{\text{isDagger}$(r_i, r_j, P)$}{
        $P \leftarrow$ \text{removeDagger($r_i, r_j, P$)}\;
    }
}
}

\Return $P$

\caption{Dagger Removal}
\label{algo:dagger}
\end{algorithm}

\begin{figure}[h]
    \centering
    \includegraphics[width=0.25\textwidth]{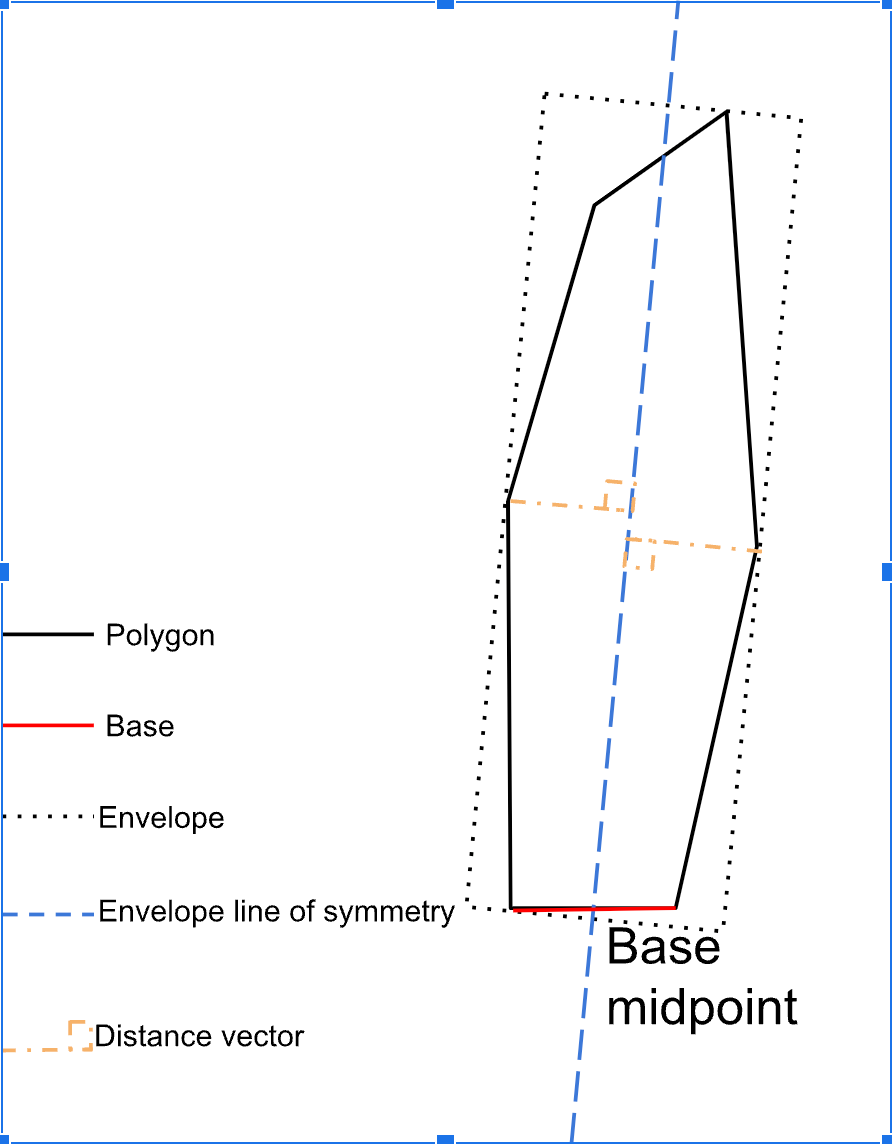}
    \caption{Dagger: Line of symmetry and Envelope}
    \label{fig:daggerenvelope}
\end{figure}

\subsection*{Dagger Identification}

The key idea of our dagger identification heuristic is to first find a rectangular envelope of the dagger points and then use the length and width of the envelope to identify the elongated dagger-like shape (see Figure \ref{fig:daggerenvelope}). 
To enable dagger identification in our heuristic we ensure that the envelope follows these constraints:
\begin{enumerate}
    \item The envelope’s line of symmetry, $L_s$, must pass through the midpoint of the {\it base edge}, i.e., the line segment formed by the base pair points.
    \item Let $L_1, L_2$ be the list of points which $L_s$ partitions $P$ into, i.e., $L_1 \cup L_2 = P$. Note that in general $L_1 \cap L_2 = \phi$, except for the points which fall on $L_s$ which are included in both lists.
    Let $d_1$ ($d_2$) be the distance of the point in $L_1$ ($L_2$) which is most distant from $L_s$. The envelope should be such that $d_1 \approx d_2$.
\end{enumerate}

\begin{figure}[h]
    \centering
    \includegraphics[width=0.4\textwidth]{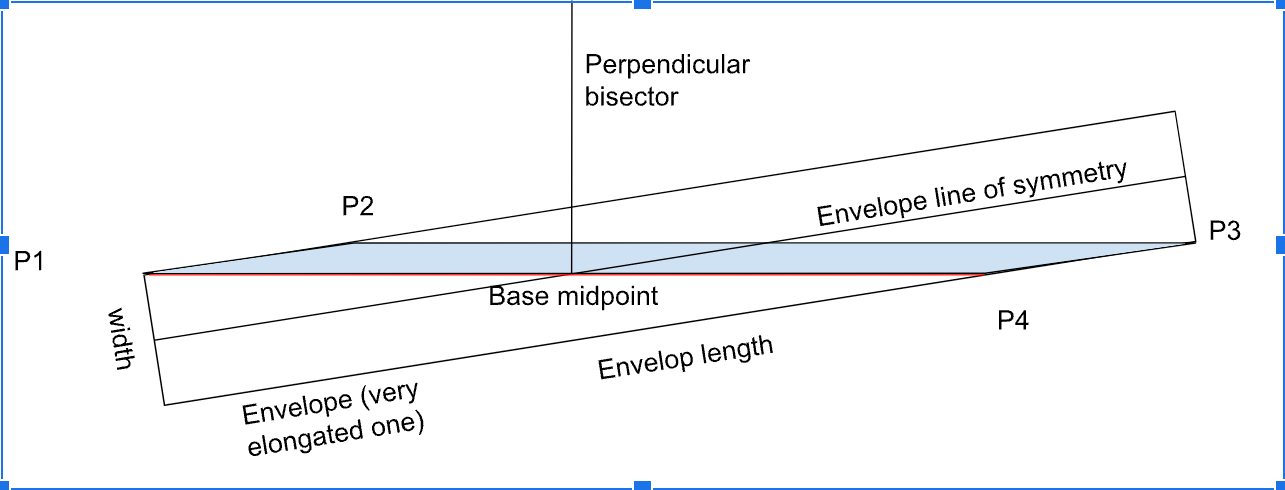}
    \caption{Case of a dagger where envelope width is smaller than the base edge length}
    \label{fig:dagger_long}
\end{figure}

\begin{algorithm}[th]
\SetAlgoLined
\KwIn{Polygon $P$, Points $q_i, q_j \in \mathcal{H}$, Dagger Threshold $d_t$, Angle Threshold $a_t$}
\KwOut{True, if $(q_i, q_j)$ is base of a dagger; False, otherwise.}
    Find the midpoints $m_b$ of the base edge $(q_i, q_j)$
    
    \For{\text{all pairs} $(q_h, q_k)| h \neq i \neq j \neq k$}{
    
    \tcp{Iteratively evaluate candidate daggers:}
    \tcp{$Q_l = \{q_i,\ldots, q_h,\ldots,q_k,\ldots,q_j\}$}
    
        Find midpoint $m_t$ of line segment $(q_h, q_k)$
        
        Calculate vector $V = m_t - m_b$
    
        \tcp{Consider vectors with angles which are not too steep}    
        \If{angle between $V$ and base edge $(q_i, q_j) < a_t$ }{
    
            \tcp{Rotate $Q_l$ around $M_b$ using vector $V$}
            $T_l \leftarrow \text{Rotate}(Q_l, V, M_b)$
            
            \tcp{Calculate minimum rectangle envelope}
            $R \leftarrow \text{MinimumRectangleEnvelope}( [T_i..T_j])$
            
            Cond1: $abs(\min_yR) \approx abs(\max_yR)$
            
            Elength = $max_xR - min_xR$
            
            Ewidth = $max_yR - min_yR$
            
            Criterion:  Elength / (max(Ewidth, Base length)
            
            \If{Cond1 and Criterion > $d_t$}{
                \Return True
            }
    
        }
    }
\Return False
\caption{Algorithm isDagger: checks if the the set of points considered, with $q_i,q_j$ as base points, forms a dagger}
\label{algo:isdagger}
\end{algorithm}

We also have to take into account the envelope’s orientation with respect to the base edge.
In many cases, the envelope satisfying the constraints above is rotated with respect to the base, and the width of the envelope is much smaller than the base edge (see Figure \ref{fig:dagger_long}).
Therefore, to identify a dagger, we check if the ratio, \emph{Envelope length / $\max$(envelope width, base edge length)}, is greater than a configurable threshold.

\begin{figure}[h]
    \centering
    \includegraphics[width=0.5\textwidth]{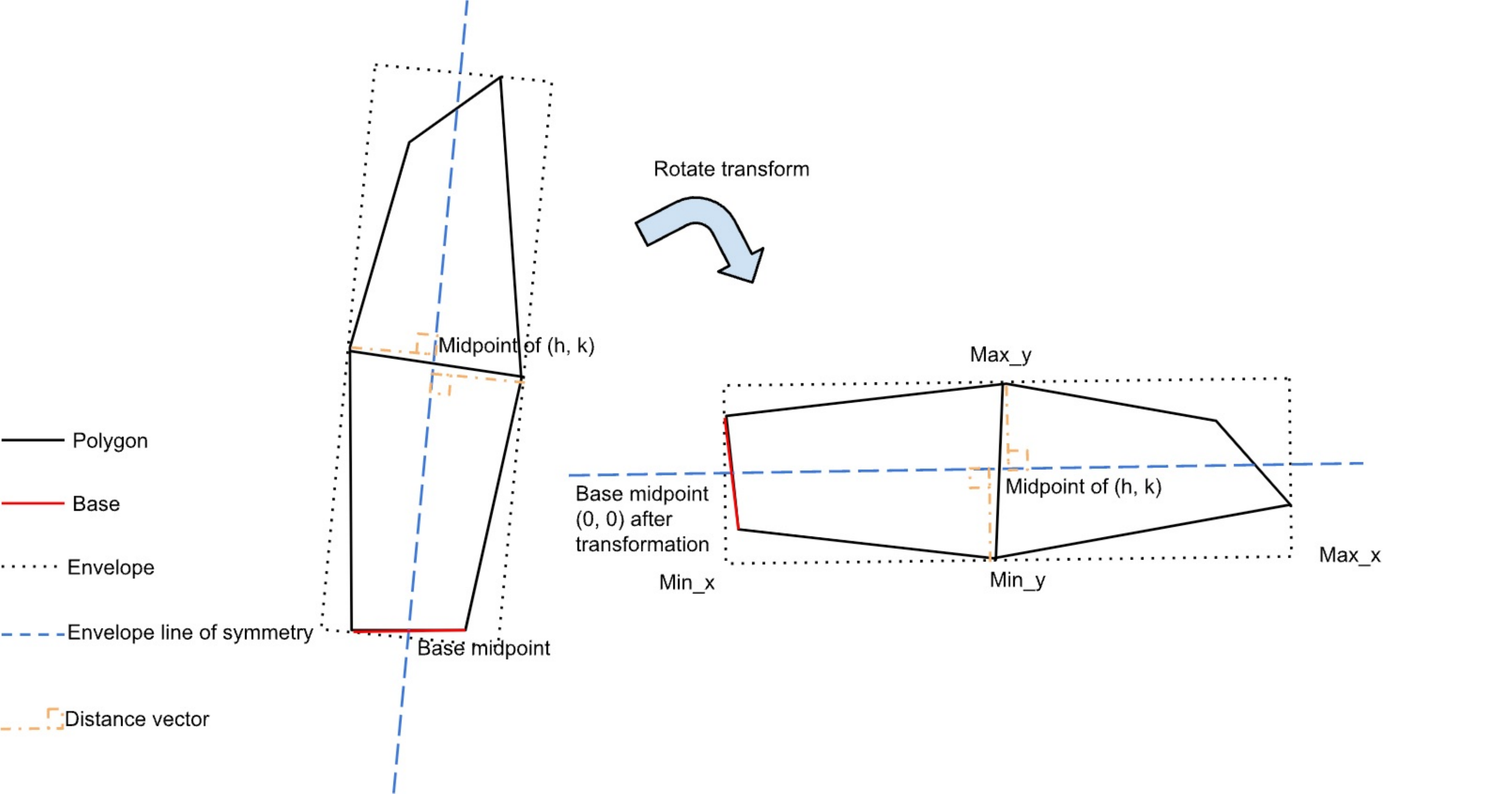}
    \caption{Rotation for dagger identification}
    \label{fig:dagger_rotation}
\end{figure}

Our dagger identification heuristic (Algorithm \ref{algo:isdagger}) is as follows.
Given a candidate base pair $q_i, q_j \in P$, we first find the base edge and its midpoint ($m_b$).
We then iterate through all pairs of points $(q_h, q_k)$ to find `eligible' rectangular envelopes.
To find such envelopes,
we first find the midpoints $m_t$ of the line segment $(q_h, q_k)$, and calculate the $V = m_t - m_b$.
This vector is a candidate line of symmetry.
We only consider those vectors that do not form a steep angle with respect to the base edge (by choosing those whose angle with the base edge is below a threshold $a_t$).
For the chosen candidate line of symmetry,
we apply a rotating transformation (see Figure \ref{fig:dagger_rotation}
on all the points in the candidate dagger $Q_l = \{q_i,\ldots, q_h,\ldots,q_k,\ldots,q_j\}$
using the vector $V$ such that the base midpoint is the origin and $V$ forms the horizontal axis after the rotation.

This transformation allows us to easily compute the minimum rectangular envelope (whose sides are parallel to the rotated coordinate axes) on the list of transformed points $T_l = \{t_i,\ldots, t_h,\ldots,t_k,\ldots$ $,t_j\}$.
The first required constraint on the envelope is true by construction ($V$ passes through $m_b$, the midpoint of the base edge).
The second constraint required can now be checked by $abs(min_y) \approx abs(max_y)$ since the absolute values of $min_y$ and $max_y$ give us the maximum distances from points in each half of plane cut by the envelope's line of symmetry.
The width of the envelope is given by $max_y - min_y$ and it's length is $max_x - min_x$.
With these values, we can check our ratio-based criterion to determine if the the set of points considered, with $q_i,q_j$ as base points, forms a dagger.

Note that the time complexity of Algorithm \ref{algo:isdagger} is $O(N^3)$, and that of Algorithm \ref{algo:dagger} is $O(N^5)$, where $N$ is the number of vertices of each polygon.

\subsection*{An Efficient Heuristic}

To improve runtime complexity, we can improve the isDagger algorithm by using a binary search algorithm to find the line of symmetry that passes through the dagger base. 

\begin{figure}[h]
    \centering
    \includegraphics[width=0.3\textwidth]{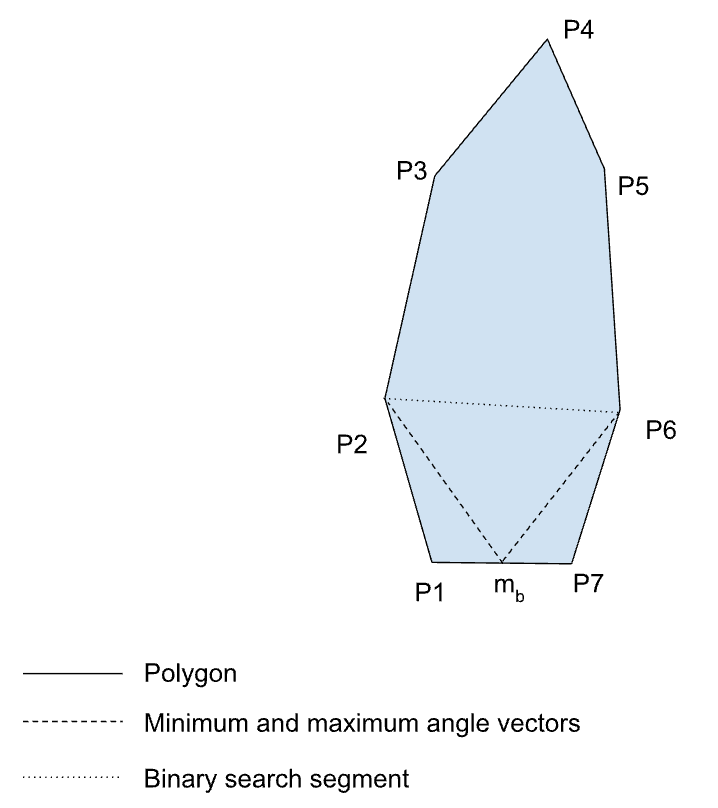}
    \caption{Schematic for calculation for binary search based heuristic}
    \label{fig:daggerheuristic}
\end{figure}

The idea of this binary search is as follows:
First, we define the concept of $d_{\delta}$ (distance-delta) value for a line of symmetry: 
For any candidate enclosing envelope, if we take the line of symmetry of that candidate envelope (which passes through the base midpoint $m_b$), and then calculate the difference between the maximum distance from any inner points on the left-half-plane of the line of symmetry and the maximum distance from any inner points on the right-half-plane of the line of symmetry, we will get a $d_{\delta}$  value. 
We will try to find the candidate envelope with $d_{\delta} \approx 0$  by doing a binary search on different positions of the line of symmetry that pass through $m_b$.
See figure \ref{fig:daggerheuristic}.

To define the bounds of the binary search: we first calculate the angle between the base edge vector and the vector from $m_b$ to each of the inner points (excluding the 2 base points). 
From this list of angles, we find the inner point with the smallest angle and the inner point with the largest angle.
We know that all of the other inner points will lie inside the arc made from the vectors from $m_b$ to the inner points with the smallest and the largest angle. Because of this, we know that for the two lines of symmetry which pass through these inner points with the smallest and the largest angle, the $d_{\delta}$ value of them will have \textit{different} signs (one value will be positive, the other will be negative).

With these 2 inner points (with smallest and largest angle) as the two bounds of the search segment (let's call them left and right bounds), we will then repeatedly do the following:
Calculate the line of symmetry that passes through $m_b$ and the mid point of the search segment.
Calculate  $d_{\delta}$ value of this line of symmetry.
If $d_{\delta} \approx 0$, we will return this line of symmetry.
Otherwise, if the sign of $d_{\delta}$ value we calculated is the same as the sign of $d_{\delta}$ value of the line of symmetry that passes through the left bound, we will update the midpoint of the search segment to be the new left bound. 
Else, we will update the midpoint of the search segment to be the new right bound.
If the length of the search segment is $\approx 0$, we return the line of symmetry that passes through $m_b$ and the midpoint of the search segment.

With this binary search algorithm, the runtime complexity of dagger identification is reduced to $O(NlogE)$, with $E$ being the length of the max distance between any pair of points in the dagger candidate polygon, and the complexity of the entire Dagger Removal algorithm is reduced to $O(N^3 log E)$.

\subsection{A Note on BAT and Dagger Removal}
\label{app:notebat}

BAT stitching and deduplication steps cannot be ``removed'' in a practical implementation, such as ours, that takes multiple satellite images as inputs. 
Hence an ablation without BAT cannot be done in our setting. 
Without BAT and dagger removal steps the pipeline comprises only ML model inference which is done on individual images and these images may be temporally and spatially misaligned -- the exact problem which BAT addresses.
The dagger removal step is optional and we conduct this ablation in section \ref{sec:syseval} (see Table \ref{tab:system_comparison}). 
BAT optimality is difficult to define since there is no ground truth available. BAT evaluates polygon detection quality based on several criteria/heuristics. Eg, ``under-stitching'' occurs when adjacent tiles lack complementary detections. This creates incomplete shapes gauged by an ``unstitched perimeter'' metric—edges perfectly aligned with tile boundaries.

\subsection{BAT parameters: Tuning, Sensitivity and Output Quality}
\subsubsection{Tuning BAT Parameters}
Parameter tuning is necessary to balance performance, accuracy, and completeness.
\begin{itemize}
    \item \textit{Image Stack}: Using more images for validation increases the computation cost per detection, while using more images for candidate selection increases completeness at the risk of introducing a few stale shapes. For ML models with high recall and false positives, higher early-exit reject thresholds drive more aggressive filtering of doubtful detections. Reducing the stack with quality thresholds improves both precision and performance.
    \item \textit{Deduplication}: Thresholds for penetration depth and overlap areas should be based on data distribution to align with desired results. E.g., setting these to 0 enables full deduplication but may severely impact recall due to image misalignment or overlapping model outputs. In practice, we pick at least $\sim$0.5m or more for depth, depending on average feature size.
    \item \textit{Bucket Size and Margins}: S2-level bucket size dictates worker load. Smaller cells increase computational overhead, while larger cells can cause memory issues in image dense areas. Higher margins, i.e., work unit overlaps, increase costs but are essential for stitching completeness/accuracy.
\end{itemize}

\subsubsection{Sensitivity of BAT Parameters}
BAT’s output can be steered to optimize for different use cases. E.g., margin size depends on the application’s specific feature size distribution.
Performance is best when BAT’s physical world knowledge matches use case dynamics. Thresholds, such as the amount of churn in tracked features, serve as proxies for these dynamics. High churn makes older imagery less useful for candidate detection or validation.
Thresholds for max-imagery-assets (candidates) and max-validation-assets (validating detections) signal world churn; higher values indicate lower churn. For ALU, we use up to 5 images for candidate detections. For building detections, which have high churn, we trust only the latest image. Both scenarios use up to 10 recent images for validation.
The max\_validating\_intersection\_area\_fraction\ threshold defines the required overlap for validation. ALU requires 25\% overlap, while buildings use 1\% overlap due to high churn and lower confidence in the alignment of later validation imagery.
BAT also regulates overlap in deduplicated output based on use case tolerance. Buildings have low tolerance: we specify a 0.5m depth overlap or 3 sqm total area overlap. Fields use a 1m depth overlap or 5\% of the smaller field's area.This prevents dropping adjacent detections and ensures high recall.

\subsubsection{Quality of BAT Outputs}
Optimality of output is difficult to define since there is no ground truth available to compare. Our system-level performance  (Sec.~\ref{sec:syseval}) and on-ground evaluation (Sec.~\ref{onground}) provides indirect evidence of optimality. BAT evaluates polygon detection quality based on several criteria such as:
\begin{itemize}
    \item \textit{Completeness}: Covering all real ground features.
    \item \textit{Precision}: Including only points belonging to real features.
    \item \textit{Instance Correctness}: Accurately representing real-world instance segmentation.
\end{itemize}
Many BAT functions are heuristically ``self-evaluating'', allowing quantitative measurement against quality assumptions. E.g., a frequent failure is ``under-stitching'', often caused when the model fails to produce a complementary detection in an adjacent tile. This results in incomplete detections that can be quantified by computing the remaining ``unstitched'' perimeter—edges perfectly aligned with tile boundaries. While theoretically possible for real features to align this way, it is extremely rare in practice. Thus, the \textbf{unstitched perimeter metric} is a reliable indicator of incomplete shapes when used with an appropriate threshold.

\section{Additional Results}


\subsection{Generalizability to other regions}
\label{app:generalizabiliy}
ALU is a scalable, data foundation layer which is currently available across several countries in the Asia Pacific (India, Indonesia, Vietnam, Malaysia, Japan) and in the African subcontinent (Kenya, Uganda, Ghana, Rwanda, Nigeria, Zambia).

To evaluate ALU's performance in other countries, we performed a zero shot inference of the ALU model, trained on labelled data in India, on a few countries in the Asia Pacific where we obtained annotated data in the same manner as described in Section~\ref{app:labeling}. 
We evaluated ALU on a total of 331, 151 and 330 images in Vietnam, Bangladesh and Malaysia respectively.
We compared the metrics described in the earlier sections and the results are shown in Table~\ref{tab:generalization_apac}. 
ALU shows comparable performance in these countries (Combined in Table~\ref{tab:generalization_apac}), with respect to the performance in India (Table~\ref{tab:panopt_comparison}).


\begin{table*}
\centering
\caption{Generalization capabilities of ALU, tested on other APAC countries. The model was trained on data in India.}
\label{tab:generalization_apac}
\begin{tabular}{lccccccccc}
\hline
\textbf{Country} & \textbf{Images} & \textbf{Instances} & \textbf{Label} & \textbf{mIoU} & \textbf{med IoU} & \textbf{OS} & \textbf{FNR} & \textbf{US} & \textbf{FPR} \\ \hline
Bangladesh & 20 & 3002 & Fields & 0.46 & 0.50 & 1.14 & 17.25 & 1.28 & 21.33 \\
& 74 & 2729 & Trees & 0.09 & 0.00 & 1.05 & 50.14 & 1.07 & 22.80 \\
& 6 & 60 & Clouds & 0.11 & 0.00 & 1.03 & 40.98 & 1.07 & 79.10 \\
& 1 & 3 & Wells & 0.00 & 0.00 & 0.00 & 100.00 & 0.00 & 100.00 \\
\hline
Malaysia & 47 & 328 & Fields & 0.56 & 0.73 & 1.29 & 20.57 & 1.09 & 83.42 \\
& 16 & 45 & Ponds & 0.09 & 0.00 & 1.00 & 90.70 & 1.00 & 92.75 \\
& 211 & 14028 & Trees & 0.14 & 0.00 & 1.06 & 44.79 & 1.23 & 28.98 \\
& 62 & 359 & Clouds & 0.21 & 0.03 & 1.01 & 33.91 & 1.41 & 37.82 \\
& 1 & 1 & Wells & 0.00 & 0.00 & - & 100.00 & - & 100.00 \\
\hline
Vietnam & 224 & 10269 & Fields & 0.61 & 0.81 & 1.15 & 17.97 & 1.07 & 21.55 \\
& 42 & 145 & Ponds & 0.00 & 0.00 & 0.00 & 100.00 & 0.00 & 100.00 \\
& 264 & 21248 & Trees & 0.12 & 0.00 & 1.20 & 68.41 & 1.09 & 12.85 \\
& 73 & 287 & Clouds & 0.20 & 0.03 & 1.03 & 45.59 & 1.36 & 26.92 \\
& 1 & 1 & Wells & 0 & 0 & 0 & 100 & 0 & 100 \\
\hline
\end{tabular}

\end{table*}

\subsection{Stratified Results}
\label{app:stratified}
\begin{figure*}
    \centering
    \includegraphics[width=0.75\linewidth]{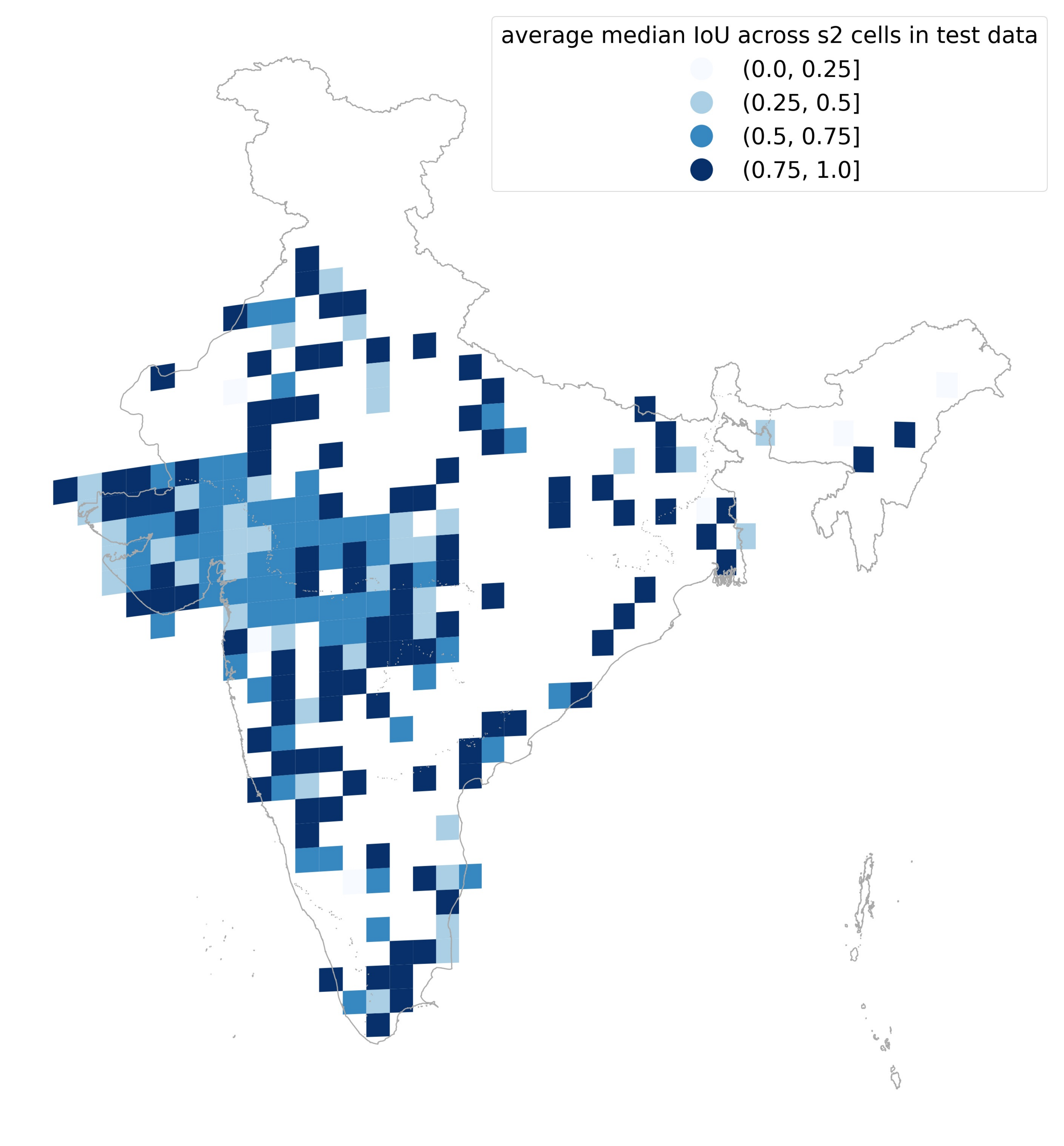}
    \caption{Choropleth depicting the average median IoU within various s2 cells (level 7) in the test dataset. A darker shade of blue indicates a higher medIoU. Best viewed in color.}
    \label{fig:choropleth_mediou}
\end{figure*}

We report ALU's performance (on the test set, class fields) across (1) agroclimatic zones, (2) agroecological regions  (3) seasons and (4) field sizes.
Agroclimatic region indicates the major climates, each of which is suitable for specific crops, while agroecological regions include other indicators like soil groups, effective rainfall etc.
The annotated satellite data also contained latitude and longitude information corresponding to each identified polygon and timestamps, based on which these stratifications were done. 
The test data covered 13 of 15 major agroclimatic~\cite{sinha201515} and 17 of 20 major agroecological~\cite{kumari2023spatio} regions of India, as described by the NITI Aayog and Planning Commission. 
In addition, we map each polygon to an agricultural season in India, namely Kharif (monsoon, June-October), Rabi (winter, October-March), and Zaid (summer, March-June). 
We calculate the median size of all fields present in each image of the test dataset. The median field size was used to group images into 3 quantiles, corresponding to small (1.82e+02 - 1.77e+03 sq. m), medium (1.77e+03 - 4.57e+03 sq. m) and large (1.77e+03 - 2.30e+05 sq. m) sized fields.
Results on each stratified group is shown in Table~\ref{tab: stratified_alu_results}. 
To obtain the spatial distribution of our accuracy, we mapped each test data point to an s2 cell, using the latitude and longitude of the field polygons. We then computed the mean of the median IoU of all images within a specific s2 cell (level 6).
Figure~\ref{fig:choropleth_mediou} shows the results as a choropleth of the average median IoU values.

\begin{table*}
\caption{Stratified results of ALU (fields) on test data split, on agroclimatic, agroecological regions, agricultural seasons and field sizes.}
\begin{tabular}{p{0.25cm}p{5cm}p{0.5cm}p{0.75cm}p{0.5cm}p{0.5cm}p{0.5cm}p{0.5cm}p{0.5cm}}
\toprule
& \textbf{Stratification} &
\textbf{mIoU} &
\textbf{medIoU} &
\textbf{OS }&
\textbf{FNR} &
\textbf{US} &
\textbf{FPR} & \textbf{Count} \\
\midrule
\multirow{13}{*}{\rotatebox{90}{\textbf{Agroclimatic Regions}}}
& Central Plateau and Hills Region & 0.57 & 0.65 & 0.98 & 5.88 & 0.99 & 13.74 & 95 \\
& East Coast Plains and Hills Region & 0.63 & 0.67 & 0.95 & 8.06 & 0.88 & 6.82 & 12 \\
& Eastern Plateau and Hills Region & 0.34 & 0.34 & 0.79 & 8.10 & 0.97 & 10.33 & 7 \\
& Gujarat Plain and Hills Region & 0.59 & 0.67 & 0.98 & 5.96 & 1.07 & 16.22 & 34 \\
& Lower Gangetic Plain Region & 0.56 & 0.66 & 0.94 & 7.12 & 0.97 & 10.33 & 69 \\
& Middle Gangetic Plain Region & 0.53 & 0.62 & 0.98 & 16.54 & 0.93 & 11.73 & 14 \\
& Southern Plateau and Hills Region & 0.64 & 0.72 & 1.03 & 3.70 & 0.97 & 4.62 & 17\\
& Trans Gangetic Plain Region & 0.55 & 0.63 & 0.93 & 6.62 & 0.96 & 11.84 & 192\\
& Upper Gangetic Plain Region & 0.53 & 0.61 & 0.89 & 6.41 & 0.94 & 15.62 & 176 \\
& West Coast Plains and Ghat Region & 0.50 & 0.55 & 0.76 & 2.49 & 0.88 & 7.58 & 7 \\
& Western Dry Region & 0.53 & 0.57 & 0.93 & 15.18 & 0.95 & 14.79 & 20 \\
& Western Himalayan Region & 0.50 & 0.57 & 1.06 & 16.46 & 1.05 & 22.53 & 14 \\
& Western Plateau and Hills Region & 0.60 & 0.67 & 1.12 & 3.42 & 1.01 & 10.08 & 15 \\
\midrule
\multirow{28}{*}{\rotatebox{90}{\textbf{Agroecological Regions}}}
& Assam and Bengal Plain Hot Subhumid to Humid & 0.55 &	0.62 & 0.93 & 5.81 &	0.92	& 12.23 &	82 \\
& Central Highlands Gujarat Plan \& Kathiawar Peninsula Semi-Arid &	0.58 &	0.65 &	0.95 &	7.01 &	0.97 &	11.63 &	138 \\
& Central Highlands Hot Subhumid (dry) &	0.58 &	0.67 &	0.98 &	2.53 &	0.99 &	13.28 &	31 \\
& Deccan Plateau and Eastern Ghats Hot Semi-Arid &	0.00 &	0.00 &	0.00 &	0.00 & 0.00 &	0.00 &	2 \\
& Deccan Plateau Hot Semi-Arid	& 0.55 &	0.63 &	0.93 &	6.89 &	0.97 &	12.69 &	130 \\
& Eastern Coastal Plain Hot Subhumid to Semi-Arid &	0.66 &	0.67 &	0.98 &	10.32 &	0.91 &	10.02 &	8 \\
& Eastern Ghats and Tamil Nadu Uplands and Deccan Plateau Hot Semi-Arid &	0.81 & 	0.95 & 1.17 &	14.29 &	1.00 &	0.00 &	1 \\
& Eastern Himalayas Warm Perhumid &	0.76 &	0.87 &	1.14 &	3.15 &	1.09 &	1.25 &	5 \\
& Easter Plain Hot Subaumid  &	0.55 &	0.63 &	0.96 &	7.21 &	1.02 &	13.01 &	48 \\
& Eastern Plateau and Eastern Ghats Hot Subhumid &	0.55 &	0.62 &	0.86 &	4.04 &	0.84 &	4.06 &	16 \\
& Islands of Andaman-Nicobar \& Lakshadweep hot humid to perhumid island &	0.52 &	0.56 &	0.95 &	6.06 &	1.13 &	10.97 &	9 \\
& Karnataka Plateau &	0.67 &	0.82 &	1.26 &	12.04 &	1.06 &	19.72 &	3 \\
& North Eastern Hills Warm Perhumid &	0.58 &	0.65 &	0.99 &	4.29 &	1.02 &	4.43 &	14 \\
& Northern Plain including Aravallis Hot Semi-arid &	0.53 &	0.61 &	0.92 &	8.77 &	0.95 &	17.71 &	62 \\
& Northern Plain Hot Subhumid &	0.56 &	0.64 &	0.91 &	4.90 &	1.07 &	21.40 &	20 \\
& Western Himalaya Warm Subhumid &	0.45 &	0.49 &	0.90 &	20.39 &	0.81 &	12.76 &	11 \\
& Western Plain Kachchh and Part of Kathiawar Peninsula Hot Arid &	0.52 &	0.60 &	0.92 &	8.09 &	0.97 &	16.38 &	92 \\
\midrule
\multirow{4}{*}{\rotatebox{90}{\textbf{Seasons}}}
& Kharif & 0.58	& 0.65	& 0.90	& 4.12	& 0.92	& 10.32 &	18 \\
& Rabi &	0.56 &	0.63 &	0.97 &	7.83 &	1.00 &	13.37 &	268 \\
& Zaid &	0.55 &	0.63 &	0.92 &	6.36 &	0.94 &	13.62 &	147 \\
\\
\midrule
\multirow{3}{*}{\rotatebox{90}{\textbf{Size}}}
& Small & 0.62 & 0.71 & 1.14 & 10.60 & 1.17 & 11.10 & 232 \\
& Medium & 0.73 & 0.83 & 1.16 & 5.73 & 1.11 & 6.32 & 231 \\
& Large & 0.31 & 0.35 & 0.51 & 4.85 & 0.56 & 21.58 & 232 \\
\bottomrule
\end{tabular}
\label{tab: stratified_alu_results}
\end{table*}

\subsection{Aggregate Analysis \& Geo-spatial Distribution}\label{geospat}

In Tables \ref{tab:counts} and \ref{tab:area}, we provide quantitative results as counts of different features and their respective areas. Certain features like trees are not counted because we only identify separated clusters of trees and not individual trees.

\begin{table*}[]
\centering
\caption{Quantitative results obtained for counts of different features across India.}
\label{tab:counts}
\footnotesize
\begin{tabular}{p{3cm}p{1.8cm}p{1.8cm}p{1.8cm}p{1.8cm}}
\toprule
\textbf{State/Region} &
\textbf{Dug well} & \textbf{Pond} & \textbf{Field}
& \begin{tabular}[c]{@{}r@{}}\textbf{Other Water} \\ \textbf{Sources}\end{tabular} \\ 
\midrule
Andaman and Nicobar Islands & 325 & 62 & 1650311 & 7694 \\
Andhra Pradesh & 109809 & 8177 & 461141085 & 450349 \\
Arunachal Pradesh & 1787 & 28 & 7406486 & 484994 \\
Assam & 20887 & 6244 & 328502451 & 285814 \\
Bihar & 66301 & 5986 & 977699495 & 197499 \\
Chandigarh & 283 & 10 & 399281 & 475 \\
Chhattisgarh & 28331 & 8057 & 328047921 & 222971 \\
DNHDD\textasciicircum & 351 & 22 & 1457193 & 2298 \\
Delhi & 5882 & 339 & 9882902 & 11955 \\
Goa & 520 & 214 & 3043349 & 17205 \\
Gujarat & 440214 & 5762 & 463669606 & 317578 \\
Haryana & 244780 & 2580 & 194075900 & 70720 \\
Himachal Pradesh & 8963 & 116 & 15299203 & 349286 \\
Jammu and Kashmir & 13869 & 190 & 70843565 & 360154 \\
Jharkhand & 47638 & 5000 & 303626746 & 118892 \\
Karnataka & 360756 & 14147 & 575248792 & 223069 \\
Kerala & 3396 & 1778 & 36379993 & 125426 \\
Leh-Ladakh & 28719 & 49 & 11361794 & 430264 \\
Madhya Pradesh & 1298066 & 4228 & 908266510 & 366407 \\
Maharashtra & 2446355 & 51459 & 959370775 & 502505 \\
Manipur & 7284 & 1486 & 22476708 & 103691 \\
Meghalaya & 1043 & 139 & 8300045 & 51539 \\
Mizoram & 176 & 24 & 1103617 & 114558 \\
Nagaland & 372 & 33 & 3122313 & 84847 \\
Odisha & 25998 & 13833 & 605175056 & 312062 \\
Puducherry & 533 & 116 & 3140176 & 3587 \\
Punjab & 230672 & 2088 & 228325877 & 62851 \\
Rajasthan & 1176891 & 15223 & 541740773 & 197025 \\
Sikkim & 257 & 4 & 291467 & 52300 \\
Tamil Nadu & 281386 & 8369 & 582253573 & 250213 \\
Telangana & 117213 & 1410 & 422307075 & 154918 \\
Tripura & 668 & 1647 & 24119670 & 106918 \\
Uttar Pradesh & 102377 & 13670 & 1666110296 & 480587 \\
Uttarakhand & 5417 & 228 & 29739159 & 286366 \\
West Bengal & 37445 & 26002 & 676609231 & 867547 \\

\textbf{Total} & \textbf{7.115206e+06} & \textbf{1.987740e+05} & \textbf{1.047411e+10} & \textbf{7.680326e+06} \\  
\bottomrule
\end{tabular}

*Area in hectares \\
\textasciicircum Dadar and Nagar Haveli and Daman and Diu

\end{table*}
\begin{table*}[]
\centering
\caption{Quantitative results obtained for area covered by different features across India.}
\label{tab:area}
\footnotesize
\begin{tabular}{p{3cm}p{1.8cm}p{1.8cm}p{1.8cm}p{1.8cm}p{1.8cm}}
\toprule
\textbf{State/Region} &
\textbf{Dug well} & \textbf{Pond} & \textbf{Field}
& \begin{tabular}[c]{@{}r@{}}\textbf{Other Water} \\ \textbf{Sources}\end{tabular}
& \textbf{Trees} \\ 
\midrule
Andaman and Nicobar Islands              & 4.02 &	11.94 &	5.67e+05 &	77190.13 & 4213989.40    \\
Andhra Pradesh                           & 1372.57	5650.74	& 1.30e+08 & 1470862.69 & 32446444.03  \\
Arunachal Pradesh & 23.66 & 18.32 & 2.74e+06 & 592515.54 & 20295496.92 \\
Assam & 259.85 & 2280.95 & 4.36e+07 & 769645.70 & 31832309.63 \\
Bihar & 792.71 & 2571.06 & 1.09e+08 & 365162.02 & 19574382.72 \\
Chandigarh & 3.38 & 3.50 & 8.44e+04 & 630.47 & 129910.81 \\
Chhattisgarh & 329.95 & 5185.59 & 4.79e+07 & 356076.04 & 33073654.86 \\
DNHDD\textasciicircum & 4.13 & 2.61 & 2.22e+05 & 9634.46 & 400895.98 \\
Delhi & 90.42 & 104.80 & 2.97e+06 & 31358.70 & 1422364.10 \\
Goa & 6.23 & 127.53 & 1.59e+06 & 75471.26 & 4218068.83 \\
Gujarat & 5387.57 & 3039.44 & 1.93e+08 & 909500.56 & 26225521.78 \\
Haryana & 4201.14 & 936.26 & 6.30e+07 & 95421.55 & 4221812.33 \\
Himachal Pradesh & 83.97 & 49.04 & 3.90e+06 & 644268.37 & 13766472.30 \\
Jammu and Kashmir & 144.59 & 98.01 & 1.47e+07 & 655232.93 & 15757994.62 \\
Jharkhand & 523.25 & 1843.25 & 2.84e+07 & 218413.91 & 20378229.61 \\
Karnataka & 4806.12 & 2298.40 & 1.81e+08 & 871656.17 & 50687173.37 \\
Kerala & 45.37 & 660.40 & 1.11e+07 & 545947.67 & 34143307.89 \\
Leh-Ladakh & 284.50 & 11.93 & 6.45e+06 & 747041.68 & 4619395.00 \\
Madhya Pradesh & 16389.38 & 2595.52 & 2.85e+08 & 1132299.62 & 52653953.84 \\
Maharashtra & 31640.75 & 6984.77 & 2.98e+08 & 1544697.30 & 64186286.00 \\
Manipur & 98.60 & 720.65 & 2.92e+06 & 89482.91 & 8222069.34 \\
Meghalaya & 12.02 & 37.55 & 2.17e+06 & 40843.27 & 9009880.10 \\
Mizoram & 2.29 & 7.47 & 9.29e+05 & 78342.83 & 8958693.30 \\
Nagaland & 5.14 & 12.03 & 1.12e+06 & 58989.73 & 7941567.45 \\
Odisha & 305.29 & 5673.06 & 7.27e+07 & 769777.83 & 49123194.92 \\
Puducherry & 7.14 & 29.99 & 7.81e+05 & 8768.58 & 253732.79 \\
Punjab & 4145.56 & 910.61 & 7.02e+07 & 94906.58 & 4395430.79 \\
Rajasthan & 13093.32 & 2134.42 & 2.82e+08 & 587644.63 & 28785725.48 \\
Sikkim & 2.88 & 1.76 & 2.39e+05 & 90027.18 & 2420271.96 \\
Tamil Nadu & 3402.30 & 1797.59 & 1.17e+08 & 593626.91 & 40031487.69 \\
Telangana & 1442.43 & 373.73 & 8.15e+07 & 564203.59 & 19558730.51 \\
Tripura & 8.06 & 636.00 & 2.56e+06 & 49263.21 & 7511192.44 \\
Uttar Pradesh & 1244.56 & 7785.76 & 3.05e+08 & 1073659.16 & 40905024.81 \\
Uttarakhand & 58.12 & 163.33 & 8.39e+06 & 451000.79 & 19451513.57 \\
West Bengal & 458.50 & 12539.00 & 6.87e+07 & 720470.44 & 27419046.65 \\

\textbf{Total}                           & \textbf{9.07e+04} & \textbf{6.73e+04} & \textbf{2.44e+09} & \textbf{1.641e+07} & \textbf{7.08e+08} \\  
\bottomrule
\end{tabular}

*Area in hectares \\
\textasciicircum Dadar and Nagar Haveli and Daman and Diu

\end{table*}

\subsection{Comparison with Census Data}
\label{app:census}

We first obtained all ALU field polygons from historic ALU data, and mapped each polygon to a specific state/union territory (UT) by checking if the latitude and longitude of each ALU polygon was within the boundaries of a state/UT. We then obtained the aggregated agricultural field area of each state/UT, by summing up the individual areas of all ALU field polygons mapped to the state/UT. This was then compared with the overall agricultural area reported in the most recent agricultural census report in India~\cite{allindiareport}. We examined the percentage change in the ALU derived agricultural area per state/UT and the corresponding values from the census. The change across all states/UTs was divided into 4 quartiles of equal space, for easier visualization. These formed various categories: $< 25^{th}$ quantile, $25^{th}$ - $50^{th}$ quantile, $50^{th}$ - $75^{th}$ quantile and $> 75^{th}$ quantile.

\begin{figure*}
    \centering
    \includegraphics[width=0.75\textwidth]{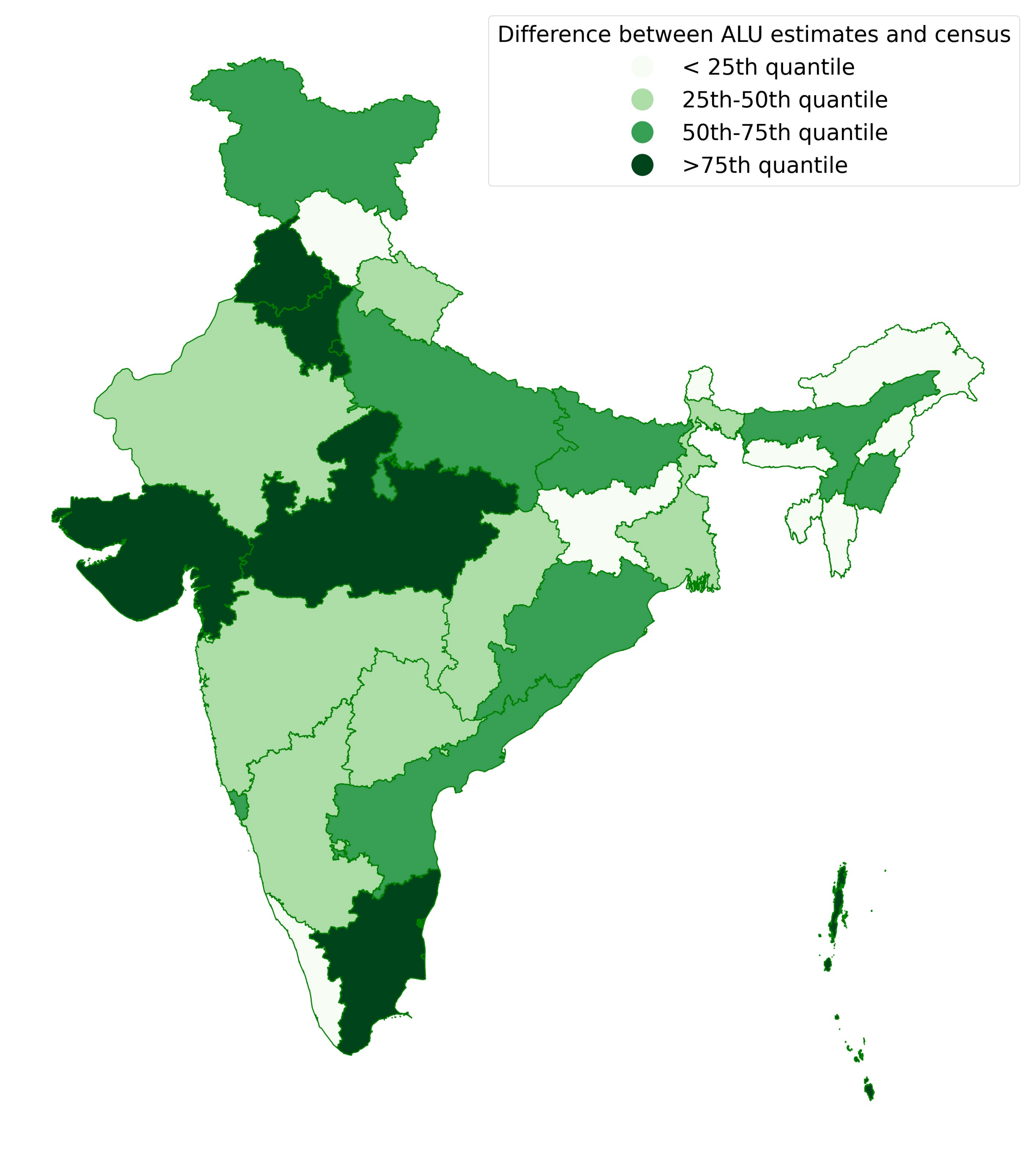}
    \caption{State-wise comparison of area of fields from ALU-based computation and agricultural census data, obtained from government sources. States in lighter shades are closer in values, while areas shaded darker differ more when comparing ALU results against census data. Best viewed in color.}
    \label{fig:alu_census}
\end{figure*}

Figure \ref{fig:alu_census} shows the choropleth comparing areas thus obtained for agricultural areas in each Indian state along with those reported in the previous census \cite{allindiareport}. 
States with a lower percentage change (lighter shaded areas) show higher concordance between ALU derived land area and census reported values, than those with a higher percentage change (darker shaded areas).
The reasons for disagreement include temporal mismatch, definition difference and model errors: the census report (2015-16) is older than ALU data (2019-21); the census defines fields in terms of operational holdings - area of entire holding may be used even when it is only partly agricultural; nevertheless, Pearson correlation (0.9275, p-val:3.16e-15) is strong between ALU estimates and census.

We observe that our system's performance varies across states.
States like Arunachal Pradesh, Himachal Pradesh, Jharkhand, Kerala, Meghalaya, Mizoram, Nagaland, Sikkim and Tripura show high agreement, while places like Tamil Nadu, Punjab, Puducherry, Madhya Pradesh, Haryana, Gujarat, Delhi, Chandigarh, Andaman and Nicobar islands show differences.

\subsection{On-Ground Validation}
\label{app:teamup}

Table \ref{tab:village_survey_results} shows the villages in which surveys were conducted and the accuracy of ALU outputs assessed through the surveys.

\begin{table*}[htbp]
\centering
\caption{On-ground validation in Telangana}
\label{tab:village_survey_results}
\begin{tabular}{@{}rlrrrrr@{}}
\toprule
\begin{tabular}[c]{@{}r@{}}\textbf{Sl.} \\ \textbf{No.}\end{tabular} & \textbf{Village} & \begin{tabular}[c]{@{}r@{}}\textbf{No. of} \\ \textbf{Surveys}\end{tabular} & \begin{tabular}[c]{@{}r@{}}\textbf{No. of} \\ \textbf{Farm Fields}\end{tabular} & \begin{tabular}[c]{@{}r@{}}\textbf{Over} \\ \textbf{Segmentation}\end{tabular} & \begin{tabular}[c]{@{}r@{}}\textbf{Under} \\ \textbf{Segmentation}\end{tabular} & \begin{tabular}[c]{@{}r@{}}\textbf{Boundary} \\ \textbf{Error}\end{tabular} \\
\midrule
1  & SUNDARAGIRI         & 5   & 60    & 10 & 0 & 0  \\
2  & ARNAKONDA           & 2   & 29    & 2  & 0 & 0  \\
3  & YASWADA             & 9   & 61    & 0  & 7 & 0  \\
4  & KHASIMPET           & 19  & 128   & 0  & 4 & 14 \\
5  & PARVELLA            & 1   & 1     & 0  & 3 & 0  \\
6  & JANGAPALLE          & 10  & 105   & 11 & 2 & 5  \\
7  & KANDUGULA           & 25  & 343   & 18 & 0 & 18 \\
8  & IRUKULLA            & 29  & 239   & 6  & 4 & 18 \\
9  & DURSHED             & 6   & 13    & 4  & 6 & 2  \\
10 & MALKAPUR            & 9   & 53    & 10 & 2 & 0  \\
11 & REKURTHI            & 13  & 25    & 23 & 0 & 3  \\
12 & VELDI               & 1   & 9     & 0  & 0 & 3  \\
13 & GATTUDUDDENAPALLE   & 2   & 12    & 0  & 0 & 0  \\
14 & RAMADUGU            & 5   & 50    & 3  & 0 & 9  \\
15 & KACHAPUR            & 7   & 64    & 0  & 0 & 4  \\
16 & ALUGUNUR            & 2   & 10    & 0  & 0 & 0  \\
17 & NUSTULAPUR          & 5   & 50    & 13 & 2 & 1  \\
18 & SAIDAPUR            & 7   & 41    & 0  & 0 & 5  \\
19 & CHALLOOR            & 7   & 76    & 2  & 0 & 30 \\
\midrule
   & \textbf{Total}      & \textbf{164} & \textbf{1,369} & \textbf{102} & \textbf{30} & \textbf{112} \\
& \textbf{Percentage (\%)} && \textbf{7.45} & \textbf{2.19} & \textbf{8.18} \\
\bottomrule
\end{tabular}
\end{table*}

\subsection{Qualitative Comparisons}
\label{app:qual}

\begin{figure*}
    \centering
    \includegraphics[width=\linewidth]{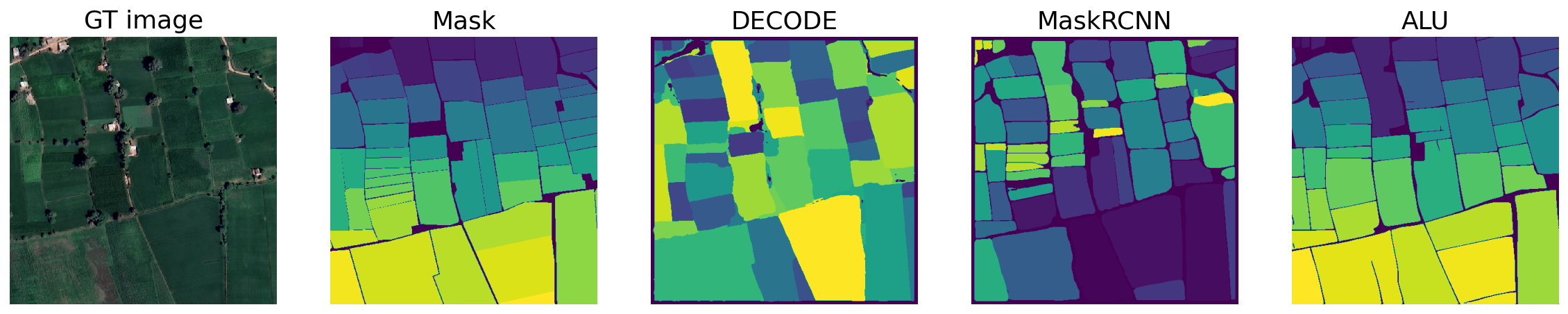}
    \includegraphics[width=\linewidth]{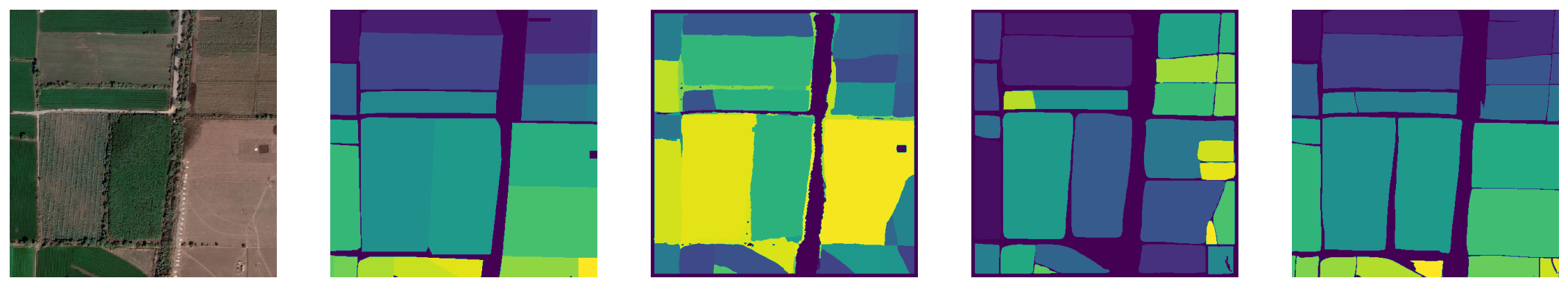}
    \includegraphics[width=\linewidth]{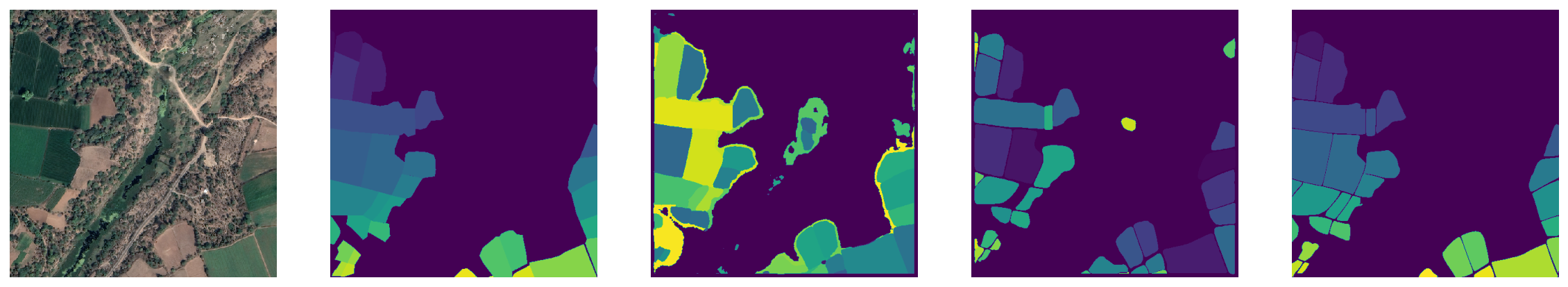}
    \includegraphics[width=\linewidth]{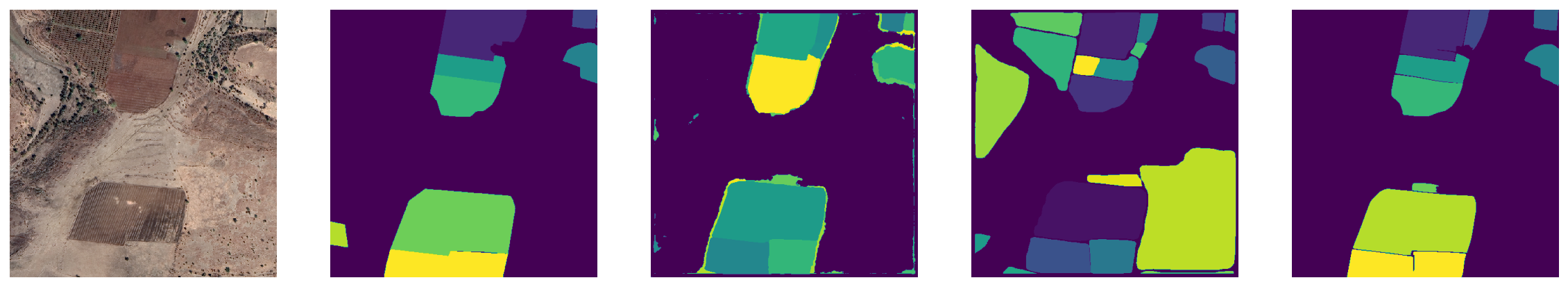}
    \includegraphics[width=\linewidth]{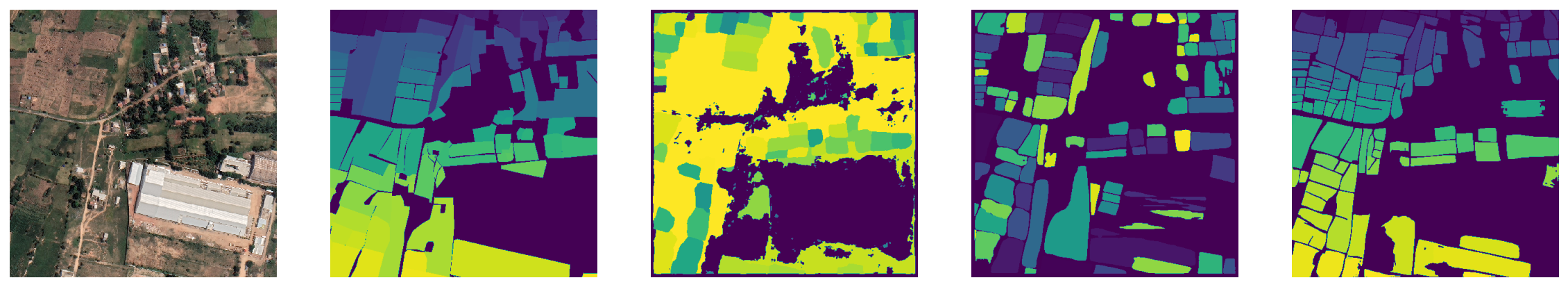}
    \caption{Each row shows an example of (left to right): the satellite image, the human-annotated ground instance mask and predicted masks from DECODE, MaskRCNN and ALU. Best viewed in color.}
    \label{tab:ground_qual}
\end{figure*}

\begin{figure*}
    \centering
    \includegraphics[width=1\textwidth]{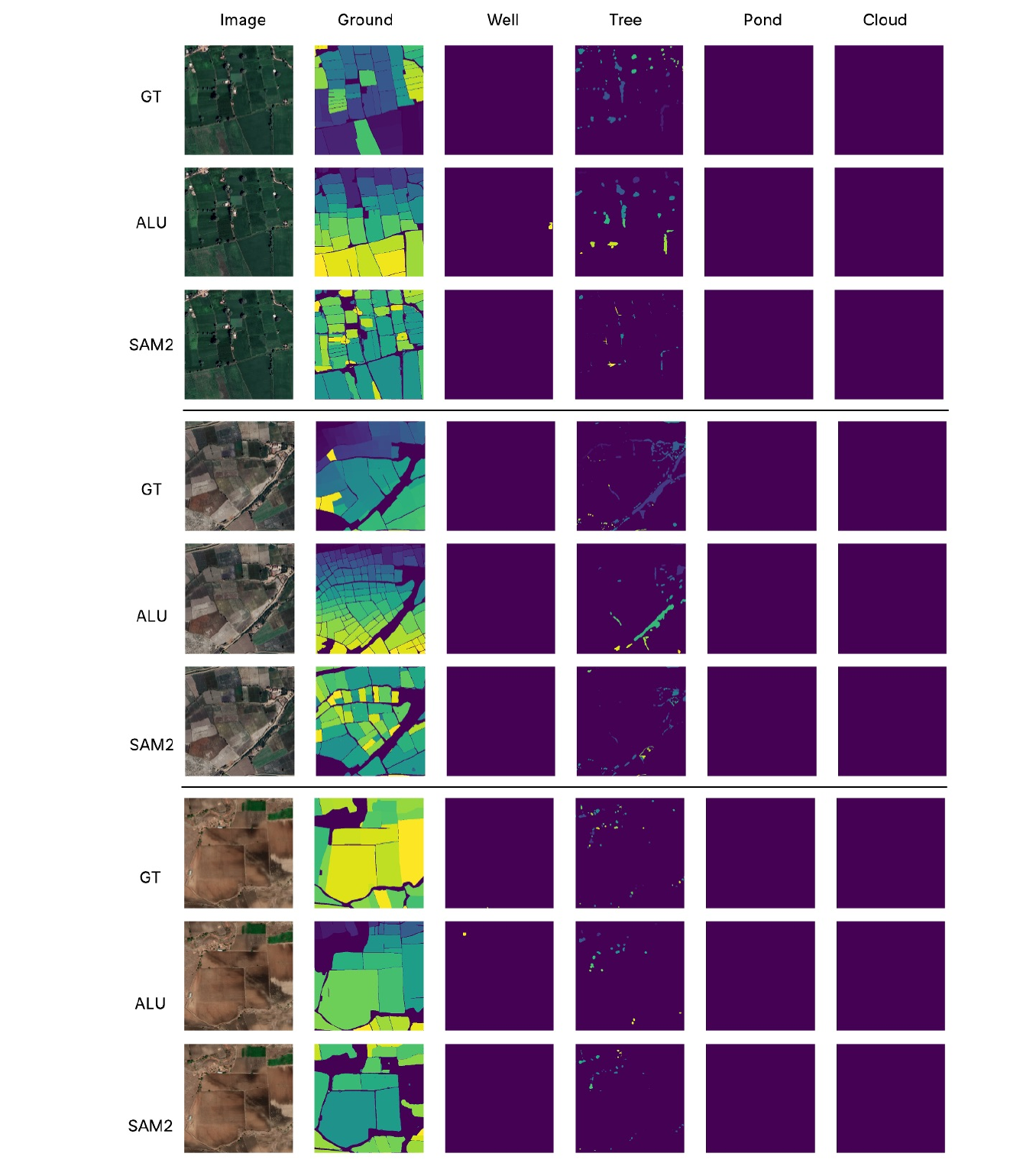}
    \caption{Ground Truth (GT) satellite image and human annotated masks (Left to Right: ground, well, tree and cloud instances). Subsequent rows show ALU (our) and SAM2 outputs. Columns 2-6 for ALU and SAM2 are the predicted masks. Best viewed in color.}
    \label{fig:panoptic}
\end{figure*}

Figure \ref{tab:ground_qual}
shows sample images, and their ground instance segmentation masks predicted by DECODE, MaskRCNN and ALU.
Fig~\ref{fig:panoptic} shows the results of panoptic segmentation from ALU and SAM2 (the second best performing model), as well as the corresponding ground truth images. As seen in the figure (rows 2 and 5, column 4), Unlike SAM2, ALU is able to identify not just the field boundaries well, but also most of the trees annotated in the dataset. However, there are also certain situations when ALU identifies wells where they do not exist (rows 2 and 8, column 3) and tends to over-segment field instances (row 5, column 2). The over-segmentation issue is also visible in Figure \ref{tab:ground_qual} (row 5, column 5).

\begin{figure*}
    \centering
    \includegraphics[width=0.49\textwidth]{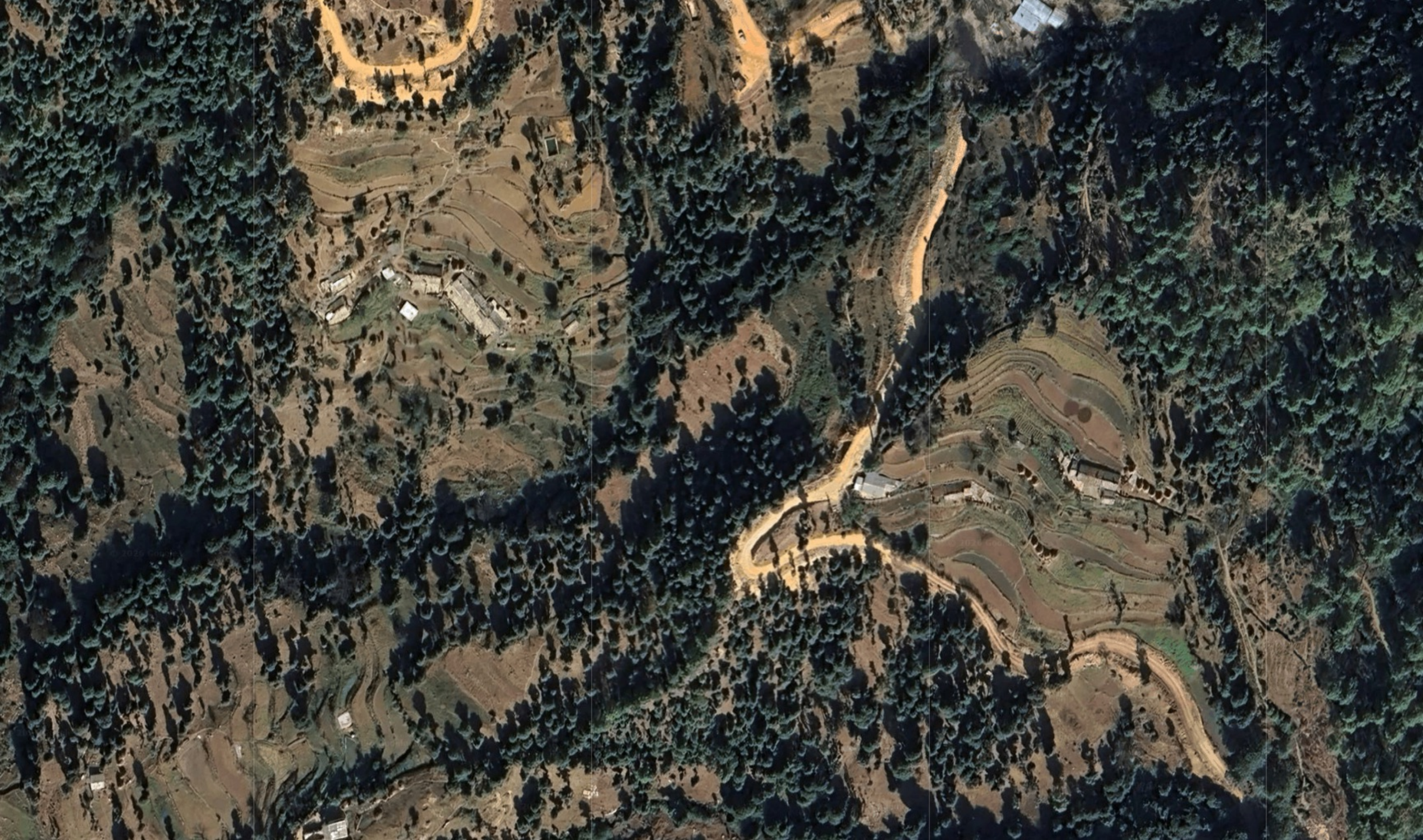}
    \includegraphics[width=0.49\textwidth]{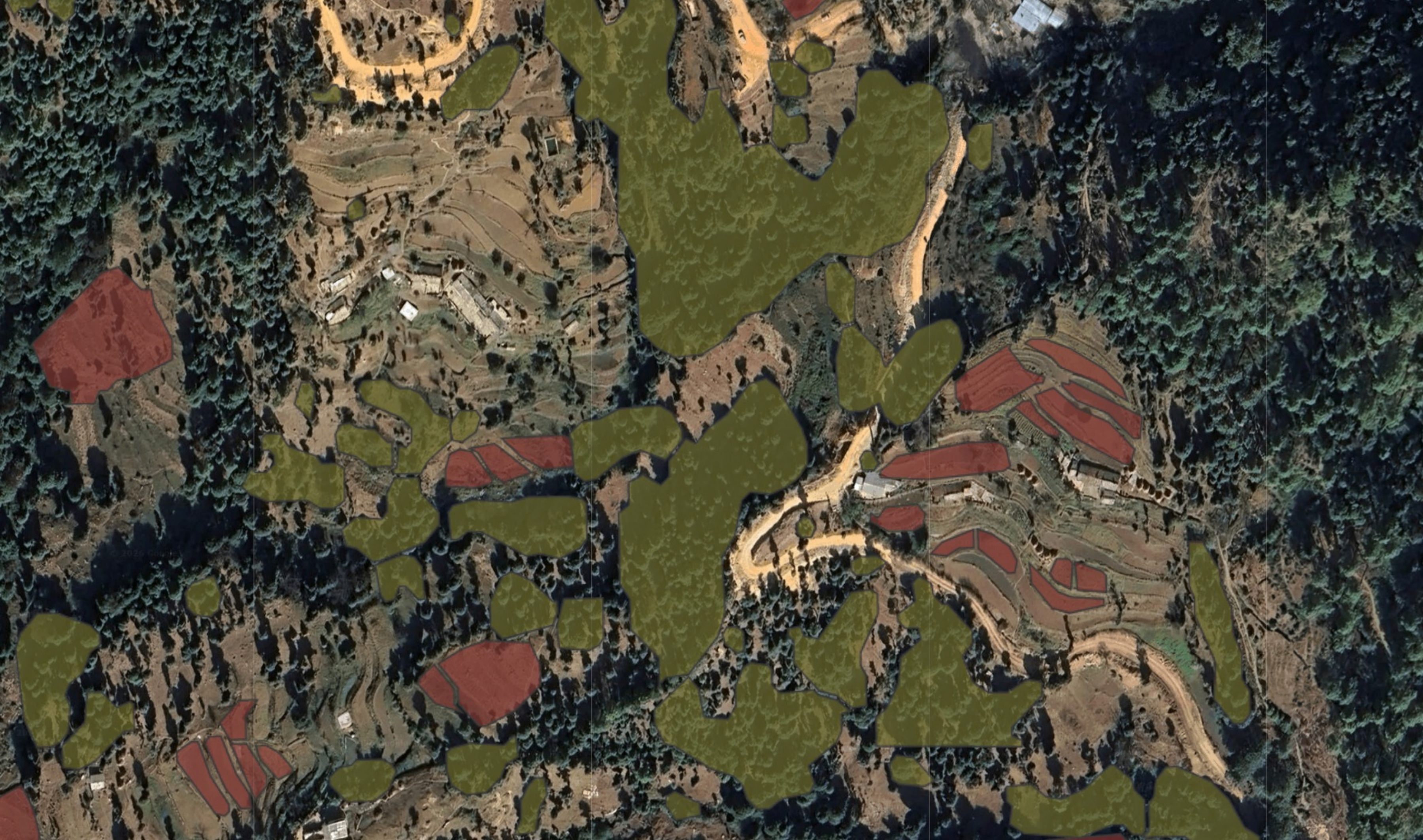}
    \includegraphics[width=0.49\textwidth]{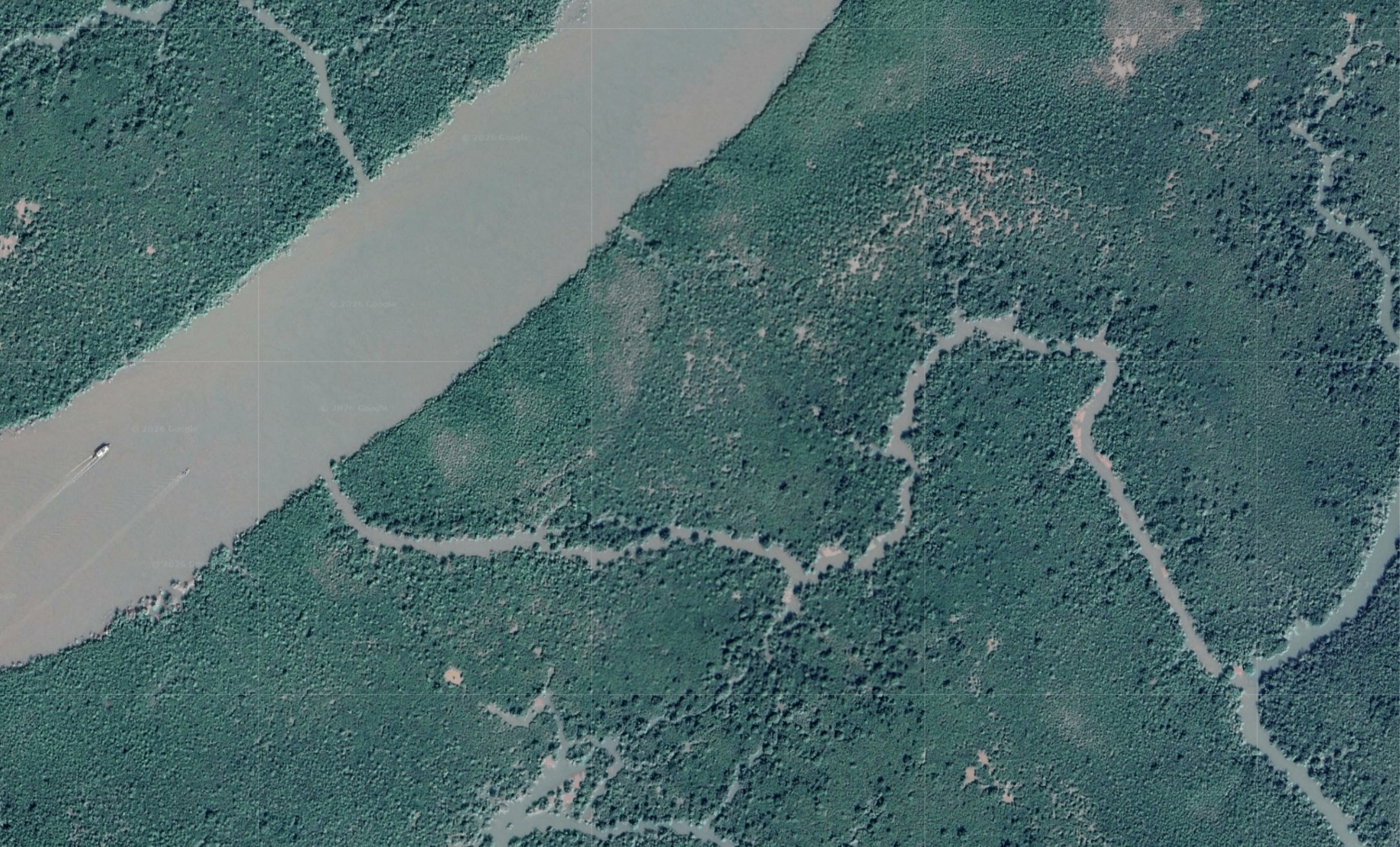}
    \includegraphics[width=0.49\textwidth]{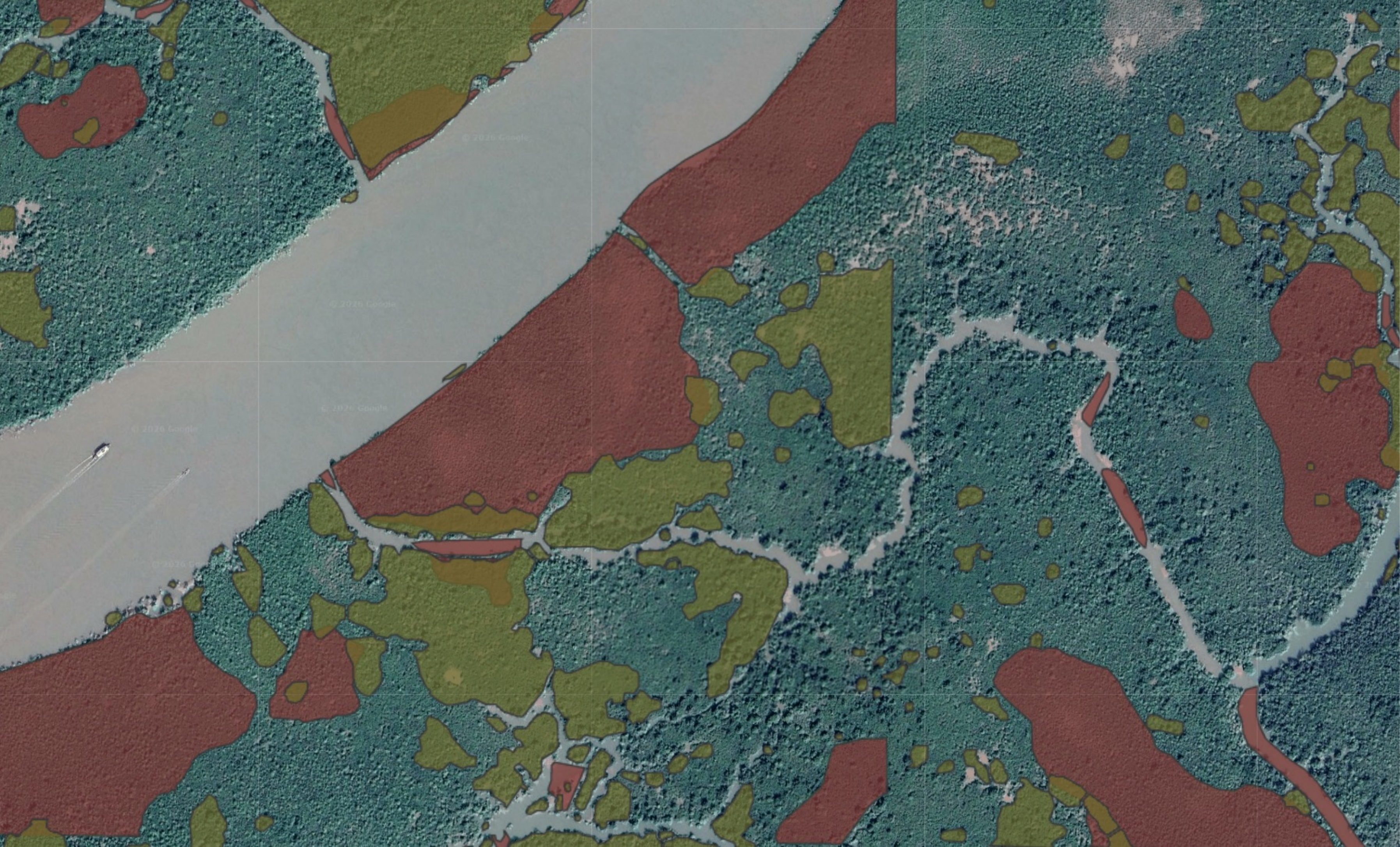}
    \includegraphics[width=0.49\textwidth]{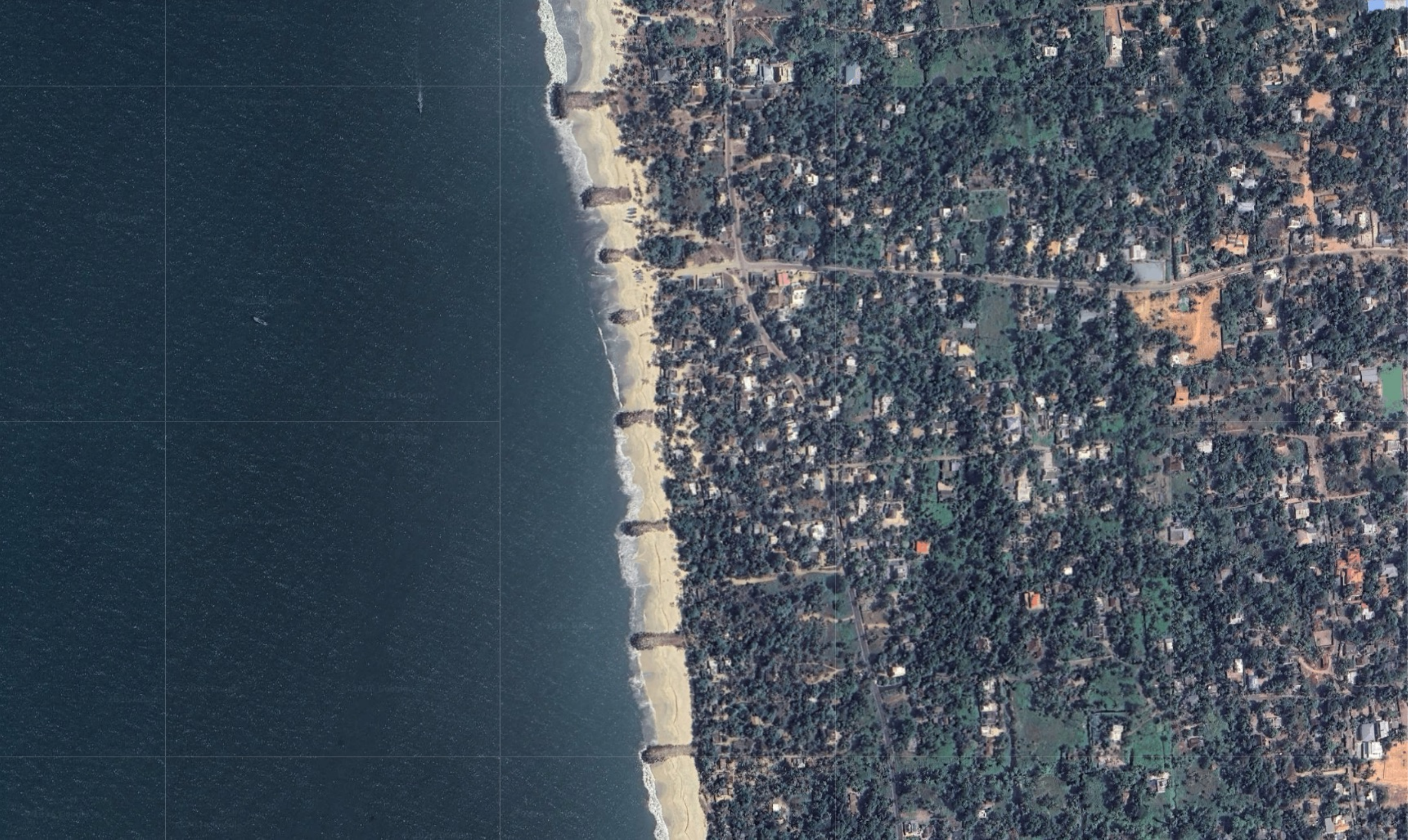}
    \includegraphics[width=0.49\textwidth]{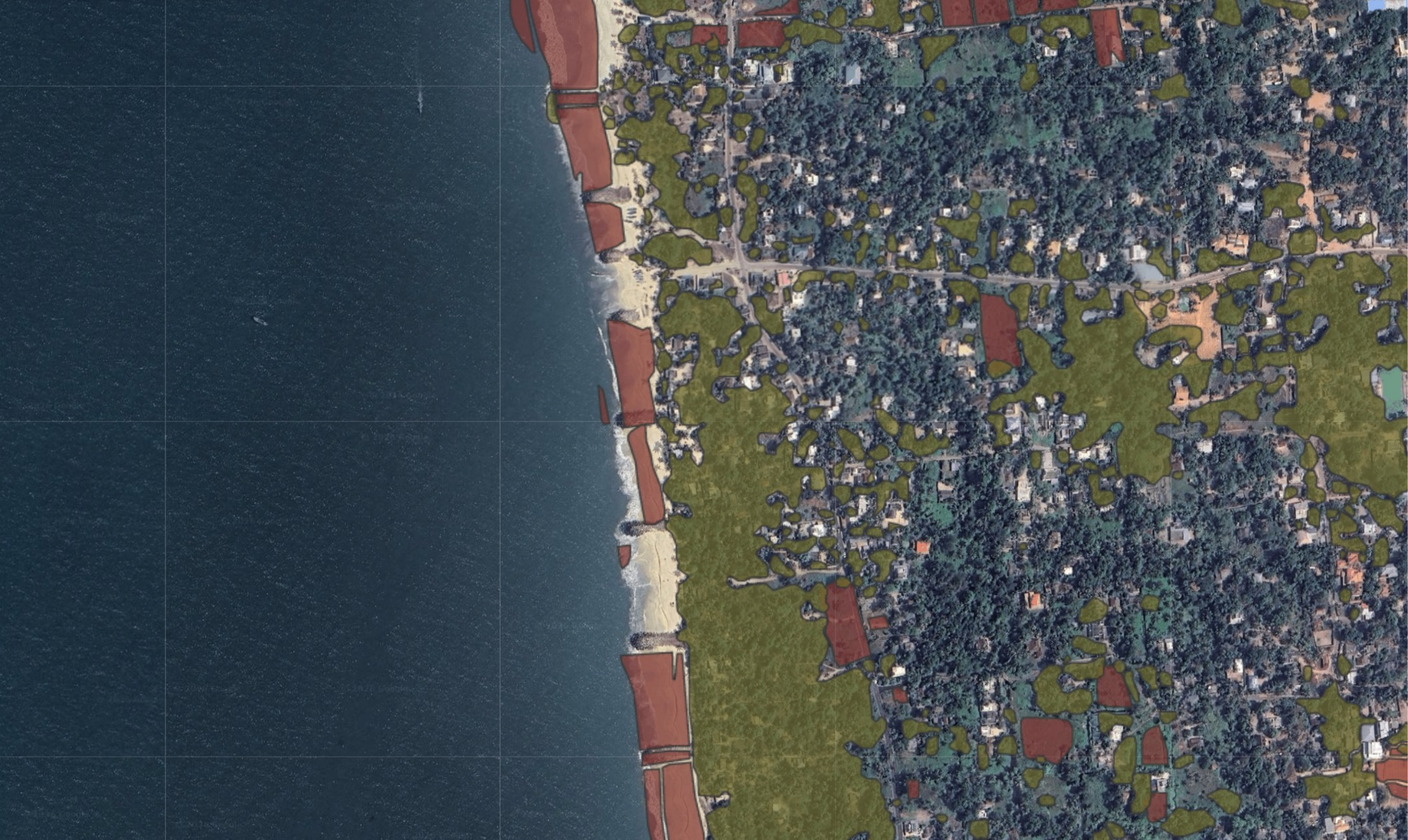}
    \includegraphics[width=0.49\textwidth]{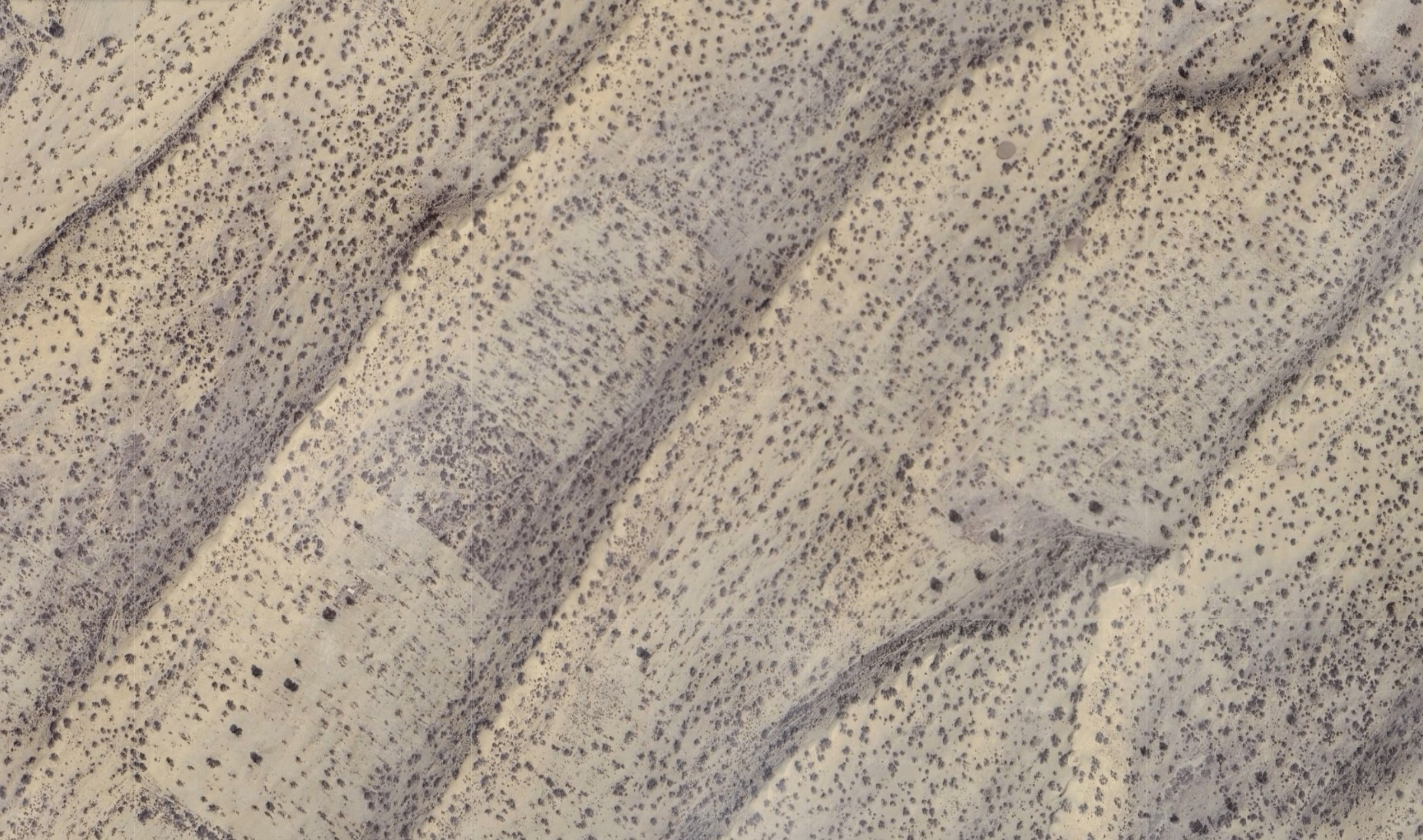}
    \includegraphics[width=0.49\textwidth]{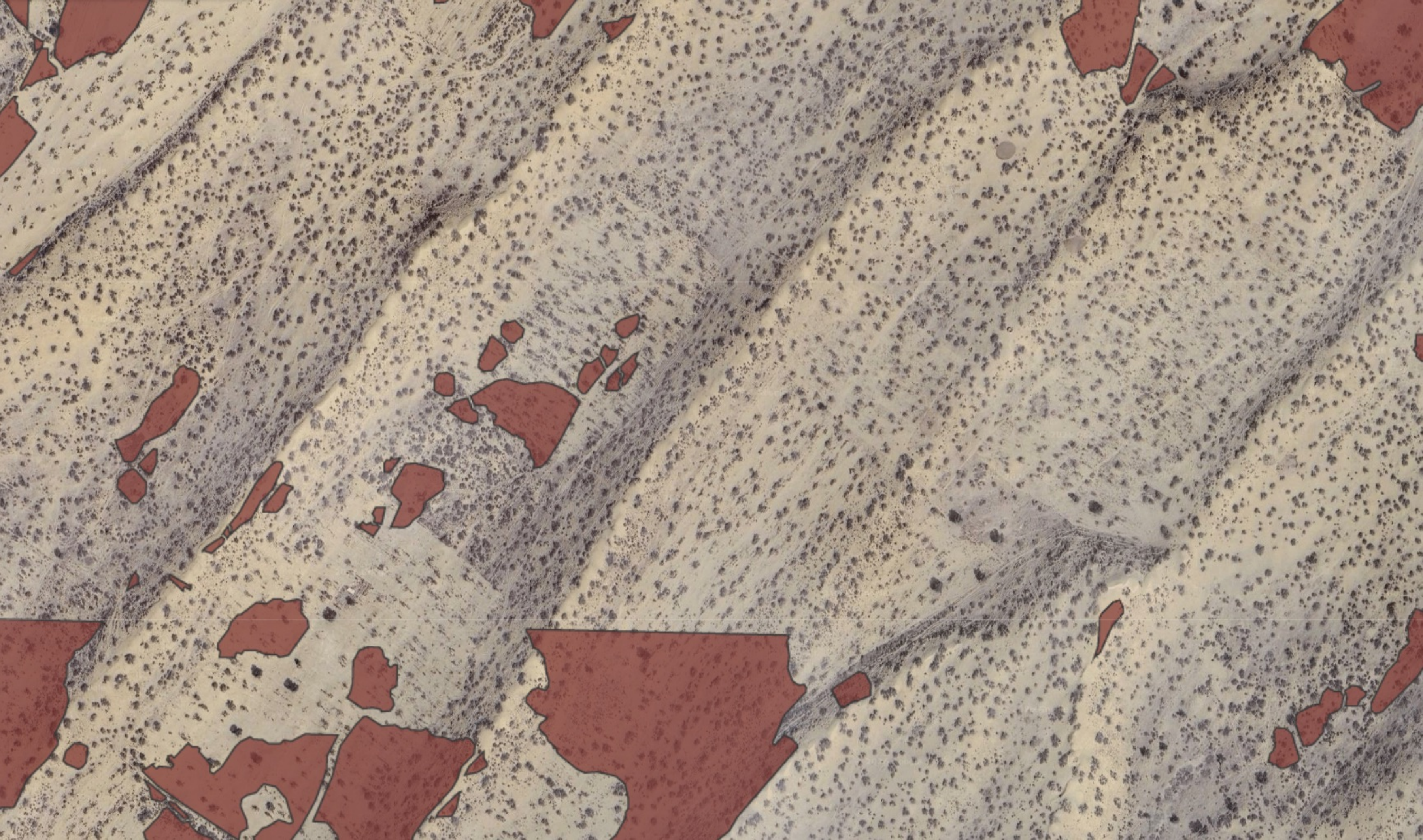}
    \includegraphics[width=0.2\textwidth]{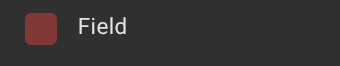}
    \includegraphics[width=0.2\textwidth]{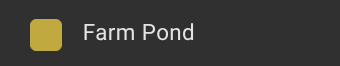}
    \includegraphics[width=0.2\textwidth]{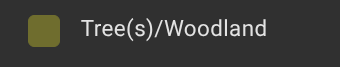}
    \caption{Errors seen in the ALU API in different parts of India (in order: Step cultivation in hilly regions of the north, mangroves in Sundarbans, coastal regions on the south-west coast, sand dunes from the deserts of the west). Left: satellite image and right: ALU outputs. Best viewed in color.}
    \label{fig:alu_api_errors}
\end{figure*}

We also examined the output of the ALU API in various parts of India, differing in terrain and agricultural practices. Figure~\ref{fig:alu_api_errors} shows the raw satellite images and the output of the ALU API. While ALU does identify fields in most parts well, it does not perform well in identifying curved field boundaries seen in step cultivation (row 1). ALU also wrongly classifies trees along river banks as fields in the mangroves of Sundarbans (row 2). Along coastal areas, ALU incorrectly identifies beaches as agricultural fields (row 3) and the peaks of sand dunes in deserts as field boundaries (row 4). These indicate the limitations of the current ALU model and opens up several possibilities for future research in this space, including translation to different regions and geographies.

\end{document}